\documentclass[10pt,letterpaper]{article}
\usepackage[left=2cm,right=2cm,top=2cm,bottom=2cm]{geometry}

\usepackage{times}
\usepackage{epsfig}
\usepackage{graphicx}
\usepackage{amsmath}
\usepackage{amssymb}

\usepackage{authblk}
\usepackage{lipsum}

\usepackage{color}
\usepackage{soul}
\usepackage{caption}
\usepackage{booktabs}
\usepackage{mdframed}
\usepackage{comment}
\usepackage{bm}
\usepackage{multirow}
\usepackage{subcaption}

\usepackage[pagebackref=true,breaklinks=true,colorlinks,bookmarks=false]{hyperref}

\begin{document}

%%%%%%%%% TITLE
\title{Learning Disentangled Latent Factors from Paired Data in Cross-Modal Retrieval: %\\
An Implicit Identifiable VAE Approach}

\author[1]{Minyoung Kim\thanks{mikim21@gmail.com}}
\author[1]{Ricardo Guerrero\thanks{ricardo.guerrero09@alumni.imperial.ac.uk}}
\author[1,2]{Vladimir Pavlovic\thanks{vladimir@cs.rutgers.edu}}
\affil[1]{Samsung AI Center Cambridge, UK}
\affil[2]{Dept. of Computer Science, Rutgers University, NJ, USA}

\maketitle

%%%%%%%%% ABSTRACT
\begin{abstract}
We deal with the problem of learning the underlying disentangled latent factors that are shared between the paired bi-modal data in cross-modal retrieval. Our assumption is that the data in both modalities are complex, structured, and high dimensional (e.g., image and text), for which the conventional deep auto-encoding latent variable models such as the Variational Autoencoder (VAE) often suffer from difficulty of accurate decoder training or realistic synthesis. A suboptimally trained decoder can potentially harm the model's capability of identifying the true factors. %Instead of  investigating how to learn the decoder better, 
In this paper we propose a novel idea of the implicit decoder, which completely removes the ambient data decoding module from a latent variable model, %we rather consider the cross-modal retrieval setting where the task of data synthesis is replaced by the task of search in a large database. 
%we rather consider the cross-modal retrieval setting where the model's conditional likelihood of the target modality data given the query needs to be accurately learned. To avoid the difficult decoder modeling in the conditional likelihood, 
via implicit encoder inversion that is achieved by Jacobian regularization of the low-dimensional embedding function. 
Motivated from the recent Identifiable-VAE (IVAE) model, %where the true parameters of the model are provably identified with additional auxiliary input, 
we modify it to incorporate the query modality data as conditioning auxiliary input, which %enhances the model identifiability, and 
allows us to prove that the true parameters of the model can be identifiable under some regularity conditions. %with the assumption of availability of the auxiliary information as a conditioning input.
Tested on various datasets where the true factors are fully/partially available, our model is shown to identify the factors accurately, significantly outperforming conventional encoder-decoder latent variable models. We also test our model on the Recipe1M, the large-scale food image/recipe dataset, where the learned factors by our approach highly coincide with the most pronounced food factors %that are widely agreed on, 
including savoriness, wateriness, and greenness. 
\end{abstract}

%%%%%%%%%%%%%%%%%%%%%%%%%%%%%%%%%%%%%%%%%%%%%%%%%%%%%%%%%%%%%%%%%%%%%%%%%%%%%%%
%%%%%%%%%%%%%%%%%%%%%%%%%%%%%%%%%%%%%%%%%%%%%%%%%%%%%%%%%%%%%%%%%%%%%%%%%%%%%%%
\section{Introduction}\label{sec:intro}

%The demand for analyzing and modeling multi-modal data has been unprecedentedly high, and received significant attention recently in computer vision communities. 
Accurately learning the underlying factors that explain the source of variability in the complex structured data (e.g., images), is one of the core problems in representation learning in computer vision. Whereas there has been considerable research work and significant success in the  unsupervised scenario recently~\cite{infogan16,aae16,beta_vae17,nica17,dip_vae18,factor_vae18,tcvae,bfvae}, there remains fundamental difficulty in identifying the true factors due to some known challenges~\cite{locatello19,disent_teh}. Having the factor labels can significantly boost the performance, but it is inherently difficult and costly to annotate data with true factors. 
As a remedy for this, we deal with the paired bi-modal data setup %in cross-modal retrieval 
as means of partial supervision. %, %a partially supervised learning scenario. 
%Considering the inherent difficulty of annotating true factors for data instances, this 
%which is deemed to require least amount of supervision cost. 
Formally, we consider paired data $({\bf x}_1,{\bf x}_2)$, where the goal is to learn latent factors that are shared between the two. Collecting paired data requires least amount of supervision, and can often be automated (e.g., web scrapping to pair image (${\bf x}_1$) and the text (${\bf x}_2$) that reside in the same web page)~\cite{salvador2017learning,marin2019learning}. 

While the CCA and its recent deep nonlinear extension, Deep CCA~\cite{dcca,deep_vcca}, are the most popular models to extract shared features (subspace) for such paired data, the dependency of CCA's cross-covariance matrix on the entire embedded data features hinders the mini-batch stochastic gradient methods from being applied in a principled manner. %, meaning that the approach may not be  scalable to large data. 
Moreover, applying the latent variable generative models like the Variational Autoencoder (VAE)~\cite{vae14} and its bi-modal extension (Bi-VAE)~\cite{wgvae}, %(and other related approaches; see Sec.~\ref{sec:related}). 
%The VAE-like purely generative models 
requires the latent-to-data synthesis/decoder modules, and learning the decoders for the complex high-dim ambient data is often very difficult. 
%\footnote{VAE synthesis is often blurry and less realistic than GAN's~\cite{gan}. Recent attempts for VAE to remedy this~\cite{vaegan,huang2018introvae,vqvae2} result in more GAN like models, that is, synthesis quality is improved at the expense of quality of the learned latent representations.}, 
A suboptimally trained decoder can potentially harm the model's capability of identifying the true factors. 

In the retrieval setup, where we let w.l.o.g., ${\bf x}_1$ be the query and ${\bf x}_2$ the search item, it is well known that the simple idea of aligning the pairs in the low-dim embedded space via cosine similarity, dubbed Cos-Sim, %which is highly scalable, 
is shown to yield very good retrieval performance~\cite{salvador2017learning,marin2019learning}. Despite this success, Cos-Sim has no capability of identifying latent variables that can be controlled explicitly by users. Although the model can be extended to incorporate an extra bottleneck auto-encoding model on top of the embedded space, there is no theoretical underpinning that such extension can help discovering the true factors. 

Our main goal in this paper is to learn the latent factors from paired data in the retrieval setup. More specifically, we would like to identify all the factors, each as a single latent variable in the model, and it would be ideal if the learned latent variables are {\em disentangled} (i.e., avoid a single latent variable impacting on the variation of two or more factors at the same time), and {\em complete} or minimally redundant (i.e., each of the underlying true factors has to be explained by only one of the latent variables, not by multiple ones).

To accomplish these desiderata and address the issues of the existing approaches, we introduce two strategies. First, we adopt the recent Identifiable-VAE (IVAE) model~\cite{ica_vae} that can provably identify the true model parameters by having additional auxiliary input on which the latent variables are conditioned. We modify IVAE by incorporating the query data ${\bf x}_1$ as conditioning auxiliary input, which enhances the model identifiability under some regularity conditions. Secondly, to remove the difficult-to-learn ambient data decoding module, we propose a novel idea of the {\em implicit decoder}. The modified IVAE model is built on the embedded space, and the relevance score required in the retrieval is defined by implicit inversion of the embedding function, which is accomplished by the Jacobian-based regularization of the embedding function.

%we rather consider the cross-modal retrieval setting where the model's conditional likelihood of the target modality data given the query needs to be accurately learned. To avoid the difficult decoder modeling in the conditional likelihood, 

Tested on various datasets where the true factors are fully/partially available, our model is shown to identify the factors accurately, significantly outperforming conventional encoder-decoder latent variable models. We also test our model on the Recipe1M, the large-scale food image/recipe dataset, where the learned factors by our approach highly coincide with the most pronounced food factors %that are widely agreed on, 
including {\em savoriness}, {\em wateriness}, and {\em greenness}.

\textbf{Notations and background on embedding}. Many approaches in cross-modal retrieval~\cite{hashing_cao,hashing_jiang,hashing_zheng,Su_2019_ICCV,real_wang,real_peng,food_howtocook,food_rich,salvador2017learning,marin2019learning,food_r2gan}, including our model, adopt the so called {\em shared space embedding} strategy, introducing mappings from ambient data to the shared space $\mathcal{V}$. More specifically, ${\bf v}_i = {\bf e}_i({\bf x}_i)$ for $i=1,2$, are the low-dim embeddings induced by the deep nonlinear (neural net) functions ${\bf e}_i(\cdot)$. 
The Cos-Sim model, for instance, learns the embedding functions by enforcing the embedded vectors of the paired data to be aligned well, with the cosine-angle of two embeddings $\cos({\bf v}_1, {\bf v}_2)$ as the alignment score. 
Its latent variable model extension discussed previously, introduces additional bottleneck encoding/decoding layers, ${\bf v}_1 \rightarrow {\bf z}$ and ${\bf z} \rightarrow {\bf v}_1'$, where the relevance score is measured by $\cos({\bf v}_1', {\bf v}_2)$. Then the latent variables ${\bf z}$ can be controlled directly (e.g., latent traversal). % under this model.
Instead of this ad-hoc model, in the next section we build the identifiable probabilistic latent variable model (IVAE) on the embedded space, while we avoid the ambient data synthesis module via Jacobian-based regularization of the embedding function.

%%%%%%%%%%%%%%%%%%%%%%%%%%%%%%%%%%%%%%%%%%%%%%%%%%%%%%%%%%%%%%%%%%%%%%%%%%%%%%%
%%%%%%%%%%%%%%%%%%%%%%%%%%%%%%%%%%%%%%%%%%%%%%%%%%%%%%%%%%%%%%%%%%%%%%%%%%%%%%%
\section{Implicit Identifiable Retrieval VAE %(IIR-VAE)
}\label{sec:main}

Our model is motivated from the recent work of the Identifiable VAE (IVAE for short)~\cite{ica_vae}, in which the true model can be provably identified by having a latent variable model with an auxiliary inputs on which the latent variables are conditioned. We modify this model to circumvent difficult synthesis model learning in the  cross-modal retrieval setup (dubbed \textbf{Retrieval IVAE} or \textbf{RIVAE} for short). %, where we introduce the Jacobian-based regularization for implicit inversion of the embedding function. 
%
%treats the query as the conditioning auxiliary input, thus increasing the chance of identifying true disentangled factors. 
%
Considering that our goal is to %make the model capable of learning 
learn the underlying disentangled latent representation that explains the sources of data variability, the provable model identifiability %is important as it would increase the chance of identifying 
allows us to recover true disentangled latent factors accurately, provided that the %underlying
data generating process admits disentangled factors. 

%For the VAE $p_{\bm\theta}({\bf v},{\bf z})$ on the embedded space, 
Unlike conventional VAE models, the IVAE adopts the so-called auxiliary input ${\bf u}$ on which the latent ${\bf z}$ is conditioned a priori, i.e.,  $P_{\bm\lambda}({\bf z}|{\bf u})$. In our retrieval setup, we regard the query ${\bf v}_1$ as the auxiliary input, and treat the target embedding ${\bf v}_2$ as the observed data. 
There are several modeling assumptions to make the IVAE model fully identifiable from the augmented data $\{({\bf v}_1,{\bf v}_2)\}$. Among others, the most important are: 
1) the conditional prior $P_{\bm\lambda}({\bf z}|{\bf v}_1)$ needs to be a (conditionally) factorized exponential family distribution with natural parameters $\bm{\lambda}$, and 
2) the output decoding process $P_{\bm\theta}({\bf v}_2|{\bf z})$ is homo-scedastic (i.e., ${\bf v}_2 = {\bf f}({\bf z}) + \bm{\xi}$ where ${\bf f}({\bf z}) = \mathbb{E}[{\bf v}_2|{\bf z}]$ and $\bm{\xi}$ is $0$-mean random and independent of ${\bf z}$). %To frame it within our model, we consider the query embedding ${\bf v}_1$ as the auxiliary input, and 
Then we define the following components to meet the IVAE's modeling assumptions:
%%%%
\begin{align}
P_{\bm\lambda}({\bf z}|{\bf v}_1) &= \mathcal{N}(\bm{\mu}({\bf v}_1), \textrm{Diag}(\bm{\sigma}({\bf v}_1))) \label{eq:dfrivae_p_z_v1} \\ 
P_{\bm\theta}({\bf v}_2|{\bf z}) &= \mathcal{N}({\bf f}({\bf z}), \eta^2 {\bf I}) \\
%P({\bf z}|{\bf v}_1) &:= \mathcal{N}({\bf z}; \bm{\mu}({\bf v}_1), \textrm{Diag}(\bm{\sigma}({\bf v}_1))) \label{eq:dfrivae_p_z_v1} \\
%P({\bf v}_2|{\bf z}) &:= \mathcal{N}({\bf v}_2; {\bf f}({\bf z}), \eta^2 {\bf I}) \label{eq:dfrivae_p_v2_z} \\
Q_{\bm\phi}({\bf z}|{\bf v}_1,{\bf v}_2) &= \mathcal{N}({\bf m}({\bf v}_1,{\bf v}_2), \textrm{Diag}({\bf s}({\bf v}_1,{\bf v}_2))) \label{eq:dfrivae_q_z_v1_v2} 
\end{align}
%%%%
where $\bm{\mu}(\cdot)$, $\bm{\sigma}(\cdot)$, ${\bf f}(\cdot)$, ${\bf m}(\cdot,\cdot)$, and ${\bf s}(\cdot,\cdot)$ are all deep neural networks (whose parameters are denoted by ${\bm\lambda}$, ${\bm\theta}$, ${\bm\phi}$, resp.) and $\eta^2$ is small fixed noise variance (e.g., $\eta=10^{-3}$). Hence, if the query embedding (auxiliary input) ${\bf v}_1$ retains all salient information about the underlying shared 
factors\footnote{This will be achieved by the cross-modal retrieval loss (\ref{eq:retr_loss}). %, we can urge the embedding vector ${\bf v}_1$ to retain all information about the shared factors.
}, then the true process of ${\bf v}_2|{\bf v}_1$, presumably following the model structure of (\ref{eq:dfrivae_p_z_v1}--\ref{eq:dfrivae_q_z_v1_v2}), can be identified accurately (due to the Theorem 1 in~\cite{ica_vae}), and so are the true factors ${\bf z}$. %The model diagram of DFR-IVAE is depicted in  Fig.~\ref{fig:dfrivae}. 

To train the RIVAE, we first consider the lower bound (ELBO or $\mathcal{L}_\textrm{LB}$) of the data likelihood $\log P({\bf v}_2|{\bf v}_1)$, where
%%%%
\begin{equation}
\mathcal{L}_\textrm{LB} = \textrm{KL}( Q({\bf z}|{\bf v}_1,{\bf v}_2)||P({\bf z}|{\bf v}_1)) - \mathbb{E}_{Q}[\log P({\bf v}_2|{\bf z})] 
%\nonumber
\label{eq:elbo_ivae}
\end{equation}
%%%%
% ii) no TC loss is required, and iii) the single-modal inference $Q({\bf z}|{\bf v}_1)$ in the conditional likelihood (\ref{eq:dfrvae_p_v2_x1}) and the retrieval loss (\ref{eq:retr_loss}) is replaced by $P({\bf z}|{\bf v}_1)$ (thus no variational approximation, but exact).
Although we maximize the ELBO wrt the IVAE parameters ${\bm\Lambda} := ({\bm\lambda}$, ${\bm\theta}$, ${\bm\phi})$, it cannot be used to optimize the embedding networks. This is because the embedding networks can determine the auxiliary input ${\bf v}_1$ and the target observed ${\bf v}_2$ in an arbitrary way just to make the IVAE model fit to $({\bf v}_1, {\bf v}_2)$ very well. As an extreme case, the constant mappings ${\bf e}_1(\cdot) = {\bf e}_2(\cdot) = {\bf v}^\textrm{const}$, can lead to a perfect-fit model that is degenerate. To learn the embedders, we introduce another learning criterion related to cross-modal prediction. % performance. 

Let $({\bf x}_1,{\bf x}_2)$
%\sim P_d({\bf x}_1,{\bf x}_2)$ 
be a matching pair from the training data, while $({\bf x}_1,{\bf x}'_2)$ a mismatch one. It is desired for the model to score them as: %these two pairs of samples as follows:
%%%%
\begin{equation}
\log P({\bf x}_2 | {\bf x}_1) \gg \log P({\bf x}'_2 | {\bf x}_1).
\label{eq:ret_ineq}
\end{equation}
%%%%
%However, we cannot directly evaluate the cross-modal prediction score $\log P({\bf x}_2 | {\bf x}_1)$ in our DFR-VAE model, since there is no decoder component that generates the ambient space data ${\bf x}_2$. 
To avoid difficult synthesis modeling for ambient data ${\bf x}_2$ in (\ref{eq:ret_ineq}), %$\log P({\bf x}_2 | {\bf x}_1)$, 
we utilize the {\em rule of the change of random variables}. Assuming that the embedding function ${\bf e}_2(\cdot)$ is {\em injective}\footnote{Formally, ${\bf e}_2({\bf x}_2) \neq {\bf e}_2(\hat{{\bf x}}_2)$ if ${\bf x}_2 \neq \hat{{\bf x}}_2$. It allows the inverse ${\bf e}_2^{-1}(\cdot)$ to be defined. It is a reasonable assumption that is also considered in~\cite{ica_vae}. %can be easily met under mild regularity conditions.
},  %\st{invertible}\footnote{\hl{REMOVE: To be more precise, we need further technical conditions to be imposed here since the embedding function ${\bf v}_T = {\bf e}_T({\bf x}_T)$ itself is {\em structurally} not bijective, e.g., considering that we often have $\dim({\bf x}_T) \gg \dim({\bf v}_T)$. Thus what needs to be imposed as a regularity condition is that the data distribution of ${\bf x}_T$ forms a low-dim {\em manifold} whose dimensionality equals $\dim({\bf v}_T)$. This technical assumption is fairly reasonable and widely accepted in many computer vision tasks (e.g., all valid facial images lie on a low-dim manifold). With this regularity condition, the Jacobian in (\ref{eq:ce_1}) can be considered as a square matrix within this implicit data manifold space, of which the determinant can be well defined.}}, 
%%%%
\begin{align}
\log P({\bf x}_2 | {\bf x}_1) &= \log P({\bf e}_2^{-1}({\bf v}_2) | {\bf x}_1) \\
%& \ \ \ \ \ \ (wrong!) = \log P({\bf v}_T | {\bf x}_I) + \log |\det\nabla {\bf e}_T({\bf x}_T) | \label{eq:ce_1x} \\
%& \ \ \ \ \ \ (correct!) 
%& 
&= \log P({\bf v}_2 | {\bf x}_1) + \log \textrm{vol} \nabla {\bf e}_2({\bf x}_2)  \label{eq:ce_1} \\
& \approx \log P({\bf v}_2 | {\bf x}_1) + \textrm{const.}
\label{eq:ce_2}
\end{align}
%%%%
%Here (\ref{eq:ce_1}) is due to the change of random variables. %In (\ref{eq:ce_2}) ``$\approx_c$'' means {\em approximately equal up to constant}, and 
To establish (\ref{eq:ce_2}), we will enforce the {\em Jacobian volume}, $\textrm{vol} \nabla {\bf e}_2({\bf x}_2)$ %$|\det\nabla {\bf e}_T({\bf x}_T)|$ 
to be constant over different points ${\bf x}_2$ as regularization %, which serves as regularization for embedding networks during training 
(See below for details). %(for details, see below how we regularize the embedding networks).
Using %$P({\bf v}_2 | {\bf x}_1) = P({\bf v}_2 | {\bf v}_1) \approx \mathbb{E}_{Q_1({\bf z}|{\bf v}_1)}[P({\bf v}_2 | {\bf z})]$, 
%%%%
\begin{equation}
P({\bf v}_2 | {\bf x}_1) = P({\bf v}_2 | {\bf v}_1) = \mathbb{E}_{P({\bf z}|{\bf v}_1)}[P({\bf v}_2 | {\bf z})],
\label{eq:dfrvae_p_v2_x1}
\end{equation}
%%%%
%and introducing additional margins, 
the desiderata (\ref{eq:ret_ineq}) can be %approximately 
encoded as the following loss: %retrieval loss: %minimization:
%hinge loss minimization:
%%%%
\begin{equation}
%\begin{align}
%\min_{\textrm{VAE},\textrm{Emb}} \Big( 1 + \mathbb{E}_{Q({\bf z}|{\bf v}_I)}[\log P({\bf v}'_T | {\bf z})] - \mathbb{E}_{Q({\bf z}|{\bf v}_I)}[\log P({\bf v}_T | {\bf z})] \Big)_+,
%\min_{\textrm{VAE},\textrm{Emb}} & \ \ \mathcal{L}_\textrm{Retrieval} := \label{eq:retr_loss} \\
%& \ \ \ \ \Big( 1 + \mathbb{E}_{Q({\bf z}|{\bf v}_I)} \big[ \log P({\bf v}'_T | {\bf z}) - \log P({\bf v}_T | {\bf z}) \big] \Big)_+  \nonumber 
%\end{align}
%\min_{\textrm{VAE},\textrm{Emb}} \ 
\mathcal{L}_\textrm{Retr} = \big( 1 + \mathbb{E}_{Q} [ \log P({\bf v}'_2 | {\bf z}) - \log P({\bf v}_2 | {\bf z}) ] \big)_+ 
\label{eq:retr_loss} 
\end{equation}
%%%%
where ${\bf v}'_2 = {\bf e}_2({\bf x}'_2)$ is the embedding of the mismatch sample ${\bf x}'_2$, %$\textrm{Emb}$ indicates the parameters of the embedding networks ${\bf e}_1(\cdot)$ and ${\bf e}_2(\cdot)$, 
and $(a)_+ = \max(0,a)$. %This way, %we build the loss function based on cross-modal retrieval $P({\bf x}_2|{\bf x}_1)$ 
That is, we implicitly invert the embedding function without a synthesis model for ${\bf x}_2$. %$P({\bf x}_2|{\bf v}_2)$. 

%This must be joint training of embedding networks and VAE

%%%%%%%%%%%%%%
%\subsubsection{Regularizing embedding networks}\label{sec:emb_reg}
\vspace{+0.5em}
\textbf{Regularizing the embedding network}. 
%Recall that 
% The derivation 
Derivation (\ref{eq:ce_2}) is valid only if %the determinant of the Jacobian of %the target modal embedder 
%${\bf e}_T({\bf x}_T)$ 
$\textrm{vol}\nabla {\bf e}_2({\bf x}_2)$ remains (approximately) constant across all plausible data samples ${\bf x}_2$. %\sim P_d({\bf x}_2)$. 
Directly regularizing this by minimizing a loss like $(\textrm{vol}\nabla {\bf e}_2({\bf x}_2) - c)^2$ for some constant $c$, is computationally prohibitive. %as one has to evaluate the Jacobian, its determinant, and the derivative of it. %, let alone optimizing.
However, since the Jacobian volume essentially measures the change in the function output due to small perturbations in the input, we can attain similar effect by enforcing the change originating from a random input 
% perturbation\footnote{Perturbation in the input space is done by pixel-wise perturbation for images, and on the word vector level for the text modality.} 
perturbation\footnote{The input space perturbation can be done pixel-wisely for images, and on the word vectors for text data.} 
to remain constant regardless of the input point. 
% To be more specific, 
Specifically, we impose the following regularization for ${\bf e}_2(\cdot)$:
%%%%
\begin{equation}
\mathcal{L}_\textrm{Reg} = \mathbb{E}_{{\bf x}_2,{\boldsymbol\epsilon}} \Big[ \big(||{\bf e}_2({\bf x}_2) - {\bf e}_2({\bf x}_2+{\boldsymbol \epsilon})|| - c \big)^2 \Big]
\label{eq:emb_reg}
\end{equation}
%%%%
where $\boldsymbol{\epsilon} \sim P(\boldsymbol{\epsilon})$ is a random sample from a noise distribution $P(\boldsymbol{\epsilon})$ with small magnitude (e.g., $||\boldsymbol{\epsilon}||=0.001$). We can also optimize $c$.

%%%%%%%%%%%
%\subsubsection{Summary}\label{sec:train_final}
\vspace{+0.5em}
\textbf{Summary}. The full training %procedure can be seen as alternation between (\ref{eq:obj_vae}) and the following 
can be written as the following optimization ($\lambda_\textrm{Retr}$ and $\lambda_\textrm{Reg}$ are the trade-off parameters). Note that the arguments in each loss $\mathcal{L}(\cdot)$ indicates which parameters should be updated regarding the loss. 
% for $\textrm{Reg}({\bf e}_2)$):
%%%%
\begin{equation}
%\max_{\textrm{VAE}} & \ \ -\textrm{ELBO} + \gamma \cdot \textrm{TC} \\
%\min_{\textrm{VAE},\textrm{Emb}} \ \ \mathcal{L}_\textrm{Retrieval} + \lambda \cdot \textrm{Reg}({\bf e}_2)
\min \ \mathcal{L}_\textrm{LB}({\bm\Lambda}) + \lambda_\textrm{Retr} \mathcal{L}_\textrm{Retr}({\bm\Lambda},{\bf W}) +  \lambda_\textrm{Reg} \mathcal{L}_\textrm{Reg}({\bf W})
\label{eq:obj_joint}
\end{equation}
%%%%

% %%%%%%%%%%%%%%%%%%%%%%%%%%%%%%%%%%%%%%%%%%%%%%%%%%%%%%%%%%%%%%%%%%%%%%%%%%%%%%%
% \subsection{Cross-Modal Retrieval and Latent Traversal}\label{sec:retrieval_traversal}

\textbf{Cross-modal Retrieval}. 
%Once our models are trained, the model needs to be able to
%To perform cross-modal retrieval with our model, %Suppose we have the test dataset for the retrieval search, denoted by $\mathcal{D}_2 \ni {\bf x}_2 \sim P_d({\bf x}_2)$. 
Basically we need to solve: $\arg\max_{{\bf x}_2 \in \mathcal{D}_2} \log P({\bf x}_2|{\bf x}_1)$, where 
${\bf x}_1$ is the query data point, and $\mathcal{D}_2$ is the search database.
%%%%
%\begin{equation}
%\arg\max_{{\bf x}_2 \in \mathcal{D}_2} \log P({\bf x}_2|{\bf x}_1).
%\end{equation}
%%%%
Using the derivation of (\ref{eq:ce_1}--\ref{eq:ce_2}), $\log P({\bf x}_2|{\bf x}_1)$ can be approximated as $\log P({\bf v}_2|{\bf z})$ with ${\bf z}\sim P({\bf z}|{\bf v}_1)$, which results in the following three-step algorithm: %for the cross-modal retrieval:
%%%%
\begin{mdframed}
\vspace{-0.3em}
\begin{enumerate}
%\vspace{-0.3em}
\item Embed the query point: ${\bf v}_1 = {\bf e}_1({\bf x}_1)$.
\vspace{-0.6em}
\item Sample ${\bf z}\sim P({\bf z}|{\bf v}_1)$.
\vspace{-0.6em}
\item Solve: $\arg\max_{{\bf x}_2 \in \mathcal{D}_2} \log P({\bf e}_2({\bf x}_2)|{\bf z})$.
%\vspace{-0.3em}
\end{enumerate}
%%%%
\vspace{-0.1em}
\end{mdframed}

\textbf{Latent Traversal}. 
While varying the value of one particular latent dimension with the rest being fixed, we inspect the change in the retrieved data. % for both modalities, 
This helps us understand what type of aspect each latent dimension corresponds to, that is, the source of variability.
%In the standard VAE models, the latent traversal is typically done by 
%At some reference latent point ${\bf z}^\textrm{Ref}$, we traverse along each $j$-th dimension (e.g., $z_j \in \{-10.0, -9.9, \dots, +9.9, +10.0\}$ and $z_i = z^\textrm{Ref}_i$ for $i \neq j$), and 
We retrieve the output for each traversed point ${\bf z}$, namely the mode of $P({\bf x}_2|{\bf z})$. %The set of generated data points ${\bf x}$ would show us the role of the latent dimension $j$ in our model. 
%Unfortunately, we cannot directly apply this approach to our model since we have no decoder component. However, we can approximately perform this task by retrieval: For each traversed point ${\bf z}$, 
We regard the conditional distribution $P({\bf v}_2|{\bf z})$ as a proxy of $P({\bf x}_2|{\bf z})$ using the derivation similar to  (\ref{eq:ce_1}--\ref{eq:ce_2}) with the regularized embedding networks. %Then $P({\bf v}_2|{\bf z})$ can be used to score the data points in $\mathcal{D}_2$, among which we select the highest scored sample as the output for ${\bf z}$. 
%This idea leads to the following three-step algorithm for
The %retrieval-based 
latent traversal algorithm is summarized as follows:
%%%%
\begin{mdframed}
\vspace{-0.3em}
\begin{enumerate}
\item Traverse ${\bf z}$ along the $j$-th dim ($j=1,\dots,d$).
\vspace{-0.6em}
\item Prepare the conditional distribution: $P({\bf v}_2|{\bf z})$.
\vspace{-0.6em}
\item Solve: $\arg\max_{{\bf x}_2 \in \mathcal{D}_2} \log P({\bf e}_2({\bf x}_2)|{\bf z})$.
\end{enumerate}
\vspace{-0.1em}
\end{mdframed}

\textbf{Comparison to Conditional VAE~\cite{cond_vae}}. 
The idea of conditioning the latent variables ${\bf z}$ %on the given information 
is similar to the Conditional VAE (CVAE)~\cite{cond_vae}. However, there are two main differences: i) The conditioning variables in the CVAE are typically class labels, and its main goal is to enrich the representational capacity of VAE to cover different regions/clusters of the data domain indexed by the class label. ii) %Along this line, 
CVAE explicitly models dependency between the observed and the conditioning variables. On the other hand, in our RIVAE, there is no direct link between the output ${\bf v}_2$ and the auxiliary ${\bf v}_1$, thus enforcing core information from ${\bf v}_1$ to ${\bf v}_2$ to flow through the bottleneck ${\bf z}$. This allows the latents to learn the salient factors more effectively.

%%%%%%%%%%%%%%%%%%%%%%%%%%%%%%%%%%%%%%%%%%%%%%%%%%%%%%%%%%%%%%%%%%%%%%%%%%%%%%%
%%%%%%%%%%%%%%%%%%%%%%%%%%%%%%%%%%%%%%%%%%%%%%%%%%%%%%%%%%%%%%%%%%%%%%%%%%%%%%%
\section{Experimental Results}\label{sec:expmt}

We test our Retrieval-IVAE model on both controlled datasets where the ground-truth factors are fully/partially available for quantitative comparison, and the large-scale Recipe1M dataset~\cite{salvador2017learning,marin2019learning} in the context of food image to recipe retrieval. We especially focus on demonstrating our model's capability of identifying the underlying true factors with high degree of disentanglement, compared to the existing approaches and baselines (See Sec.~\ref{sec:methods_datasets}).

For the cross-modal retrieval at test time, we construct the search database $\mathcal{D}_2$, of size $|\mathcal{D}_2|=1000$ or $2000$, randomly selected from the test dataset. 
%The query data ${\bf x}_1$ are determined as the instances from the modality 1 that match data in $\mathcal{D}_2$. 
%The matching modality-1 data for each item in $\mathcal{D}_2$ constitutes the query points ${\bf x}_1$. 
We repeat this procedure randomly 10 times, and run the models to report average performance. 
For the retrieval metrics, we consider the median rank (Med-R) and the recall-at-$K$ ($R@K$) with $K=1,5,10$, where $R@K$ stands for the fraction (out of $|\mathcal{D}_2|$ queries) where the true item is found by the model in its top-$K$ scored items. %The median rank (Med-R) is the median of the ranks of the true items according to the model's retrieval scores. Hence the higher $R@K$ and the lower Med-R, the better.
Although these retrieval scores are  indicative of how well the models extract shared information, the main focus in this paper is to judge the goodness of the learned latent factors. For this purpose, see the quantitative metrics in Sec.~\ref{sec:quant_measures}.

%%%%%%%%%%%%%%%%%%%%%%%%%%%%%%%%%%%%%%%%%%%%%%%%%%%%%%%%%%%%%%%%%%%%%%%%%%%%%%%
\subsection{Competing Methods and Datasets}\label{sec:methods_datasets}

%%%%
%\textbf{Competing methods}. 
Our RIVAE is %Retrieval-IVAE model (denoted by \textbf{RIVAE}) is 
compared with the following methods: %existing methods or baselines:
\begin{itemize}
%
%\vspace{-0.1em}
\item \textbf{Cos-Sim-LVM}: %\hl{Cos-Sim-Bottleneck previously} 
As described previously, we extend the Cos-Sim %cosine-similarity-based 
embedded space alignment method %(aka Cos-Sim)~\cite{salvador2017learning,marin2019learning} 
to identify/control the latent variables. That is, this extended model has %additional functions/networks for
encoder (${\bf v}_1 \rightarrow {\bf z}$) and decoder (${\bf z} \rightarrow {\bf v}_1'$), both of which roughly correspond to $P({\bf z}|{\bf v}_1)$ and $P({\bf v}_2|{\bf z})$ in our RIVAE, respectively, and we use neural nets with similar architectures for fair comparison. %(See Supplement for the details of the networks.) 
%
%\vspace{-0.1em}
\item \textbf{Bi-VAE}: This is the bi-modal extension of the VAE %model~\cite{vae14} 
via the product-of-experts approximation~\cite{wgvae}. It requires difficult synthesis (decoder) model learning for the ambient data (i.e., ${\bf z} \rightarrow {\bf x}_1$ and ${\bf z} \rightarrow {\bf x}_2$). 
%
%\vspace{-0.1em}
\item \textbf{Bi-VAE-on-$\mathcal{V}$}: As a reasonable workaround to circumvent the difficult synthesis model learning in Bi-VAE, we can think of building the bi-modal VAE%~\cite{wgvae} 
on %top of 
the embedded space ($\mathcal{V}$) instead. That is, the embedding networks are fixed (e.g.,  simply borrowed from the trained Cos-Sim model). The intuition is to regard the embeddings ${\bf v}_{1/2}$ as proxy for ambient data ${\bf x}_{1/2}$, and a similar idea was previously explored in~\cite{bivaeonv}. 
%and it corresponds to our RBi-VAE model with only the data fitting loss (VAE training alone) without the embedding loss (\ref{eq:obj_joint}). 
%
%\vspace{-0.1em}
\item \textbf{DCCA}\footnote{Other variants including DCCAE (DCCA with the  auto-encoding loss)~\cite{deep_vcca}, are often on a par with DCCA, and not considered here.}: The Deep CCA model~\cite{dcca} that learns the nonlinear mapping from inputs to the embedding vectors. % that maximizes the CCA's linear correlation score on the embedded space. 
%
%\vspace{-0.1em}
\item \textbf{RBi-VAE}: %\hl{DFR-VAE previously}
%As also discussed in the method section, 
We consider a {\em joint} model $P({\bf v}_1,{\bf v}_2,{\bf z})$ in place of the {\em conditional} $P({\bf v}_2,{\bf z}|{\bf v}_1)$ in our RIVAE. (Supplement for details) % of the model) 
Dubbed Retrieval-Bi-VAE (or RBi-VAE for short), we adopt the same Jacobian embedder regularization for the implicit embedder inversion, and it is compared to RIVAE to see how effective the identifiable model learning in RIVAE. % is in terms of the latent factor learning. 
\end{itemize}

For fair comparison, we %attempt to 
make the experimental setup as equal as possible for all competing methods. E.g., the number of latent variables adopted in the competing models is set to be the same. The details of the model architectures and optimization strategies %/hyperparameters 
are described in the Supplement.

%If not scalable, we don't run it (eg, WG-VAE for Recipe1M).

%%%%
%\textbf{Datasets}. 
And we test the above models on four datasets: 
\begin{itemize}
%
%\vspace{-0.5em}
\item \textbf{Synth} (Sec.~\ref{sec:expmt_synth}): The synthetic data generated from a nonlinear function with disentangled latent variables partitioned into shared and private factors. 
%
%\vspace{-0.5em}
\item \textbf{Sprites} (Sec.~\ref{sec:expmt_sprites4}): 
Binary images of sprites. % in two different 
The shape %two shapes (square and oval) are 
is regarded as modality, where the locations and size of the sprite are considered as shared factors. 
%
%\vspace{-0.5em}
\item \textbf{Split-MNIST} (Sec.~\ref{sec:expmt_split_mnist}): 
The bi-modal data created from the MNIST dataset by splitting images into left (modality-1) and right (modality-2) halves. %
%\vspace{-0.5em}
\item \textbf{Recipe1M} (Sec.~\ref{sec:expmt_im2recipe}): The large scale food dataset~\cite{salvador2017learning,marin2019learning} that consists of pairs of food image and recipe text including title, ingredient list, and instructions. 
\end{itemize}

%Due to the lack of space, for Sprites and Split-MNIST, we only show highlights of the results in Table~\ref{tab:sprites_split_mnist} and  Fig.~\ref{fig:sprites_splitmnist_highlights}, where full experimental results can be found in Supplement. 

%%%%%%%%%%%%%%%%%%%%%%%%%%%%%%%%%%%%%%%%%%%%%%%%%%%%%%%%%%%%%%%%%%%%%%%%%%%%%%%
\subsection{Metrics for Goodness of Learned Latents}\label{sec:quant_measures}

%As the main focus of this paper is to learn good latent representations that can capture the underlying variability of the data, 
We define metrics for the goodness of the learned latents, including the degree of disentanglement and completeness. 
Our quantitative measures are similar to the D/C/I metrics~\cite{williams18}, %used for judging goodness of the learned factors. The 
but %the key difference is that 
whereas in~\cite{williams18} they assume an offline data pool for which all %ground-truth 
factor labels are available, but in our case, not all factor labels are available for entire data instances for some datasets. %, and for only latent traversed results. 
To this end, we modify the original metrics, adapted to the latent traversal results as follows. 

%The following is how the modified D/C/I measures can be evaluated for the latent traversal results:
For each reference/query ${\bf x}_1^{\textrm{Ref}}$, we embed/encode it to obtain the latent vector ${\bf z}^{\textrm{Ref}}$. Then from this anchor point ${\bf z}^{\textrm{Ref}}$, we traverse along the $z_j$ axis for each latent dimension $j=1,\dots,d$, i.e., varying the value of $z_j$ while freezing the rest dimensions. We then collect a set of retrieved items ${\bf x}_2$, each of which corresponds to each of the traversed points along $z_j$ axis. And for each retrieved item, its true factor values $[f_1,\dots,f_K]$ are looked up. This way, we collect the paired $(z_j,f_k)$ data, and estimate the Pearson's correlation coefficient.  In particular, we deal with $c_{jk} = |\textrm{Corr}(z_j,f_k)|$, the absolute correlation score, where the larger $c_{jk}$ implies that the latent variable $z_j$ is more related to the true factor $f_k$. We repeat this procedure for many reference/query points, and take the average scores $c_{jk}$, which helps marginalizing out the instance-specific impact on the correlations between latents and true factors. 

Then with the $(d \times K)$ correlation table ${\bf c}$ (whose $(j,k)$-entry is $c_{jk}$), we measure the three metrics similarly as~\cite{williams18}. First, the {\em Disentanglement} metric ($D$) measures the degree of dedication of each latent variable $z_j$ in predicting $f_k$ against others $f_{-k}$. Intuitively, we have perfect disentanglement if each latent variable $z_j$ is correlated with only a single true factor $f_k$. On the other hand, if $z_j$ is correlated with multiple true factors at the same time, it is deemed entangled. To capture this idea, each $j$-th row of the table ${\bf c}$ is normalized to a probability distribution, specifically,
%%%%
\begin{equation}
p_{jk} = \frac{\rho(c_{jk})} {\sum_{k'=1}^d \rho(c_{jk'})},
\end{equation}
%%%%
where $\rho()$ is positive monotonic increasing (e.g., $\rho(c) = e^{\alpha c}$ for some $\alpha>0$). Then we compute the normalized ($[0,1]$-scaled) entropy, $H_j = - (1/{\log K}) \sum_k p_{jk} \log p_{jk}$, 
%%%%
% \begin{equation}
% H_j = \frac{\sum_{k=1}^d -p_{jk} \log p_{jk}} {\log K},
% \end{equation}
%%%%
and we define $D = 1 - (1/d) \sum_{j=1}^d H_j$ (the higher, the better disentangled). 
The {\em Completeness} metric ($C$) captures the degree of exclusive contribution of $z_j$ in predicting $f_k$ against others $z_{-j}$. As it essentially aims for minimal redundancy, we would achieve perfect completeness if variability of each true factor $f_k$ is explained by only a single latent variable $z_j$, instead of multiple latents. The metric $C$ can be computed similarly as $D$, by replacing all row-wise operations with column-wise. Likewise, the higher $C$ is, the better. Lastly, the {\em Informativeness} metric ($I$) measures how informative each latent variable is in predicting a true factor. To this end, for each $f_k$, we find the best predictor $z_j$ (i.e., $c_k^* = \max_{1\leq j\leq d} c_{jk}$), and define $I$ as the average of $c_k^*$ over $k=1,\dots,K$.

%%%%%%%%%%%%%%%%%%%%%%%%%%%%%%%%%%%%%%%%%%%%%%%%%%%%%%%%%%%%%%%%%%%%%%%%%%%%%%%
\subsection{Synthetic Data (Synth) %\hl{This section contains results using the CCA model, hence needs to be removed.
}\label{sec:expmt_synth}

%\hl{emphasize disentanglement (marginal improvement in retrieval performance)}

To demonstrate our model's capability of identifying the shared latent factors from bi-modal data, we devise a synthetic data setup as follows. First we generate 4-dim factors, ${\bf f} = [{\bf f}^S, f^1, f^2]$ which are all iid samples from $\mathcal{N}(0,1)$ with $\dim({\bf f}^S)=2$ and $\dim(f^1)=\dim(f^2)=1$. The ambient data points are then generated by the nonlinear functions, ${\bf x}_1 = {\bf G}_1({\bf f}^S, f^1)$ and ${\bf x}_2 = {\bf G}_2({\bf f}^S, f^2)$ where ${\bf G}_1(\cdot)$ and ${\bf G}_2(\cdot)$ are two-layer neural networks that output 50-dim vectors. % (i.e., $\dim({\bf x}_1)=\dim({\bf x}_2)=50$). 
So our intention is that ${\bf f}^S$ serves as the {\em shared factors} that govern the data variability in the two modalities, while $f^1$ and $f^2$ are the {\em private factors} that only affect individual modalities, being independent on the other. 
%
%Note that the key to accurate cross-modal retrieval is to correctly identify the shared factors ${\bf f}^S$ from a query input ${\bf x}_1$, and also separate them from the private $f^1$, so that the shared factors can be used to find the matching data instance ${\bf x}_2$ from the target modality. 
Additionally, it would be desirable if the model can further disentangle the two individual factors in ${\bf f}^S$. %, we could have better understanding, interpretability, and control of the real factors.

%%%%
\begin{table}%[t]
%\vspace{-0.4em}
\centering
\caption{(Synth) Retrieval performance with 1000 test samples as the search set. (averaged over 10 random runs).
}
\label{tab:synth_retrieval}
\small
%\footnotesize
%
\vspace{-1.0em}
\begin{tabular}{|c||c|c|c|c|}
\hline
Method & $R@1 \uparrow$ & $R@5 \uparrow$ & $R@10 \uparrow$ & Med-R $\downarrow$ \\
\hline\hline
%\st{Cos-Sim} & \st{0.26} & \st{0.66} & \st{0.82} & \st{3.00} \\
%\hline
Cos-Sim-LVM & 0.02 & 0.14 & 0.27 & 24.00 \\
\hline
Bi-VAE~\cite{wgvae} & 0.01 & 0.05 & 0.10 & 101.00 \\
\hline
%WG-VAE (Linear)~\cite{wgvae} & oom & oom & oom & oom \\
%\hline
Bi-VAE on $\mathcal{V}$ & 0.20 & 0.58 & 0.76 & 4.00 \\
\hline
DCCA~\cite{dcca} & 0.29 & 0.62 & 0.81 & 3.00 \\
\hline
RBi-VAE & 0.33 & 0.81 & 0.92 & 2.00 \\
\hline
%RBi-VAE ($\gamma=50.0$) & 0.83 & 1.00 & 1.00 & 1.00 \\
%\hline
RIVAE & 0.35 & 0.80 & 0.92 & 2.00 \\
\hline
\end{tabular}
%%
% \vskip -0.1in
\vspace{-0.8em}%\centering
\end{table}
%%%%

%From $3,000$ paired data samples $({\bf x}^1,{\bf x}^2)$ generated from the above model, the training, validation, and test sets are randomly split with equal proportions. %We first train the Cos-Sim model. 
For the embedding networks ${\bf e}_{1/2}(\cdot)$, we use two-layer neural networks, and the embedding dimension is set as $\dim({\bf v}_{1/2})=3$. %The results are shown in Table~\ref{tab:synth}. 
%For our DFR models %-IVAE and RBi-VAE models, %we use the CCA model on top of the embedding space with the 
We set $\dim({\bf z})=2$ for the latent spaces, which matches the number of true shared factors. The retrieval performance of the competing models is summarized in Table~\ref{tab:synth_retrieval}. Our RIVAE outperforms Cos-Sim-LVM and DCCA. %, while the generative RBi-VAE without the TC loss ($\gamma=0.0$) performs the best. 
%indicating that our latent representation learning also helps improving the retrieval scores over the previous retrieval approaches that merely aim to align or maximize the correlation of the embedded vectors in the shared embedded space. 
The poor performance of the Bi-VAE model implies that suboptimally trained decoders for high-dimensional noisy ambient data can degrade the retrieval performance significantly. Even the Bi-VAE-on-$\mathcal{V}$ trained on the fixed embedded space exhibits performance comparable to Cos-Sim-LVM, although it still underperforms our model. %This also signifies the importance of joint training of VAE and the embedding networks in our RBi-VAE model through the embedding loss. 

%%%%
\begin{figure}%[t!]
\vspace{-0.5em}
  \centering
  \begin{subfigure}{0.315\textwidth}
    \centering
    \includegraphics[trim = 0mm 3mm 0mm 6mm, clip, scale=0.313]{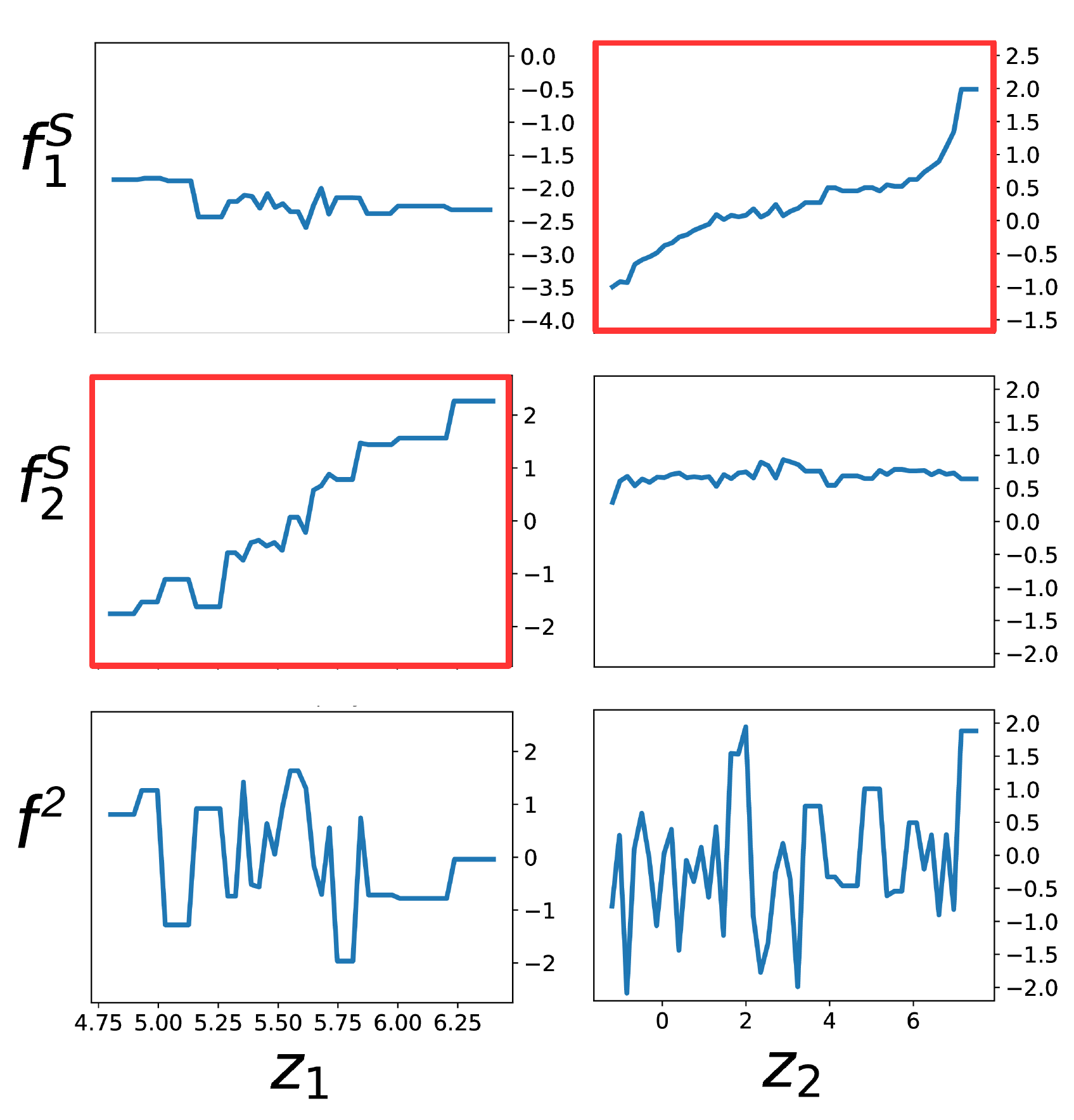}\caption{RIVAE}
    \label{fig:trv_synth_dfrivae}
  \end{subfigure}
  %\
  \begin{subfigure}{0.315\textwidth}
    \centering
    \includegraphics[trim = 0mm 3mm 0mm 6mm, clip, scale=0.313]{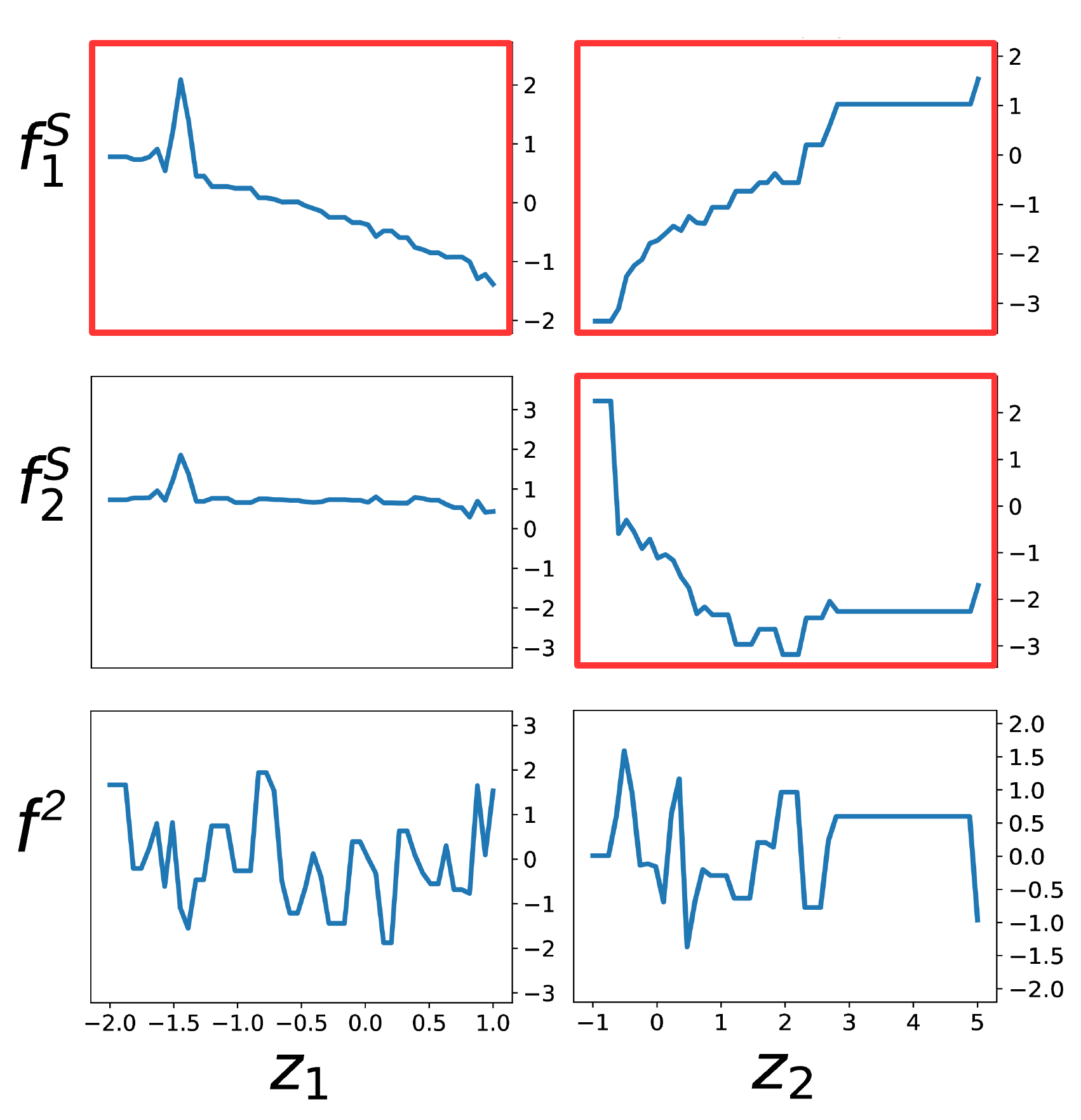}\caption{RBi-VAE}
    \label{fig:trv_synth_dfrvae}
  \end{subfigure}
  %\
  \begin{subfigure}{0.315\textwidth}
    \centering
    \includegraphics[trim = 0mm 3mm 0mm 6mm, clip, scale=0.313]{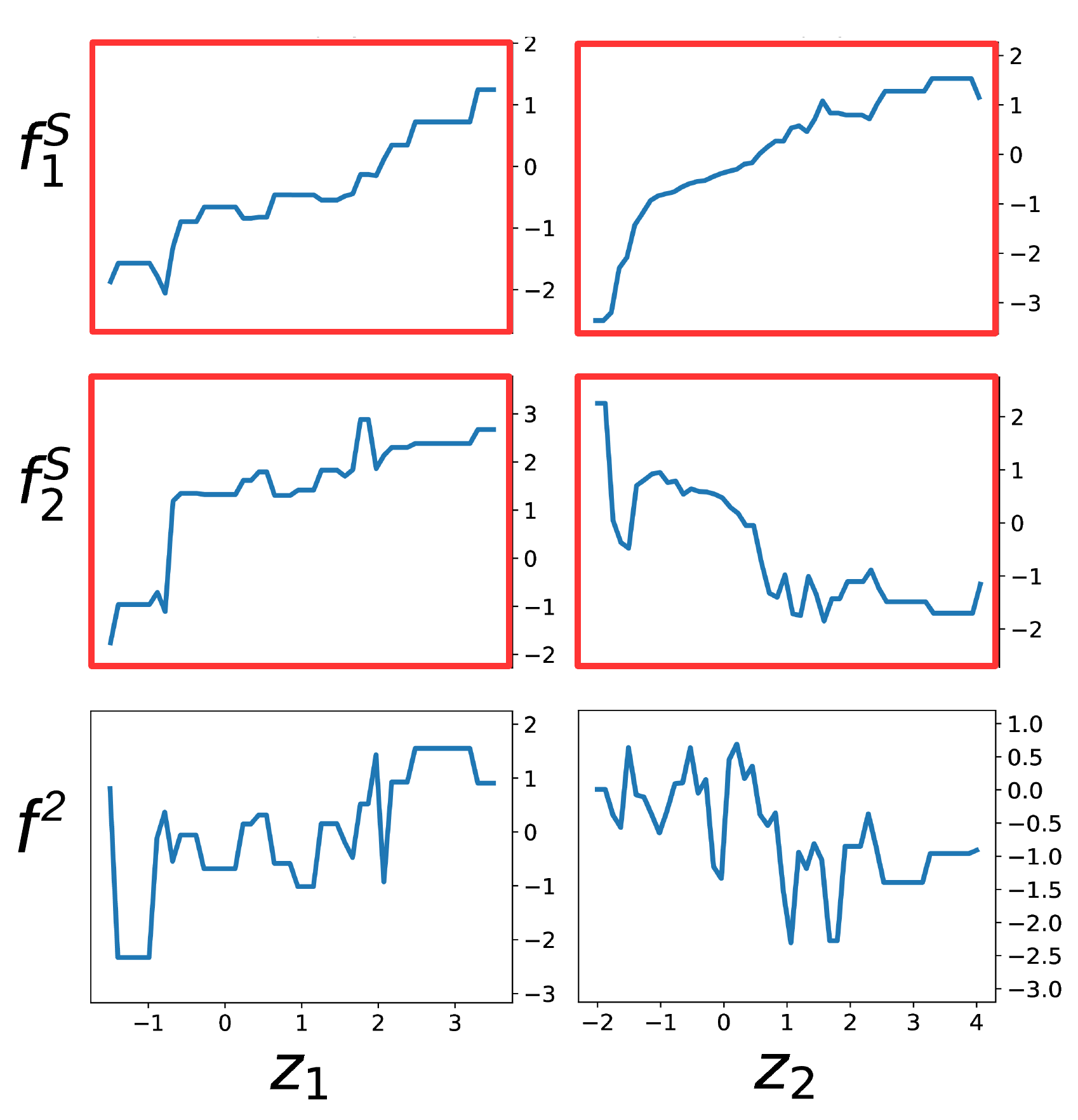}\caption{DCCA}
    \label{fig:trv_synth_dcca}
  \end{subfigure}
% \begin{center}
% \includegraphics[trim = 0mm 1mm 0mm 6mm, clip, scale=0.223]{figs/synth_dfr_ivae_f_vs_z.pdf} \ \
% \includegraphics[trim = 0mm 1mm 0mm 6mm, clip, scale=0.223]{figs/synth_dfr_vae_gam_0_f_vs_z.pdf} \ \
% % \includegraphics[trim = 0mm 1mm 0mm 6mm, clip, scale=0.225]{figs/synth_dfr_vae_gam_10_f_vs_z.pdf}
% \includegraphics[trim = 0mm 1mm 0mm 6mm, clip, scale=0.223]{figs/synth_dcca_f_vs_z.pdf}
% \end{center}
\vspace{-0.5em}
\caption{(Synth) True factors vs.~latent variables. %in latent traversal for 
%in RIVAE. %traversed for out RIVAE (Left), RBi-VAE w/ $\gamma=10.0$ (Middle), and DeepCCA (Right). 
Each column shows traversal of one latent $z_j$ with the other fixed. % and the rows show the individual 
The Y axes are true factors %(from top to bottom, $f^S_1$, $f^S_2$, and $f^2$), 
obtained from retrieved items ${\bf x}_2$. Red boxes indicate %are marked if the plots have 
significant changes in the true {\em shared factors}, %due to latent traversal, 
i.e., high correlation. 
%Visually, our RIVAE attains nearly one-to-one correspondence between the shared factors and the latents: $f_2^S$ explained by $z_1$ alone, and $f_1^S$ identified in $z_2$. Note that $f_1^S$ remains nearly unchanged while $z_1$ is varied (similarly for $f_2^S$ vs.~$z_2$). Also, the private (non-shared) factor $f^2$ of the retrieved items behaves almost randomly along the traversal $z_j$, which is expected (and desired) because $f^2$ is independent of the given query ${\bf x}^1$. RBi-VAE w/ $\gamma=10.0$ (not shown here) yielded nearly similar disentanglement quality as $\gamma=0.0$. 
%There is no considerable entanglement found here since no single latent variable $z_j$ captures two or more factors in it. 
}
\label{fig:synth_trv}
%\vspace{-1.5em}
\end{figure}
%%%%

Next we inspect the learned latent representations. We are especially interested in the correspondence between the learned latent variables ${\bf z}$ and the true shared factors ${\bf f}^S$. A desirable result would be exclusive one-to-one correspondence, where a single latent variable $z_j$ affects only one shared factor, independent from the other. To this end, we perform latent traversal. %: varying one variable $z_j$ while fixing the other as described in Sec.~\ref{sec:retrieval_traversal}. 
%, and for each traversal point ${\bf z}$ we select the best data instance in the target modality via ${\bf x}_T^*=\arg\max_{{\bf x}_T\in\mathcal{D}_T}\log P({\bf e}_T({\bf x}_T)|{\bf z})$ (recall Sec.~\ref{sec:retrieval_traversal}).
For the retrieved item ${\bf x}_2$, we look up its true factor values,  $(f^S_1, f^S_2, f^2)$, and plot each against $z_j$. The results %on the test retrieval set of size 1000 
are shown in Fig.~\ref{fig:synth_trv}. Notably for our RIVAE, each latent variable corresponds to only one true shared factor exclusively, implying that the disentangled factors are accurately identified. RBi-VAE partially identifies $f_1^S$ in $z_1$, but both shared factors are entangled in $z_2$. We also run the latent traversal with the DCCA, which is done by converting the learned CCA model to the dual-view latent variable linear Gaussian model following~\cite{pcca}. %, and performing cross-modal inference in closed forms. 
However, as this maximum-likelihood estimated model only fits well to the data, we see that each of the learned latent variables retain both factors entangled in it. 
%DCCA latent traversal: For the retrieval, one can form the Bach-Jordan's linear dual-view latent variable model on top of the $\bm{\phi}_{r}({\bf x}_{r})$ space, then apply cross modal inference as usual. 
Finally, the quantitative D/C/I metrics in Table~\ref{tab:synth_quant} show that our RIVAE yields significantly better latent representations than competing models.

%%%%
\begin{table}%[t!]
%\vspace{-2.0em}
\centering
\caption{(Synth) Goodness of the learned latents. %(averaged over 10 random runs).
}
\label{tab:synth_quant}
%\footnotesize
\small
\vspace{-1.0em}
\begin{tabular}{|c||c|c|c|}
\hline
 & Disent. $\uparrow$ & Comple. $\uparrow$ & Inform. $\uparrow$ \\
\hline\hline
Cos-Sim-LVM & 0.6169 & 0.7613 & 0.5486 \\ \hline
Bi-VAE & 0.4429 & 0.6822 & 0.6902 \\ \hline
Bi-VAE on $\mathcal{V}$ & 0.0520 & 0.8725 & 0.5630 \\
\hline
DCCA & 0.0017 & 0.2766 & 0.1371 \\ \hline
RBi-VAE & 0.0063 & 0.6900 & 0.3923 \\
\hline
RIVAE & ${\bf 0.9186}$ & ${\bf 0.8782}$ & ${\bf 0.8995}$ \\
\hline
\end{tabular}
%%
% \vskip -0.1in
\vspace{-0.8em}
\end{table}
%%%%

%%%%%%%%%%%%%%%%%%%%%%%%%%%%%%%%%%%%%%%%%%%%%%%%%%%%%%%%%%%%%%%%%%%%%%%%%%%%%%%
\subsection{Sprites}\label{sec:expmt_sprites4}

Using the benchmark dSprites dataset~\cite{dsprites}, we devise an experimental setup for the bi-modal retrieval task. First, we assume that the shape of sprites induces the modalities, specifically, ${\bf x}_1$ is {\em square}, and ${\bf x}_2$ {\em oval}. We then consider only the X, Y positions and the scale of the sprite as the underlying shared factors, with the other %varying source {\em rotation} 
factors being fixed. There are 32 variations in each of the X, Y positions and 6 variations in scale, which are independent from one another, resulting in 6144 samples. Image size is $(64 \times 64)$ pixels. %The three shared factors are denoted by: $f_1 = X$-pos, $f_2 = Y$-pos, and $f_3 = $ scale. 
%Sample pairs are shown in Fig.~\ref{fig:samples_sprites_split_mnist}. 
The dimension of the embedded space $\mathcal{V}$ is set to 10, and the latent space $\textrm{dim}({\bf z})=3$, which matches ground-truth.

\begin{figure}%[t!]
  \centering
  \begin{subfigure}{0.315\textwidth}
    \centering
    \includegraphics[trim = 0.5mm 2.5mm 1mm 1mm, clip, scale=0.615]{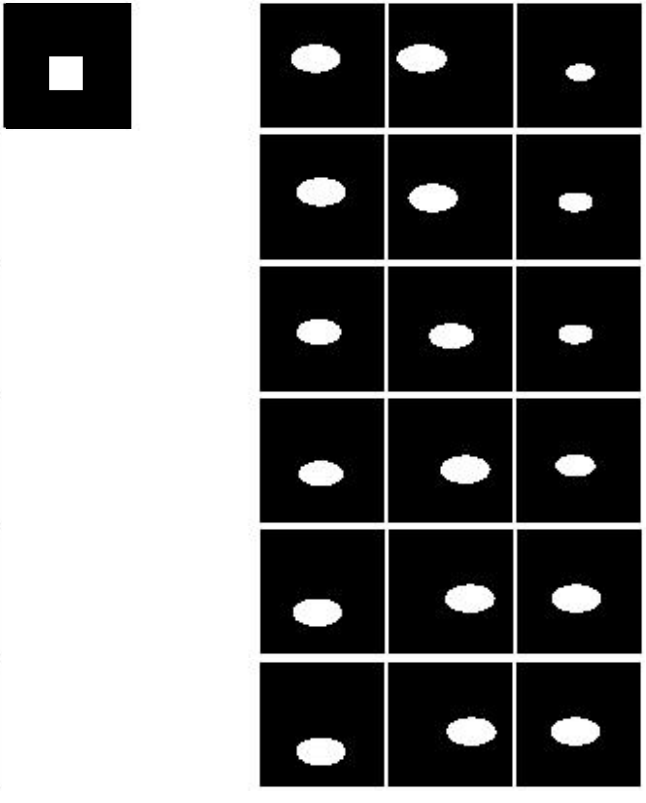} \ \ \ \ \ \ \ \ \caption{RIVAE}
    \label{fig:img_trv_sprites_dfrivae}
  \end{subfigure}
  \ \
%   \begin{subfigure}{0.235\textwidth}
%     \centering
%     \includegraphics[trim = 20mm 2mm 1mm 1mm, clip, scale=0.425]{figs/sprites4_dfrvae_gam0_latent_traversal.pdf}\caption{RBi-VAE $(0.0)$}
%     \label{fig:img_trv_sprites_dfrvae_gam0}
%   \end{subfigure}
%   \ \
  \begin{subfigure}{0.315\textwidth}
    \centering
    \includegraphics[trim = 20mm 2mm 1mm 1mm, clip, scale=0.625]{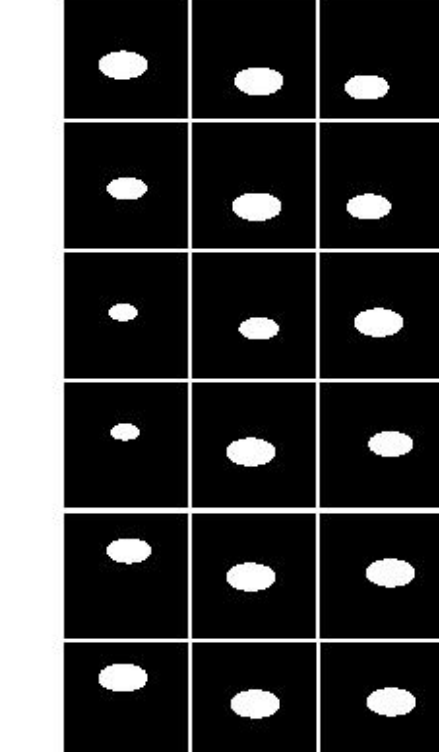}\ \ \ \caption{RBi-VAE}
    \label{fig:img_trv_sprites_dfrvae}
  \end{subfigure}
  \ \
  \begin{subfigure}{0.315\textwidth}
    \centering
    \includegraphics[trim = 20mm 2mm 1mm 1mm, clip, scale=0.625]{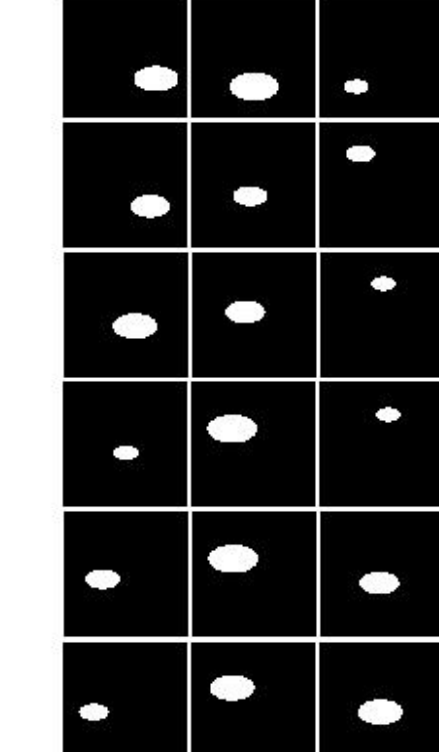}\caption{DCCA}
    \label{fig:img_trv_sprites_dcca}
  \end{subfigure}
\vspace{-0.8em}
\caption{(Sprites) Retrieved images from latent traversal. %\textbf{Left}: RIVAE. \textbf{Middle}: RBi-VAE with $\gamma=10.0$ and $50.0$. \textbf{Right}: DeepCCA~\cite{dcca}. 
The top left image is the query image that determines the reference point ${\bf z}^\textrm{Ref}$ % where the latent traversal is performed 
(the same for all models). For each model, each of the three columns depicts the retrieved images due to $z_1$, $z_2$, and $z_3$ changes (progresses vertically). Visually, it is clear that with  RIVAE, varying $z_1$ alone results in change in the $Y$-pos with the scale and X-pos intact, $z_2$ exclusively affects the X-pos, and $z_3$ only affects the scale. Such interpretation is not clear for other models. %However, the factors are significantly entangled in the latent variables of the DCCA model. 
}
\label{fig:sprites4_trv}
\vspace{-0.5em}
\end{figure}

The retrieval results on 1000 randomly selected search set are summarized in Table~\ref{tab:sprites4_retrieval}. Most approaches yield near perfect performance except for the decoder training models (Bi-VAE %\footnote{As stated in the main paper, %the poor performance of the Bi-VAE model 
%this implies that suboptimally trained decoders for high-dimensional ambient data can degrade the retrieval performance significantly.} 
 and Bi-VAE-on-$\mathcal{V}$), while our RIVAE performs marginally the best. 
%After training our model, 
The plots of the ground-truth factors of the retrieved images due to the latent space traversal, similar to Fig.~\ref{fig:synth_trv}, % are shown in Fig.~\ref{fig:sprites4_f_vs_z}. 
can be found in the Supplement, where our RIVAE shows near one-to-one correspondence between the true and learned factors, while other models exhibit considerable entanglement. %More specifically, for RIVAE, we see that $z_1$ {\em exclusively} corresponds to $f_3$ (highlighted by red box), $z_2=f_2$ and $z_3=f_1$, signifying that our model learns the disentangled latent representation very accurately.  
%On the other hand, for the DCCA, there is no exclusive correspondence, but significant entanglement. E.g., change in $z_1$ results in considerable changes in both $f_1$ and $f_2$, indicating that the scale and X-pos factors are entangled in the latent variable $z_1$. 
%our model identifies the disentangled latent representations, more specifically $z_1$ corresponds to Y-pos, while $z_2$ explains X-pos.
We also visualize the retrieved images obtained by latent traversal in Fig.~\ref{fig:sprites4_trv}. The result again verifies the high quality of disentanglement in the learned factors, very close to the true latent factors. 
The D/C/I scores in  Table~\ref{tab:sprites_quant} also support this claim quantitatively. 

%%%%

% %%%%
% \begin{figure}%[t!]
% %\vspace{-1.0em}
% \begin{center}
% \includegraphics[trim = 0mm 2mm 1mm 1mm, clip, scale=0.455]{figs/sprites4_latent_traversal.pdf} \ \ \ \ %\ \ \ \ \ \ 
% \includegraphics[trim = 20mm 2mm 1mm 1mm, clip, scale=0.455]{figs/sprites4_dfrvae_gam10_latent_traversal.pdf} \ \ \ \ 
% \includegraphics[trim = 20mm 2mm 1mm 1mm, clip, scale=0.455]{figs/sprites4_dfrvae_gam50_latent_traversal.pdf} \ \ \ \ 
% \includegraphics[trim = 20mm 2mm 1mm 1mm, clip, scale=0.455]{figs/sprites4_dcca_latent_traversal.pdf} 
% \end{center}
% \vspace{-1.8em}
% \caption{(Sprites) Retrieved images from latent traversal. \textbf{Left}: RIVAE. \textbf{Middle}: RBi-VAE with $\gamma=10.0$ and $50.0$. \textbf{Right}: DeepCCA~\cite{dcca}. 
% The top left image is the reference point ${\bf z}^\textrm{Ref}$ (the same for all models). Next to it, we depict the retrieved images due to $z_1$, $z_2$, and $z_3$ changes (progresses vertically). Visually, it is clear that with our DR-IVAE, varying $z_1$ alone results in change in the $Y$-pos with the scale and X-pos intact, $z_2$ exclusively affects the X-pos, and $z_3$ only affects the scale. However, it is not clear in the DeepCCA's traversal results.
% }
% \label{fig:sprites4_trv}
% \vspace{-0.5em}
% \end{figure}
% %%%%

%%%%
\begin{table}%[t]
%\vspace{-0.4em}
\centering
\caption{(Sprites) Retrieval performance among a 1000 randomly selected search set (averaged over 10 random runs). %\hl{Note: This RIVAE is the new conditional model (ICA-VAE, aka IVAE) conditioned on ${\bf v}_I$, hence we did not use the TC term in our loss function, and only maximize the ELBO.}
}
\label{tab:sprites4_retrieval}
%\footnotesize
\small
\vspace{-1.0em}
\begin{tabular}{|c||c|c|c|c|}
\hline
Method & $R@1 \uparrow$ & $R@5 \uparrow$ & $R@10 \uparrow$ & Med-R $\downarrow$ \\
\hline\hline
%\st{Cos-Sim} & \st{0.97} & \st{1.00} & \st{1.00} & \st{1.00} \\
%\hline
Cos-Sim-LVM & 0.13 & 0.46 & 0.67 & 6.00 \\
\hline
Bi-VAE~\cite{wgvae} & 0.02 & 0.07 & 0.10 & 127.75 \\
\hline
%WG-VAE (Linear)~\cite{wgvae} & oom & oom & oom & oom \\
%\hline
Bi-VAE on $\mathcal{V}$ & 0.69 & 0.97 & 0.99 & 1.00 \\
\hline
DCCA~\cite{dcca} & 0.96 & 1.00 & 1.00 & 1.00 \\
\hline
RBi-VAE & 0.99 & 1.00 & 1.00 & 1.00 \\
\hline
%RBi-VAE ($\gamma=50.0$) & 0.83 & 1.00 & 1.00 & 1.00 \\
%\hline
RIVAE & 1.00 & 1.00 & 1.00 & 1.00 \\
\hline
\end{tabular}
%%
% \vskip -0.1in
%\vspace{-1.5em}%\centering
\end{table}
%%%%

%%%%
\begin{table}[t!]
%\vspace{-2.0em}
\centering
\caption{(Sprites) Goodness of the learned latents. %(averaged over 10 random runs).
}
\label{tab:sprites_quant}
%\footnotesize
\small
\vspace{-1.0em}
\begin{tabular}{|c||c|c|c|}
\hline
 & {Disent. $\uparrow$} & {Comple. $\uparrow$} & {Inform. $\uparrow$} \\
\hline\hline
{Cos-Sim-LVM} & 0.6664 & 0.5634 & 0.8417 \\ \hline
{Bi-VAE} & 0.6821 & 0.0779 & 0.1259 \\ \hline
{Bi-VAE on $\mathcal{V}$} & 0.1367 & 0.6812 & 0.3119 \\
\hline
{DCCA} & 0.4163 & 0.1178 & 0.1432 \\ \hline
{RBi-VAE} & 0.5814 & 0.4863 & 0.7234 \\
\hline
\small{RIVAE} & ${\bf 0.8280}$ & ${\bf 0.9133}$ & ${\bf 0.8476}$ \\
\hline
\end{tabular}
%%
% \vskip -0.1in
\vspace{-0.8em}
\end{table}
%%%%

%%%%%%%%%%%%%%%%%%%%%%%%%%%%%%%%%%%%%%%%%%%%%%%%%%%%%%%%%%%%%%%%%%%%%%%%%%%%%%%
\subsection{Split-MNIST}\label{sec:expmt_split_mnist}

Following the setup in~\cite{dcca}, we form a retrieval setup from the MNIST dataset~\cite{Lecun98mnist} by taking the left half of each image as modality-1 and the right half as modality-2. So, each view contains images of size $(H=28 \times W=14)$ pixels. %Some example samples are shown in Fig.~\ref{fig:samples_sprites_split_mnist}. 
Note that the shared factors for both modalities would be the digit class and the writing style, which are deemed independent (disentangled) from each other. We followed the standard data splits, where 2000 images are randomly sampled from the test set to serve as the retrieval search set.

%%%%
\begin{table}%[t!]
%\vspace{-1.0em}
\centering
\caption{(Split-MNIST) Retrieval performance with 2000 randomly selected test images as the search set. % (averaged over 10 random runs).
}
\label{tab:split_mnist_retrieval}
%\footnotesize
\small
\vspace{-1.0em}
\begin{tabular}{|c||c|c|c|c|}
\hline
Method & $R@1 \uparrow$ & $R@5 \uparrow$ & $R@10 \uparrow$ & Med-R $\downarrow$ \\
\hline\hline
%\st{Cos-Sim} & \st{0.49} & \st{0.81} & \st{0.89} & \st{1.90} \\
%\hline
Cos-Sim-LVM & 0.19 & 0.46 & 0.61 & 6.35 \\
\hline
Bi-VAE~\cite{wgvae} & 0.01 & 0.03 & 0.05 & 524.80 \\
\hline
Bi-VAE on $\mathcal{V}$~\cite{wgvae} & 0.21 & 0.51 & 0.66 & 5.30 \\
\hline
DCCA~\cite{dcca} & 0.47 & 0.79 & 0.87 & 2.00 \\
\hline
RBi-VAE & 0.43 & 0.81 & 0.91 & 2.00 \\
\hline
%RBi-VAE ($\gamma=50.0$) & 0.34 & 0.76 & 0.88 & 2.00 \\
%\hline
RIVAE & 0.52 & 0.89 & 0.96 & 1.00 \\
\hline
\end{tabular}
%%
% \vskip -0.1in
%\vspace{-0.8em}\centering
\end{table}
%%%%

%%%%
\begin{figure}%[t!]
%\vspace{-0.8em}
\begin{center}
\includegraphics[trim = 0mm 0mm 0mm 16mm, clip, scale=0.825]{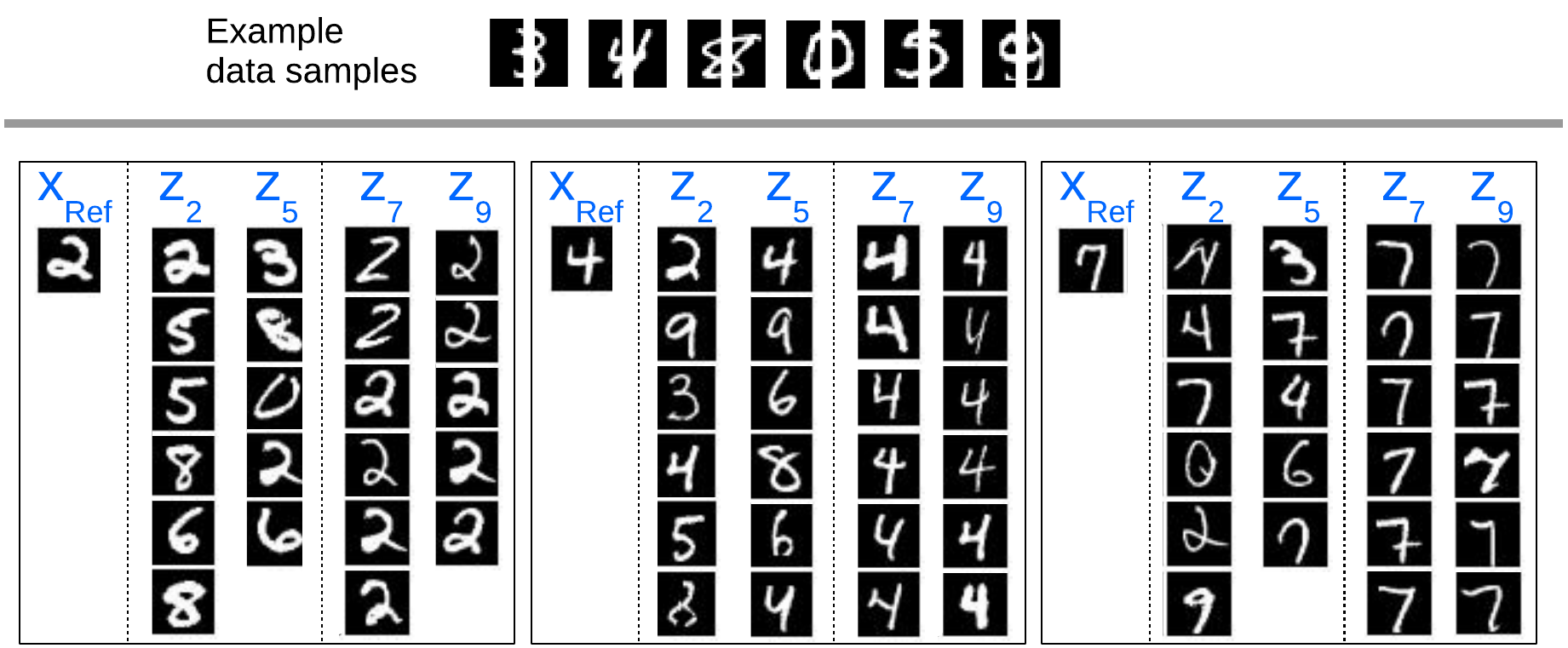}
\end{center}
\vspace{-1.5em}
\caption{(Split-MNIST) For three query references, latent traversal along four latent variables ($z_2,z_4,z_7,z_9$) in RIVAE. Visually, $z_2, z_5 = $ %correspond to 
digit class, 
%factors while 
$z_7,z_9=$ %explain variability of 
writing style. See Supplement for other query references.
}
%\vspace{-1.4em}
\label{fig:split_mnist_highlight}
\end{figure}
%%%%  

%The dimension for the embedded space is set to 50, and the latent space
We set $\textrm{dim}(\mathcal{V})=50$ and 
$\textrm{dim}({\bf z})=10$. The retrieval scores are reported in Table.~\ref{tab:split_mnist_retrieval}. Consistent with previous  experiments, our RIVAE attains the best scores. To see how the two underlying factors, writing style and digit class, are disentangled in the learned latent variables for our RIVAE, we show the latent traversal results visually in Fig.~\ref{fig:split_mnist_highlight}. %For the four highlight latent variables ($z_2$, $z_5$, $z_7$, $z_9$), we can see that two explain the writing style variation, while the other two correspond to the digit class. 
Refer to the caption of the figure for details.

%%%%%%%%%%%%%%
\subsubsection{Quantitative Analysis for Split-MNIST}

%The pair in this data is (${\bf x}_1 =$ left half, ${\bf x}_2 =$ right half), where 
Although we know that the shared factors between the left and right halves deem to be partitioned to those related to digit class and those non-digit related (e.g., writing style), there are only digit labels ($0\sim 9$) available, and it is not even clear how the writing style can be formally described or specified. 
%
%-- Hence, under this situation, to have reasonable quantitative measures of the learned factors, we devise the following two aspects that are reasonable enough to capture the underlying variability in the data: 1) Diversity in digit transitions along the latent traversal, and 2) Non-digit variation.
Hence, we devise some reasonable quantitative measures that can reflect how the underlying variability in digit transitions is captured in the model's latent variables. 

%\textbf{Diversity in digit transitions}. 
Let's say that we traverse along the axis $z_j$ in the latent space, while fixing the rest dimensions of ${\bf z}$.  After retrieving the data items, we record the number of unique digit transitions (e.g., $0 \to 2$ or $3 \to 8$) in the retrieved images in the traversal. The results, over all dimensions, can be summarized into a $(d \times 45)$ table, where $d = \dim({\bf z})$ and $45$ ($= 10C2$) is the number of direction-free transitions (e.g., $0 \to 2$ and $2 \to 0$ are regarded identical). Note that discarding the directions makes sense considering traversal in the reverse direction.
More concretely, if the digit classes of the retrieved items for the traversal $z_5$ are: $2\to 2\to 2\to 3\to 8\to 8\to 9\to 3\to 2\to 2\to 2$, then the fifth row of the table has value $1$ at four columns, $2\to 3$, $3\to 8$, $8\to 9$, $3\to 9$, with all the other entries $0$. There will be one such table for each reference query (${\bf x}_1^\textrm{Ref}$). And we will collect many (e.g., 1000) such  tables/queries, and take the average to get the global statistics. 

The key idea is that this (averaged) digit transition table can tell us possible digit transition types specific to each latent dimension. For instance, $z_0$ has large entries in the table for the transitions among digits $(1,4,7)$, while $z_1$ covers the clique $(2,3,5,8)$, and so on. And we expect that if latent factors are well trained, the transition cliques for different dimensions tend to be less overlapped, and overall the union of the cliques has good coverage of all 45 possible transitions. The former is related to {\em disentanglement} of the latent variables, and the latter being {\em thoroughness} or {\em coverage}, how many different %digit
transitions the learned latent variables can capture. 

%%%%
\begin{table}%[t!]
%\vspace{-2.0em}
\centering
\caption{(Split-MNIST) Goodness of the learned latents. Two quantitative measures regarding the variability in digit transitions. Interpretation: {\em overlap} is related to disentanglement (the lower, the better), and {\em coverage} to thoroughness (the higher, the better). The differences in the last column. %See text for the details. %(averaged over 10 random runs).
}
\label{tab:split_mnist_quant}
%\footnotesize
\small
\vspace{-1.0em}
\begin{tabular}{|c||c|c|c|}
\hline
 & \small{Overlap $\downarrow$} & \small{Coverage $\uparrow$} & \small{$C-O$ $\uparrow$} \\
\hline\hline
\small{Cos-Sim-LVM} & 0.0168 & 0.8881 & 0.8713 \\ \hline
\small{Bi-VAE} & ${\bf 0.0091}$ & 0.4863 & 0.4772 \\ \hline
\small{Bi-VAE on $\mathcal{V}$} & 0.0181 & 0.6418 & 0.6237 \\ \hline
\small{DCCA} & 0.0191 & 0.6377 & 0.6186 \\ \hline
\small{RBi-VAE} & 0.0189 & 0.8932 & 0.8743\\ 
\hline
\small{RIVAE} & 0.0121 & ${\bf 0.9675}$ & ${\bf 0.9554}$ \\
\hline
\end{tabular}
%%
% \vskip -0.1in
\vspace{-0.8em}
\end{table}
%%%%

To formalize, let's say $D = (d \times 45)$ be the table of averaged digit transitions. We normalize each row of $D$ as a probability distribution. Then we measure the overlap between two rows $i$ and $j$, e.g., $overlap(i,j) = \frac{1}{45} \sum_{k=1}^{45} \min(D[i,k], D[j,k])$, similarly as the histogram intersection. Then we measure the average overlap over all $i \neq j$. The smaller the overlap is, the better.
And, for the {\em coverage}, we take the union of the rows, e.g., simply the average of the rows in $D$. Then we compute the entropy of the union. The larger the entropy is, the union covers more digit transitions. %-- Then we report ({\em coverage} - {\em thoroughness}) as the goodness score of diversity in digit transitions. The higher,  the better. 
The results are summarized in Table~\ref{tab:split_mnist_quant}, and we see that the learned latent variables in our RIVAE exhibit low overlap and the highest coverage among others. %the competing approaches. 

\subsection{Food Image to Recipe Retrieval (Recipe1M)}\label{sec:expmt_im2recipe}

Recipe1M~\cite{salvador2017learning} is the dataset comprised of about 1M cooking recipes (titles, instructions, ingredients) and images. 
In this work, a subset of about $0.4$M recipes containing at least one image, no more than 20 ingredients or instructions, and at least one ingredient and instruction was used.
Data is split into $70\%/15\%/15\%$ train/validation/test sets.
% 
% 
%The task of a retrieval model is to find a matching cross-modal sample ${\bf x}_2$ given a query ${\bf x}_1$.
% The task of the retrieval model is to given a query modality from $\textbf{x1}$ find the matching cross-modal recipe pair in $\textbf{x2}$. 
The underlying embedding networks used 
% in the following experiments 
in this experiments 
combines architectural and training strategies from \cite{salvador2017learning,wang2019learning,chen2018deep}, and the related Cos-Sim model performs comparable to the state-of-the-arts.
% methods for the retrieval task. 
%This method, referred to as Cos-Sim,  serves as the initial $\mathcal{X} \mapsto \mathcal{V}$ mapping function for all comparing methods. 
% Similar to~\cite{salvador2017learning,wang2019learning,chen2018deep}, 
%Cos-Sim has a 
The embedded space has $\textrm{dim}(\mathcal{V})=1024$, and the latent variables $\textrm{dim}({\bf z})=30$.  %In order more closely compare to other VAE methods, 
%For baseline comparison, a bottleneck structure, in the form of two fully connected layers ($\mathcal{V}^{1024}\mapsto\mathcal{Z}^{30}\mapsto\mathcal{V}^{1024}$), is introduced in the embedded space, referred to as Cos-Sim (bottleneck).
%
% and the goal of the work proposed here is to disentangle possible hidden latent factors, we further reduce the dimensionality of Cos-Sim by introducing a bottleneck in the form of two fully connected layers ($\mathcal{V}^{1024}-\mathcal{Z}^{30}-\mathcal{V}^{1024}$) in latent space, we refer to this baseline as Cos-Sim (bottleneck).
%
%%%%
\begin{table}%[t!]
%\vspace{-1.0em}
\centering
\caption{(Recipe1M) Retrieval performance with size 1000 random %ly selected test recipes as the 
search set (averaged over 10 random runs).}
\label{tab:recipe1m_retrieval}
%\footnotesize
\small
\vspace{-1.0em}
\begin{tabular}{|c||c|c|c|c|}
\hline
Method & $R@1 \uparrow$ & $R@5 \uparrow$ & $R@10 \uparrow$ & Med-R $\downarrow$ \\
\hline\hline
%\st{Cos-Sim} & \st{0.48} & \st{0.78} & \st{0.85} & \st{1.90} \\
%\hline
Cos-Sim-LVM & 0.45 & 0.74 & 0.82 & 2.00 \\
\hline
Bi-VAE~\cite{wgvae} & Failed & Failed & Failed & Failed \\
\hline
Bi-VAE on $\mathcal{V}$~\cite{wgvae} & 0.22 & 0.45 & 0.55 & 7.70 \\
\hline
DCCA~\cite{dcca} & Failed & Failed & Failed & Failed \\
\hline
RBi-VAE & 0.29 & 0.56 & 0.66 & 4.00 \\
\hline
%RBi-VAE ($\gamma=50.0$) & 0.34 & 0.76 & 0.88 & 2.00 \\
%\hline
RIVAE & 0.39 & 0.70 & 0.79 & 2.00 \\
\hline
\end{tabular}
%%
% \vskip -0.1in
\vspace{-0.8em}%\centering
\end{table}
%%%%
%
Table~\ref{tab:recipe1m_retrieval} shows retrieval performance. % for competing and proposed models. 
%It can be seen that 
%both ``Cos-Sim'' models outperform all VAE-based methods, however, it should be noted that Cos-Sim methods are not able to discover any latent factors, as was evident during empirical traversals of their embedded spaces. 
%In fact, only RIVAE was able to discover meaningful hidden factors. Note that Recipe1M has no information pertaining to any factors.
% it only provides recipes as text and images, 
% therefore, 
The Cos-Sim-LVM attains the best retrieval performance even with the introduced bottleneck layers, %to capture the latent factors, 
where our RIVAE performs nearly comparably to it. Note that both DCCA~\cite{dcca} and  Bi-VAE~\cite{wgvae} completely failed to converge. % as both are intractable for large complex datasets. 

%%%%%%%%%%%%%%
\subsubsection{Quantitative Analysis for Recipe1M}

In the Recipe1M dataset, there are no labels available for the ground-truth factors, those that commonly govern the variability of recipes and food images, deemed to be {\em food factors}. To measure the goodness of the learned latent variables of the competing methods, we aim to select a small subset of food factors that look the most pronounced and capturing the shared variability in recipes and images. To this end, we re-crape all the recipes in the Recipe1M that are associated with the Internet domain \url{food.com}, and parse their keywords and categories. Then we form the recipe {\em tags} as the unique terms union between keywords and categories. This subset represents about $50\%$ of the whole dataset.

%we collect the food tags/annotations from the nutrient DBs publicly available. There were ??? many tags, and they are hierarchical in nature, eg, dish type $=$ dessert, ... \hl{Elaborate more about how the food tags are gathered...}

Then we manually group the tags that are related to one another, and among the tag groups, we choose 8 factors the most dependent on the latent variables by visually inspecting the latent traversal results for the competing methods. 
They are: 1) wateriness, 2) greenness, 3) stickiness, 4) oven-baked-or-not, 5) food container longishness (e.g., bottle/cup or plate), 6) grains, 7) savory-or-dessert, and 8) fruit-or-no-fruit. They are intuitively very appealing. 
%For "greenness", we can say it's a factor related to color, but also obvious from text/recipe. So included.
For the association between the tags and these 8 factors, refer to the Supplement. 
All these factors that consider in this analysis are ordinal, and we consider 5 scales/levels for each factor (e.g., wateriness$=1$ means very dry food, while wateriness$=5$ implies containing lots of water). 

We sample about 20 random reference/query images %that serve as queries 
(${\bf x}_1^\textrm{Ref}$) in the traversal/retrieval. %For each reference, we find the (30-dim) latent vector, and
For the 100 traversal points along each latent dimension, we collect retrieved items (${\bf x}_2)$, and manually label the values of the 8 food factors. Then we select 10 latent dimensions that have the highest correlations with the 8 factors, and form a $(10 \times 8)$  correlation table. The D/C/I measures are %then evaluated as 
summarized in Table~\ref{tab:recipe1m_quant}. As shown, our RIVAE attains the highest scores among the competing models by large margin.

%%%%
\begin{table}%[t!]
%\vspace{-2.0em}
\centering
\caption{(R1M) Goodness of the learned latents. %(averaged over 10 random runs).
}
\label{tab:recipe1m_quant}
%\footnotesize
\small
\vspace{-1.0em}
\begin{tabular}{|c||c|c|c|}
\hline
 & {Disent. $\uparrow$} & {Comple. $\uparrow$} & {Inform. $\uparrow$} \\
\hline\hline
{Cos-Sim-LVM} & 0.6203 & 0.6027 & 0.5983 \\ \hline
{Bi-VAE} & Failed & Failed & Failed \\ \hline
{Bi-VAE on $\mathcal{V}$} & 0.5128 & 0.5423 & 0.5670 \\
\hline
{DCCA} & Failed & Failed & Failed \\ \hline
{RBi-VAE} & 0.4954 & 0.6079 & 0.5664 \\ 
\hline
{RIVAE} & ${\bf 0.8569}$ & ${\bf 0.8500}$ & ${\bf 0.8615}$ \\
\hline
\end{tabular}
%%
% \vskip -0.1in
%\vspace{-0.8em}
\end{table}
%%%%

%%%%%%%%%%%%%%
\subsubsection{Qualitative (Visual) Analysis for Recipe1M}

\begin{figure}%[t!]
%\vspace{-1.0em}
\begin{center}
% % \includegraphics[width=1.0\linewidth]{figs/R1M/word-clouds/dimension2_word-clouds.png} \\\\
% \includegraphics[height=2.0cm]{figs/R1M/traversals/dim0/dimension-0_anchor-0_traversal.jpg} \hspace{0.15cm}
% \includegraphics[height=2.0cm]{figs/R1M/traversals/dim1/dimension-1_anchor-7_traversal.jpg} \\ \\
% \includegraphics[height=2.0cm]{figs/R1M/traversals/dim2/dimension-2_anchor-19_traversal.jpg}  \hspace{0.15cm}
% \includegraphics[height=2.0cm]{figs/R1M/traversals/dim3/dimension-3_anchor-18_traversal.jpg} 
\includegraphics[width=1.0\linewidth]{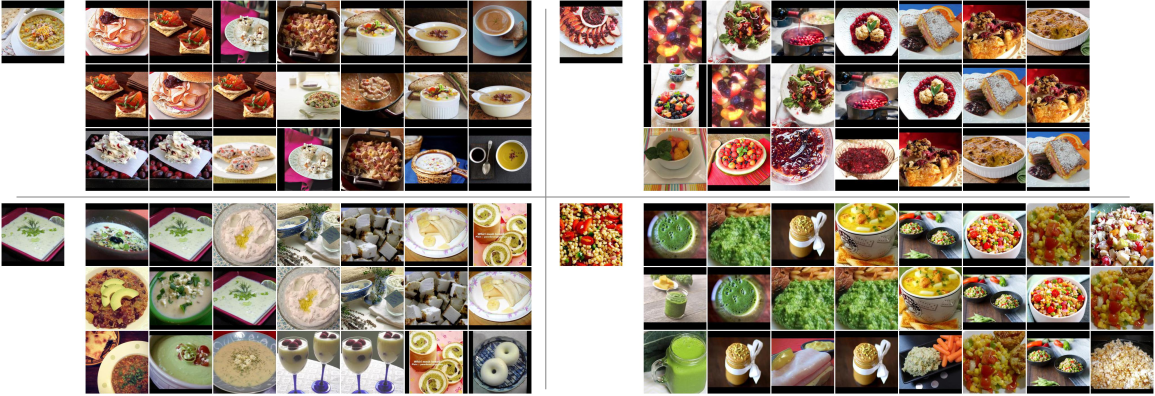} 
\end{center}
\vspace{-1.8em}
\caption{Retrieved images from latent traversal. \textbf{Top left} latent variable for {\em wateriness},
%$z_0$ (non watery - watery)
\textbf{top right} {\em fruit-no-fruit}, 
%$z_1$ (non noodle - noodle), 
\textbf{bottom left} {\em savoriness}, %$z_2$ (savory - sweet) 
and \textbf{bottom right} {\em greenness}. %$z_3$ (green - non green). 
%\textbf{Image grids}: 
For each panel, the query image is shown on the leftmost, and each column has top-3 retrieved items at each traversal point. % in traversal $[\mu_i-10\sigma_i,\mu_i+10\sigma_i]$. %Note that NNs are only shown for points with unique top-1, hence, there are different amounts of retrieved images depending on the query. 
Larger images, more examples, and other discovered factors can be found in Supplement. 
}
\label{fig:traversal}
%\vspace{-1.5em}
\end{figure}

Next we qualitatively assess the discovered hidden factors through visual inspection of the retrieved items from latent traversal. We also generate the word cloud images using the ingredients in the retrieved recipes. 
For the four latent variables that have the highest correlation with the factors: wateriness, greenness, savoriness, and fruit-or-no-fruit, 
%. See supplementary material for a complete list of all dimensions' interpretation.  
% 
% Figs.~\ref{fig:traversal2} and \ref{fig:wordclouds2} illustrate some of the results possible with RIVAE.
% 
Fig.~\ref{fig:traversal} shows traversal results for top-3 retrieved items, which are visually very coherent to the corresponding true factors.  %over $[\mu_i-10\sigma_i,\mu_i+10\sigma_i]$ for four query images in four dimensions $z_i \in \mathcal{Z}^{30}$ for RIVAE, where $\mu_i=\mathbb{E}_{{\bf x}_1\sim P_d}[P(z_i|{\bf v}_1)]$ is the mean of the aggregated conditional prior, and $\sigma_i$ is standard deviation defined similarly. These dimensions correspond to {\em wateriness}, {\em noodles}, {\em savoriness}, and {\em greenness}, assessed qualitatively. See Supplement for more retrieved examples and other factors discovered. 
%hidden factors that qualitative assessment indicates they are related to wateriness, savoriness, noodles and greenness.

%%%%%%%%%%%%%%%%%%%%%%%%%%%%%%%%%%%%%%%%%%%%%%%%%%%%%%%%
%%%%%%%%%%%%%%%%%%%%%%%%%%%%%%%%%%%%%%%%%%%%%%%%%%%%%%%%
%%%%%%%%%%%%%%%%%%%%%%%%%%%%%%%%%%%%%%%%%%%%%%%%%%%%%%%%
\begin{figure}%[t!]
%\vspace{-1.5em}
\begin{center}
\includegraphics[width=0.95\linewidth]{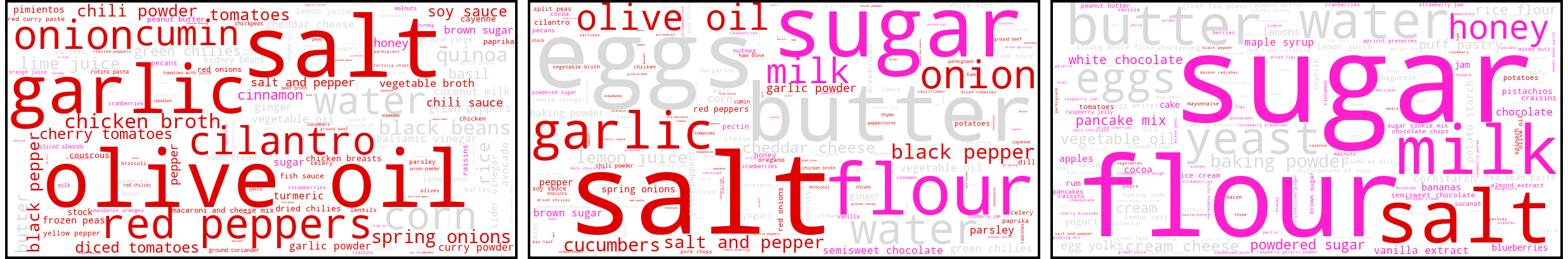} 
\end{center}
\vspace{-1.8em}
\caption{Ingredient word clouds for the latent variable {\em savoriness}, %$z_2$ (factor of ``savory - sweet'') 
obtained from top-10 retrieved items over 20 queries. % at $\mu_2-10\sigma_2$ (left), $\mu_2$ (middle), and $\mu_2+10\sigma_2$ (right). 
Ingredient color indicates typical use; red $=$ savory and pink $=$ non-savory (sweet).}
\label{fig:wordcloud}
%\vspace{-1.5em}
\end{figure}
%%%%%%%%%%%%%%%%%%%%%%%%%%%%%%%%%%%%%%%%%%%%%%%%%%%%%%%%

\begin{figure}%[t!]
%\vspace{-1.0em}
\begin{center}
\includegraphics[width=0.95\linewidth]{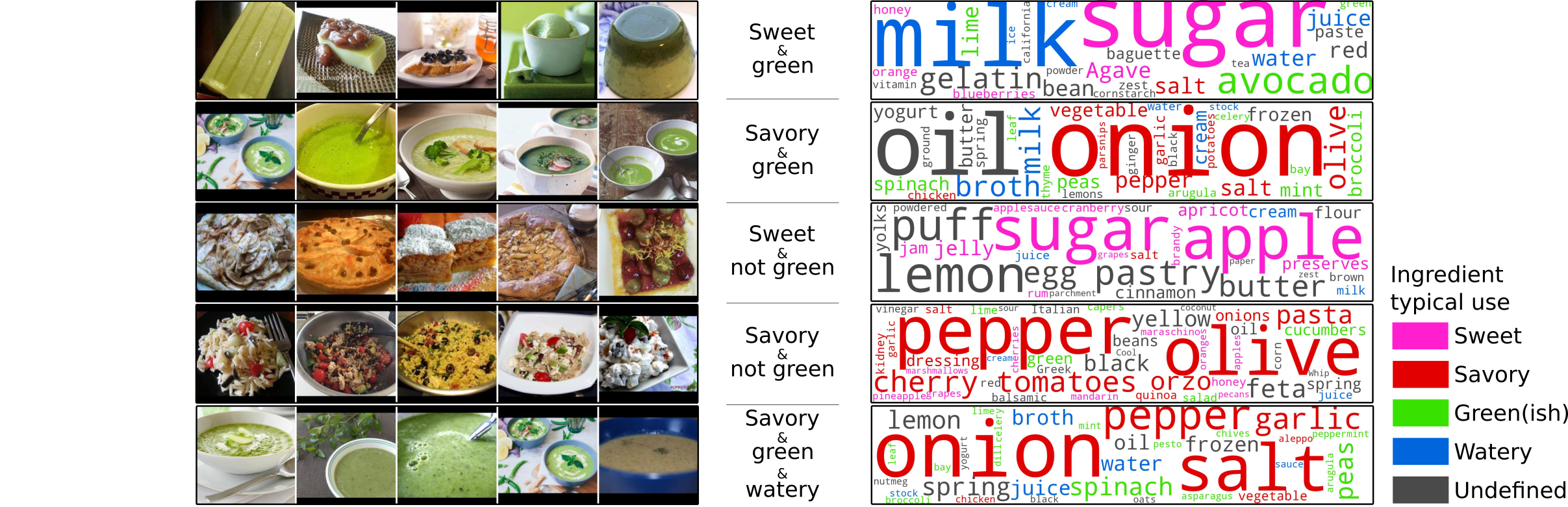}
\end{center}
\vspace{-1.8em}
\caption{\textbf{Left}: Retrieved images from activation of multiple latent variables. %Each row corresponds to retrieved items after assigning extreme  values for the $z_i$'s (indicated in the \textbf{Middle}). %$z_i=\mu_i \pm 10\sigma$. 
% such that $z_i$ activate the factors mentioned in the caption.
\textbf{Middle}: indicates which combinations of latent variables %(savory, green, watery) 
are activated. 
\textbf{Right}: Ingredient word clouds of the retrieved recipes. 
% Ingredient color indicates typical use, with red-savory, pink-sweet, blue-watery and green-green.
% Notice that in the bottom left set of images the fifth seem "brownish", however, this image corresponds to a broccoli soup, which is generally green.
}
\label{fig:traversal2}
%\vspace{-1.5em}
\end{figure}

In Fig.~\ref{fig:wordcloud}, we show the ingredient word clouds generated from the top-10 retrieved items 
%nearest neighbors (NN) 
from 20 query data points 
for the latent variable corresponding to the {\em savoriness}. % where we assign $\mu_2-10\sigma_2$, $\mu_2$, and $\mu_2+10\sigma_2$ to the latent variable $z_2$ (factor of ``savory - sweet''). 
We see that the word cloud on the left end % ($z_2=\mu_2-10\sigma_2$) 
contains ingredients mostly associated with savory dishes, %and similar interpretations for others. 
the middle contains both savory and sweet ingredients, while the right end has typical dessert ingredients. 
This highlights the main advantage of our model where it accurately identifies true factors, and allows us to directly control the latent variables to generate (retrieve) desired data items. This feature can also be easily extended to {\em manipulation of multiple factors}. %For instance, we control two or three latent variables simultaneously, to obtain the combined effects in the retrieval results.  %Due to the lack of space, we leave the results in the Supplement. 
Fig.~\ref{fig:traversal2} %(left) 
shows top-5 retrieved examples when we %turn on/off 
control two or three latent variables corresponding to {\em savoriness}, {\em wateriness}, and {\em greenness}, simultaneously. 
% $z_2$ (savory), $z_3$ (green), and $z_0$ (watery), simultaneously. 
%after translating a query point along two or even three dimensions, $z_2$ (savory), $z_3$ (green), and $z_0$ (watery). 
%Here, it can be seen that the translation 
%This activation in the latent space successfully transforms the  manipulated latent factors to a matching recipe. % that matches the selected factors.  %Fig.~\ref{fig:traversal2} (right) shows word clouds associated to the set of images retrieved in Fig.~\ref{fig:traversal2} (left). It can be observed that ingredients can explain the desired latent factors.

%%%%%%%%%%%%%%%%%%%%%%%%%%%%%%%%%%%%%%%%%%%%%%%%%%%%%%%%
%%%%%%%%%%%%%%%%%%%%%%%%%%%%%%%%%%%%%%%%%%%%%%%%%%%%%%%%

%%%%%%%%%%%%%%%%%%%%%%%%%%%%%%%%%%%%%%%%%%%%%%%%%%%%%%%%%%%%%%%%%%%%%%%%%%%%%%%
\subsection{Ablation Study: %Impact of 
Embedder Regularization}\label{sec:ablation}

In our Retrieval-IVAE model, we employed the embedder regularization loss, specifically (\ref{eq:emb_reg}). To verify the impact of this regularization term, we run our RIVAE model without this loss term, and the retrieval results are summarized in Table~\ref{tab:ablation}, in conjunction with the results with the embedder regularization.
The result demonstrates that without embedding network regularization, either the performance becomes considerably degraded %(Synth and Split-MNIST), 
or %the model 
simply fails. % to work (Sprites).

%%%%
\begin{table}%[t!]
%\vspace{-1.0em}
\centering
\caption{%Retrieval performance for our 
RIVAE with/without embedder regularization. %(\ref{eq:emb_reg}).
}
\label{tab:ablation}
%\footnotesize
\small
\vspace{-1.0em}
\begin{tabular}{|c||c||c|c|c|c|}
\hline
Dataset & Reg. & $R@1\uparrow$ & $R@5\uparrow$ & $R@10\uparrow$ & Med-R $\downarrow$ \\
\hline\hline
%RBi-VAE ($\gamma=0.0$) & 0.41 & 0.86 & 0.96 & 2.00 \\
%\hline
%RBi-VAE ($\gamma=0.0$) & 0.35 & 0.85 & 0.95 & 2.00 \\
%\hline
\multirow{2}{*}{Synth} & Yes & 0.35 & 0.80 & 0.92 & 2.00 \\
\cline{2-6}
 & No & 0.29 & 0.73 & 0.90 & 3.00 \\
\hline\hline
\multirow{2}{*}{Sprites} & Yes & 1.00 & 1.00 & 1.00 & 1.00 \\
\cline{2-6}
 & No & 0.01 & 0.03 & 0.06 & 144.40 \\
\hline\hline
%
%\multirow{2}{*}{Split-MNIST} 
Split- & Yes & 0.52 & 0.89 & 0.96 & 1.00 \\
\cline{2-6}
MNIST & No & 0.35 & 0.75 & 0.88 & 2.00 \\
\hline
\end{tabular}
%%
% \vskip -0.1in
%\vspace{-0.8em}\centering
\end{table}
%%%%

%%%%%%%%%%%%%%%%%%%%%%%%%%%%%%%%%%%%%%%%%%%%%%%%%%%%%%%%%%%%%%%%%%%%%%%%%%%%%%%
%%%%%%%%%%%%%%%%%%%%%%%%%%%%%%%%%%%%%%%%%%%%%%%%%%%%%%%%%%%%%%%%%%%%%%%%%%%%%%%
\section{Conclusion}

%In this paper 
We have proposed a novel Retrieval-IVAE model, extension of the identifiable VAE for cross-modal retrieval, where it completely removes the ambient data decoding module via implicit encoder inversion that is achieved by Jacobian regularization of the low-dimensional embedding function. 
The model is shown to have the capability of learning disentangled and complete latent representations that can explain the variability of the bi-modal data,  
%Our RIVAE is shown to identify true  factors more accurately than existing approaches both 
on both controlled and the large-scale Recipe1M datasets.  As our future work, we plan to apply our approaches to retrieval tasks with more complex structured data in multiple modalities greater than two, including video, text, and audio data. %, and other human sensing data. 

% \newpage 

% {\small
% \bibliographystyle{ieee_fullname}
% \bibliography{main}
% }

{\small
\bibliographystyle{ieee_fullname}
\bibliography{main}
}

\newpage

\appendix

\begin{center}
%\vspace*{\fill}
\LARGE Supplementary Material
%\vspace*{\fill}
\end{center}This supplement consists of the following materials: 
%%%%
\begin{itemize}
\item Experimental details (Sec.~\ref{sec:supp_model_optim}).
\item Additional experimental results (Sec.~\ref{sec:supp_extra_expmts}).
    \begin{itemize}
    \item Sprites: Plots of true factors vs.~latent variables (Sec.~\ref{sec:supp_expmt_sprites4})
    \item Split-MNIST: Visual results of latent traversal (Sec.~\ref{sec:supp_expmt_split_mnist})
    \item Receipe1M: Visual results of latent traversal  (Sec.~\ref{sec:supp_expmt_im2recipe}) %: More qualitative results
    \end{itemize}
\item Recipe1M: Association between food factors and tags from \url{food.com} (Sec.~\ref{sec:supp_food_tag_assoc}).
\item Details of the joint model,  Retrieval-Bi-VAE (RBi-VAE) (Sec.~\ref{sec:supp_rbi_vae}).
\end{itemize}
%%%%

%%%%%%%%%%%%%%%%%%%%%%%%%%%%%%%%%%%%%%%%%%%%%%%%%%%%%%%%%%%%%%%%%%%%%%%%%%%%%%%
%%%%%%%%%%%%%%%%%%%%%%%%%%%%%%%%%%%%%%%%%%%%%%%%%%%%%%%%%%%%%%%%%%%%%%%%%%%%%%%
\section{Experimental Details}\label{sec:supp_model_optim}

%%%%%%%%%%%%%%%%%%%%%%%%%%%%%%%%%%%%%%%%%%%%%%%%%%%%%%%%%%%%%%%%%%%%%%%%%%%%%%%
\subsection{Network Architectures} 

%\textbf{Embedding networks}. 
\subsubsection{Embedding networks}
\begin{figure}[t!]
%\vspace{-0.2em}
\begin{center}
\raisebox{-.5\height}{\includegraphics[width=0.95\linewidth]{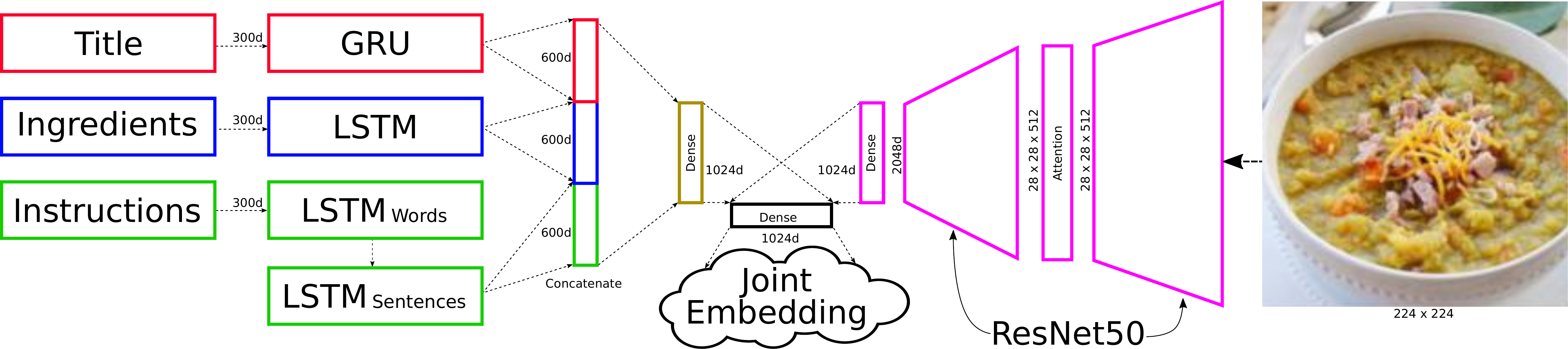}}
\end{center}
\vspace{-1.3em}
\caption{(Recipe1M) Diagram of the embedding networks.}
\label{fig:supp_embedder_diagram}
%\vspace{-1.5em}
\end{figure}

For all competing models that employ the embedding networks ${\bf e}_1(\cdot)$ and ${\bf e}_2(\cdot)$, we use the same network architectures. They are Cos-Sim-LVM, Bi-VAE-on-$\mathcal{V}$, DCCA, RBi-VAE, and our RIVAE. The learned embedding network parameters of the Cos-Sim-LVM are used as initial parameters for the rest models.  Detailed embedding network architectures for different benchmark datasets are defined as follows:
%%%%
\begin{itemize}
    \item \textbf{Synth}: A fully connected network with one hidden layer  of 25 hidden units. The $tanh()$ nonlinear link is used. 
    \item \textbf{Sprites}: Four 2D convolutional layers (filter size $(4 \times 4)$ pixels) with the number of channels 32, 32, 64, 64, followed by two fully connected layers with output dimensions 128 and $\textrm{dim}(\mathcal{V})=10$. The rectified linear unit (ReLU) is used as nonlinear components. 
    \item \textbf{Split-MNIST}: Four fully connected layers of hidden dimension 1024 are applied with the ReLU nonlinearity. 
    \item \textbf{Recipe1M}: 
    % We follow the same model architectures as~\cite{salvador2017learning,marin2019learning}.
    Our model combines architectural and training strategies from \cite{salvador2017learning,wang2019learning,chen2018deep}, mainly, hard-example mining, a modality-shared fully connected layer, and full RNN text encoder. Word embeddings used by our model are initialized from a recipe1M pre-trained word2vec 300 dimensional model, fine-tuned during training. 
    RNN layers are all bi-directional and have a 300-dimensional hidden state, with forward and backward states concatenated into a single 600-dimensional vector.
    Additionally, similar to other works, our image embedding module is based on ResNet50 pre-trained on ImageNet, however, we have introduced an attention module at the 28x28 level. This attention module helps the image encoder focus on the parts of the image that are related to the dish of food as described by the text recipe. Fig.~\ref{fig:supp_embedder_diagram} gives an schematic of the general architecture of the embedding modules.
    % \hl{More details??...}
\end{itemize}
%%%%
As usual practice, we also apply the $L_2$ normalization to the final outputs of the embedding networks to make the embedded vectors ${\bf v}$ unit norm. 

%\textbf{Model-Specific Network Architectures}. 
\subsubsection{Model-Specific Network Architectures}
\vspace{-0.7em}
For each competing model, we employ the following network architectures (and strategies):
%%%%
\begin{itemize}
    \item \textbf{RIVAE}: For the two conditional models $P({\bf z}|{\bf v}_1)$ and $P({\bf v}_2|{\bf z})$, we use fully connected networks with two hidden layers of 10 hidden units, and the leaky-ReLU with slope 0.2 is used for nonlinearity. For the Recipe1M, we used a single hidden layer with 100 hidden units, and the same leaky-ReLU nonlinearity as before.
    \item \textbf{RBi-VAE}: (See Sec.~\ref{sec:supp_rbi_vae} for the details.) For the bi-modal VAE model on the embedded spaces $\mathcal{V}$, we run several different model architectures, either linear or nonlinear, ranging from simpler to more complex architectures for the latter, and we report the ones with the highest performance on the validation sets. 
    % For the nonlinear networks, we use fully connected layers for both encoder and decoder networks with one or three hidden layers.
    For the nonlinear networks, in both encoder and decoder we use one or three fully connected hidden layers.
    %The nonlinear network architectures %include: the convolutional/transposed-convolutional networks of similar structures as the embedding networks above, and fully connected networks with one or three hidden layers.
    Note that %for our DFR-VAE, 
    when we adopt a linear (Gaussian) model in the place of the VAE, all inferences can be done analytically without introducing variational densities. For all datasets, the linear Gaussian models perform better than the nonlinear VAE models. 
    For the discriminator networks, used for estimating the total correlation loss via the density ratio trick, we adopted 6 fully connected layers with hidden dimension 300. For the Recipe1M, we also used 6 fully connected layers, however, with only 100 hidden dimensions.
    \item \textbf{Bi-VAE}: We need to model the ambient encoder $P({\bf z}|{\bf x}_i)$ and decoder networks $P({\bf x}_i|{\bf z})$ for $i=1,2$. Similarly as RBi-VAE, we run different (linear/nonlinear and simple/complex) network architectures, and select the best performing one. The nonlinear network architectures include: the convolutional and transposed-convolutional networks of similar structures as the embedding networks above, and the fully connected networks with one or three hidden layers.
    \item \textbf{Bi-VAE-on-$\mathcal{V}$}: This model fixes the embedding networks and learns only the bi-modal VAE model built on the embedded spaces. Hence, the overall model architectures are identical to the RBi-VAE. 
    \item \textbf{DCCA}: Instead of building the nonlinear feature extraction networks ${\bf z} = \bm{\phi}({\bf x})$ from the scratch, we compose the embedding networks ${\bf v} = {\bf e}({\bf x})$ with the additional nonlinear mappings ${\bf z} = {\bf g}({\bf v})$ to form $\bm{\phi}$ (i.e., $\bm{\phi} = {\bf g} \circ {\bf e}$). The network ${\bf g}$ is defined as fully connected networks with three hidden layers with hidden dimension 10, hence fairly comparable to RBi-VAE. For the Recipe1M, similarly to our RIVAE, we used a single hidden layer with 100 hidden units as the network ${\bf g}$.
\end{itemize}
%%%%

%%%%%%%%%%%%%%%%%%%%%%%%%%%%%%%%%%%%%%%%%%%%%%%%%%%%%%%%%%%%%%%%%%%%%%%%%%%%%%%
\subsection{Optimization Strategies}

We use the Adam optimizer~\cite{adam} for training of all models. The batch size is 64 and the maximum number of epochs is 2000. For our RIVAE, at the first stage we train the IVAE module alone with the embedding networks fixed, then at the second stage we jointly train the entire networks. The joint training starts at epoch 100 except for the Split-MNIST dataset (epoch 5). The learning rates are: 0.005 (Synth), 0.001 (Sprites), and 0.0001 (Split-MNIST). We also decay the learning rates by roughly half at epoch 200 and 1000 (at epoch 20 and 50 for the Split-MNIST). 
% \hl{Recipe1M optimization hyperparameters??}
The trade-off parameter $\lambda$ for the embedding regularization  $\textrm{Reg}({\bf e}_2)$ is fixed as $0.1$ throughout the experiments. 
For the Recipe1M the first stage (IVAE only) was trained for 20 epochs using a learning rate of 0.001. The second stage (joint embedding and IVAE) observes annealing of the learning by a factor of 0.1 at epochs 30 and 60.

%%%%%%%%%%%%%%%%%%%%%%%%%%%%%%%%%%%%%%%%%%%%%%%%%%%%%%%%%%%%%%%%%%%%%%%%
%%%%%%%%%%%%%%%%%%%%%%%%%%%%%%%%%%%%%%%%%%%%%%%%%%%%%%%%%%%%%%%%%%%%%%%%
\section{Additional Experimental Results}\label{sec:supp_extra_expmts}

%%%%%%%%%%%%%%%%%%%%%%%%%%%%%%%%%%%%%%%%%%%%%%%%%%%%%%%%%%%%%%%%%%%%%%%%%%%%%%%
\subsection{Sprites: Plots of true factors vs.~latent variables
}\label{sec:supp_expmt_sprites4}

\begin{figure}%[t!]
  \centering
  \begin{subfigure}{0.321\textwidth}
    \centering
    \includegraphics[trim = 1mm 3mm 1mm 7mm, clip, scale=0.225]{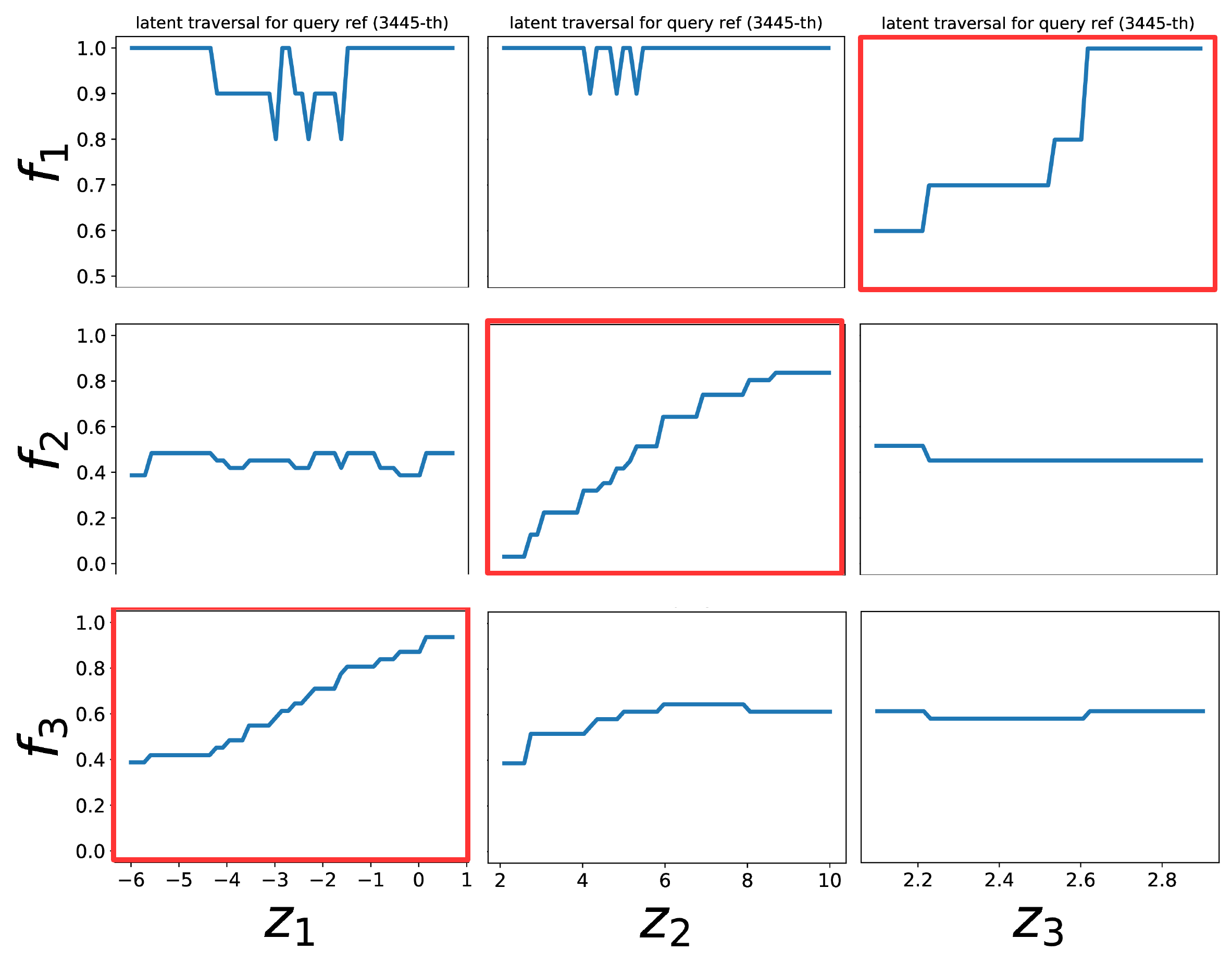}\caption{RIVAE}
    \label{fig:supp_trv_sprites_dfrivae}
  \end{subfigure}
  \ \
  \begin{subfigure}{0.321\textwidth}
    \centering
    \includegraphics[trim = 1mm 3mm 1mm 7mm, clip, scale=0.225]{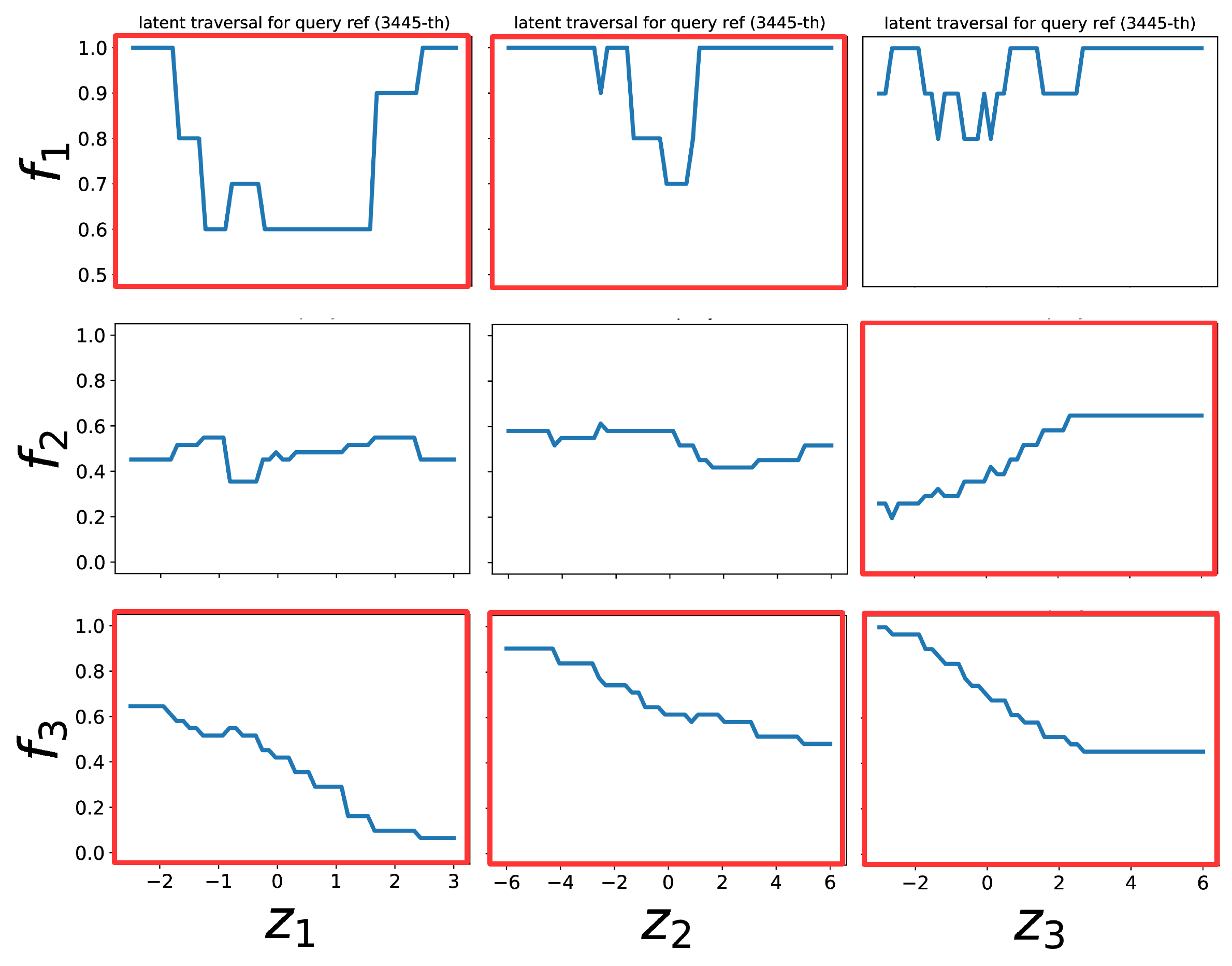}\caption{RBi-VAE %($\gamma=10.0$)
    }
    \label{fig:supp_trv_sprites_dfrvae10}
  \end{subfigure}
  \ \
  \begin{subfigure}{0.321\textwidth}
    \centering
    \includegraphics[trim = 1mm 3mm 1mm 7mm, clip, scale=0.225]{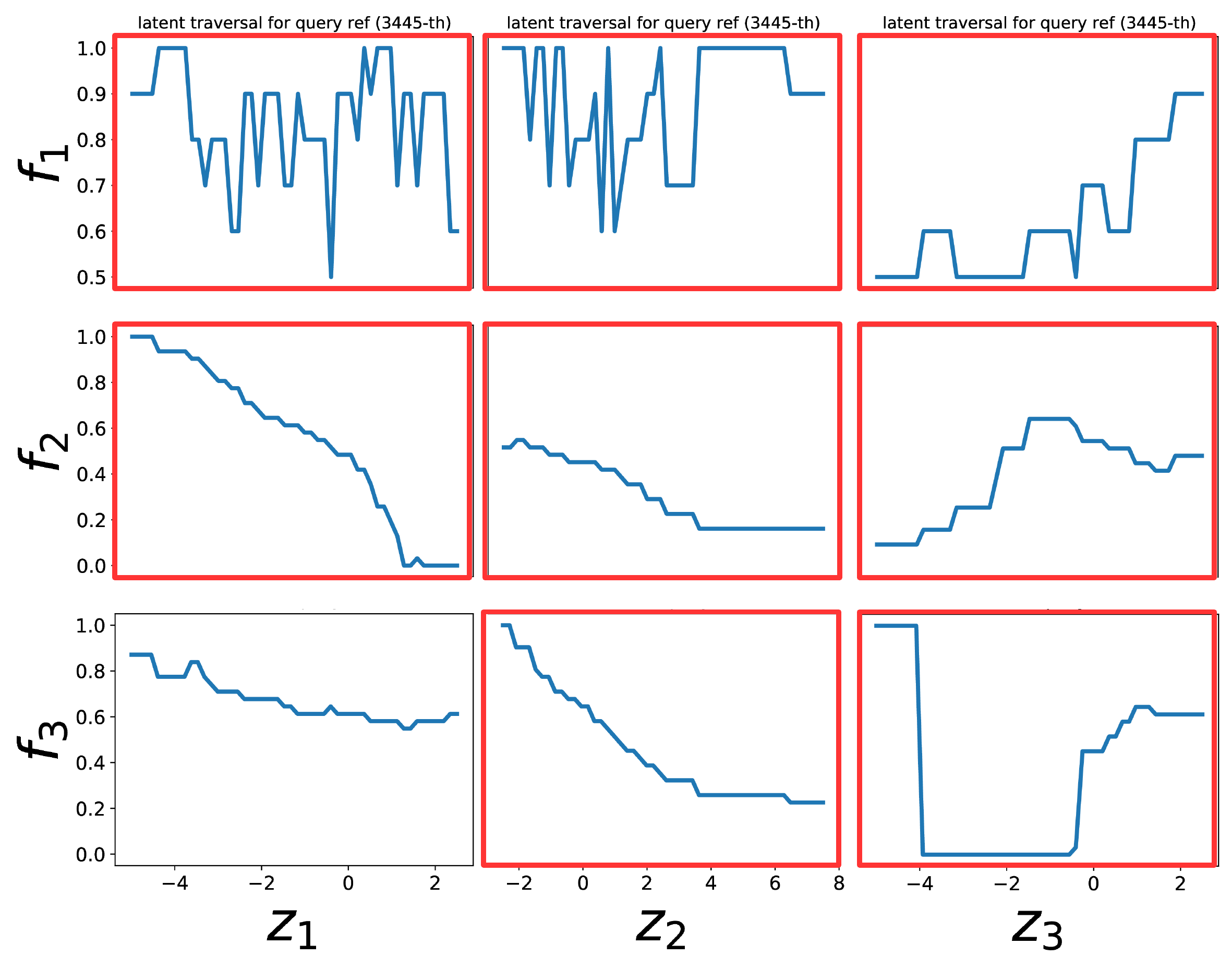}\caption{DCCA}
    \label{fig:supp_trv_sprites_dcca}
  \end{subfigure}
\vspace{-0.6em}
\caption{(Sprites) True factors ($f_1=$ scale, $f_2=$ X-pos, and $f_3=$ Y-pos) of the retrieved images vs.~latent traversed points. %\textbf{Top-Left}: DFR-IVAE. \textbf{Top-Right}: DFR-VAE ($\gamma=10.0$). \textbf{Bottom-Left}: DFR-VAE ($\gamma=50.0$). \textbf{Bottom-Right}: DeepCCA~\cite{dcca}. 
%Similar interpretations as Fig.~\ref{fig:supp_synth_trv}. 
%For instance, bottom left panel shows how the retrieved image's Y-pos ($f_3)$ changes due to the change in $z_1$. Since the other two panels above it (scale $f_1$ vs.~$z_1$ and X-pos $f_2$ vs.~$z_1$) almost remain unchanged, we can conclude that 
Each column shows traversal of one latent $z_j$ with the other fixed. % and the rows show the individual 
The Y axes are true factors %(from top to bottom, $f^S_1$, $f^S_2$, and $f^2$), 
obtained from retrieved items ${\bf x}_2$. Red boxes indicate %are marked if the plots have 
significant changes in the true factors, due to latent traversal, i.e., high correlation. 
}
\label{fig:supp_sprites4_f_vs_z}
\vspace{-0.5em}
\end{figure}

Recall that we assumed that the shape from sprites induces the modalities, specifically, ${\bf x}_1$ is a {\em square} image, and ${\bf x}_2$ an {\em oval}, and considered only the X, Y positions and the scale of the sprite as the underlying shared factors, with the other varying source {\em rotation} being fixed.
The plots of the ground-truth factors of the retrieved images due to the latent space traversal are shown in Fig.~\ref{fig:supp_sprites4_f_vs_z}. Our RIVAE shows near one-to-one correspondence between the true and learned factors, while other models exhibit considerable entanglement. More specifically, for RIVAE, we see that $z_1$ {\em exclusively} corresponds to $f_3$ (highlighted by red box), $z_2=f_2$ and $z_3=f_1$, signifying that our model learns the disentangled latent representation very accurately.  
On the other hand, for the DCCA, there is no exclusive correspondence, but significant entanglement. E.g., change in $z_1$ results in considerable changes in both $f_1$ and $f_2$, indicating that the scale and X-pos factors are entangled in the latent variable $z_1$. 

%%%%
%%%%

%%%%%%%%%%%%%%%%%%%%%%%%%%%%%%%%%%%%%%%%%%%%%%%%%%%%%%%%%%%%%%%%%%%%%%%%%%%%%%%
\subsection{Split-MNIST: Visual results of latent traversal}\label{sec:supp_expmt_split_mnist}

\begin{figure}[t!]
%\vspace{-1.0em}
\begin{center}
\includegraphics[trim = 0mm 0mm 0mm 0mm, clip, scale=0.485]{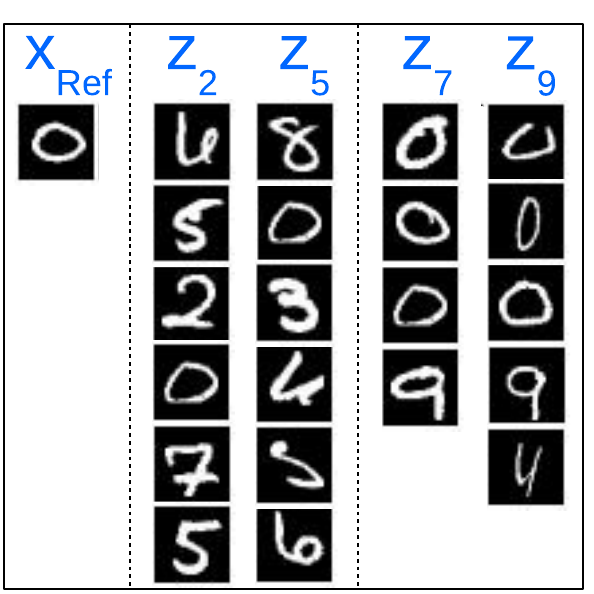} \ \
\includegraphics[trim = 0mm 0mm 0mm 0mm, clip, scale=0.485]{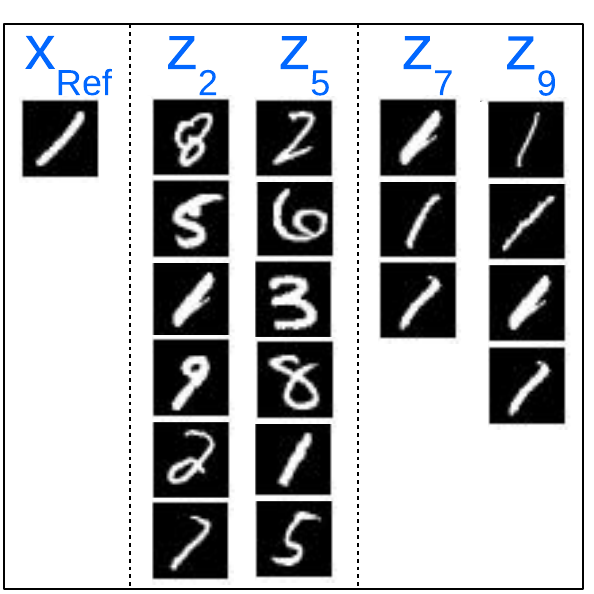} \ \
\includegraphics[trim = 0mm 0mm 0mm 0mm, clip, scale=0.485]{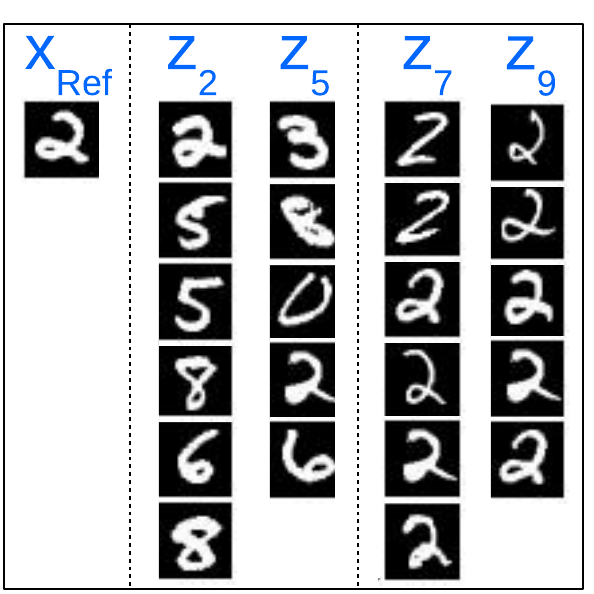} \ \
\includegraphics[trim = 0mm 0mm 0mm 0mm, clip, scale=0.485]{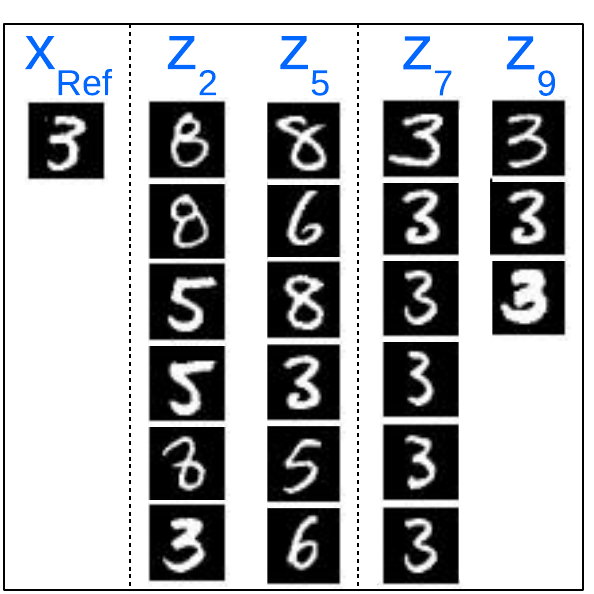} \ \
\includegraphics[trim = 0mm 0mm 0mm 0mm, clip, scale=0.485]{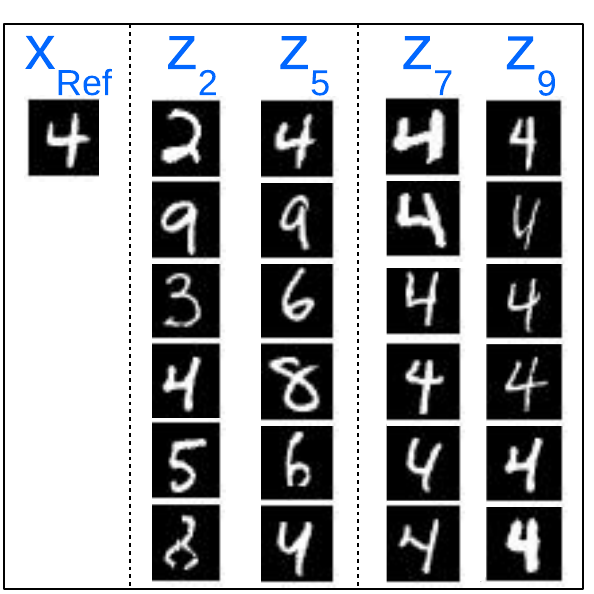} \\
\includegraphics[trim = 0mm 0mm 0mm 0mm, clip, scale=0.485]{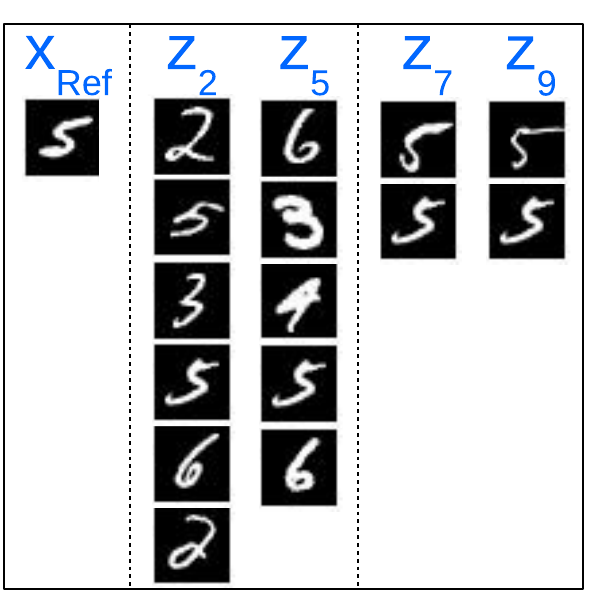} \ \
\includegraphics[trim = 0mm 0mm 0mm 0mm, clip, scale=0.485]{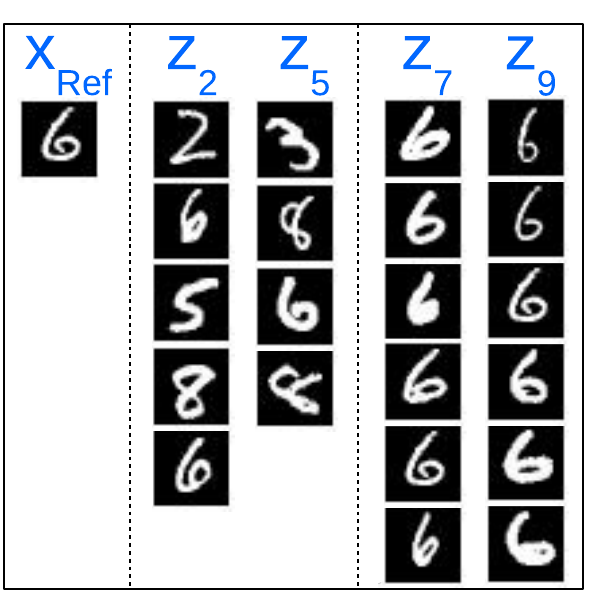} \ \
\includegraphics[trim = 0mm 0mm 0mm 0mm, clip, scale=0.485]{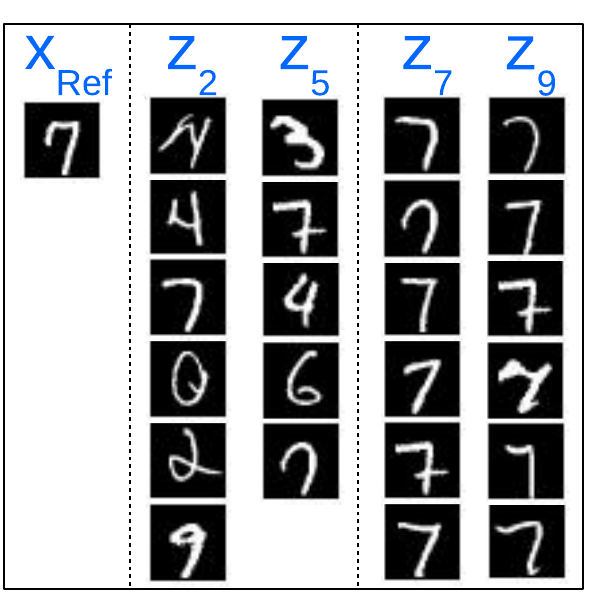} \ \
\includegraphics[trim = 0mm 0mm 0mm 0mm, clip, scale=0.485]{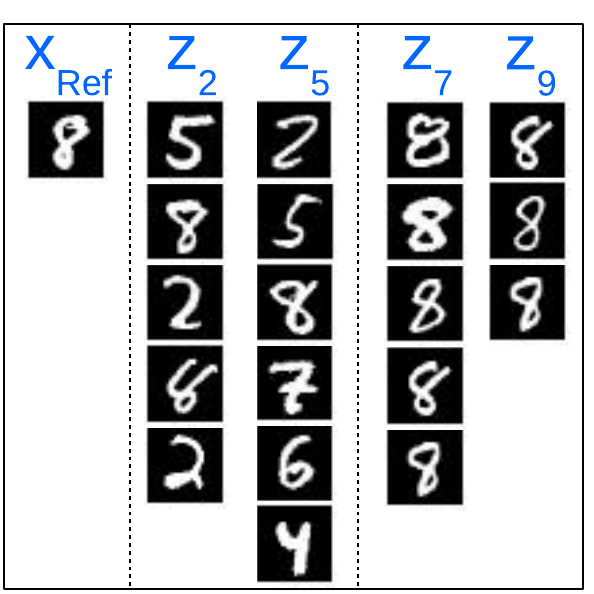} \ \
\includegraphics[trim = 0mm 0mm 0mm 0mm, clip, scale=0.485]{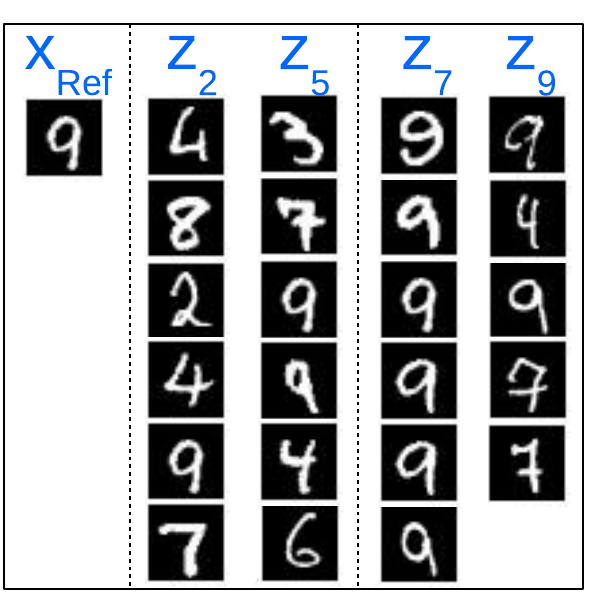}
\end{center}
\vspace{-1.5em}
\caption{(Split-MNIST) Latent traversal in our RIVAE model. Among $\textrm{dim}({\bf z})=10$ latent variables, we highlight four ($z_2$, $z_5$, $z_7$, $z_9$); for each we perform traversal by varying the latent values while fixing the rest variables. We depict 10 cases (panels), where each panel consists of: i) ${\bf x}_\textrm{Ref}=$ the anchor/reference input whose latent vector ${\bf z}$ is inferred using its left half ${\bf x}_1$, ii) each column ($z_j$) %next to ${\bf x}_\textrm{Ref}$ 
contains the retrieved images (the completed images are shown by merging the retrieved right half images ${\bf x}_2$ with their corresponding ground-truth left halves) obtained by varying $z_j$ while the rest variables being fixed. Visually, it is clear varying $z_2$/$z_5$ leads to dominant changes in digit class with writing style roughly remaining intact. Traversal along $z_7$/$z_9$ produces large variation in style with digit classes remaining mostly unchanged. This result shows that our RIVAE  accurately disentangles the two main factors, writing style and digit class, in the learned latent representation.
}
\label{fig:supp_split_mnist_trv}
\vspace{-0.5em}
\end{figure}

For this dataset, we formed a bi-modal setup by taking the left half of each image as modality-1 and the right half as modality-2. The shared factors for both modalities would be the digit class and the writing style, which are independent (disentangled) from each other. 
To see how the two underlying factors, writing style and digit class, are disentangled in the learned latent variables for our RIVAE, we show the latent traversal results in Fig.~\ref{fig:supp_split_mnist_trv}. For the four highlight latent variables ($z_2$, $z_5$, $z_7$, $z_9$), we can see that the first two explain the digit class variation, while the other two correspond to the writing style. Refer to the caption of the figure for details.

%%%%
%%%%

%%%%%%%%%%%%%%%%%%%%%%%%%%%%%%%%%%%%%%%%%%%%%%%%%%%%%%%%%%%%%%%%%%%%%%%%%%%%%%%%%%%%%%
\subsection{Receipe1M: Visual results of latent traversal}\label{sec:supp_expmt_im2recipe}

\begin{figure}[t!]
%\vspace{+1.5em}
  %
  \centering
  \begin{subfigure}{1.0\textwidth}
    \centering
    \includegraphics[width=1.0\linewidth]{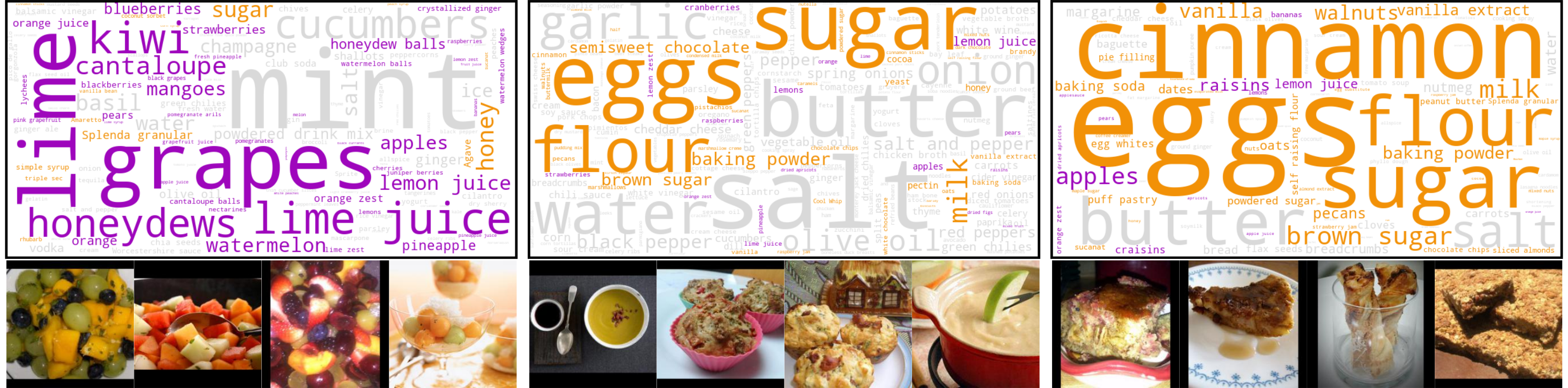}\vspace{-0.5em}\caption{$z_{23}$ ({\em fruit-no-fruit}): fruit ingredients as purple -- non-fruit ingredients as orange.}
    \label{fig:supp_wordcloud_z23}
  \end{subfigure}
  \\ \vspace{+0.3em}%
  \centering
  \begin{subfigure}{1.0\textwidth}
    \centering
    \includegraphics[width=1.0\linewidth]{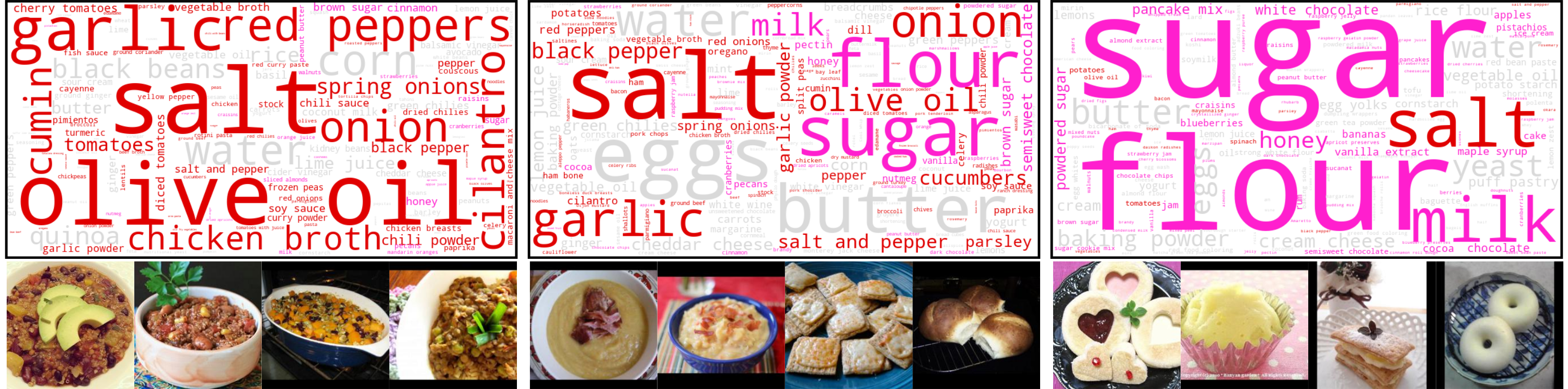}\vspace{-0.5em}\caption{$z_{2}$ ({\em savoriness}): savory ingredients as red -- non-savory as pink.}
    \label{fig:supp_wordcloud_z2}
  \end{subfigure}
  \\ \vspace{+0.3em}%
  \centering
  \begin{subfigure}{1.0\textwidth}
    \centering
    \includegraphics[width=1.0\linewidth]{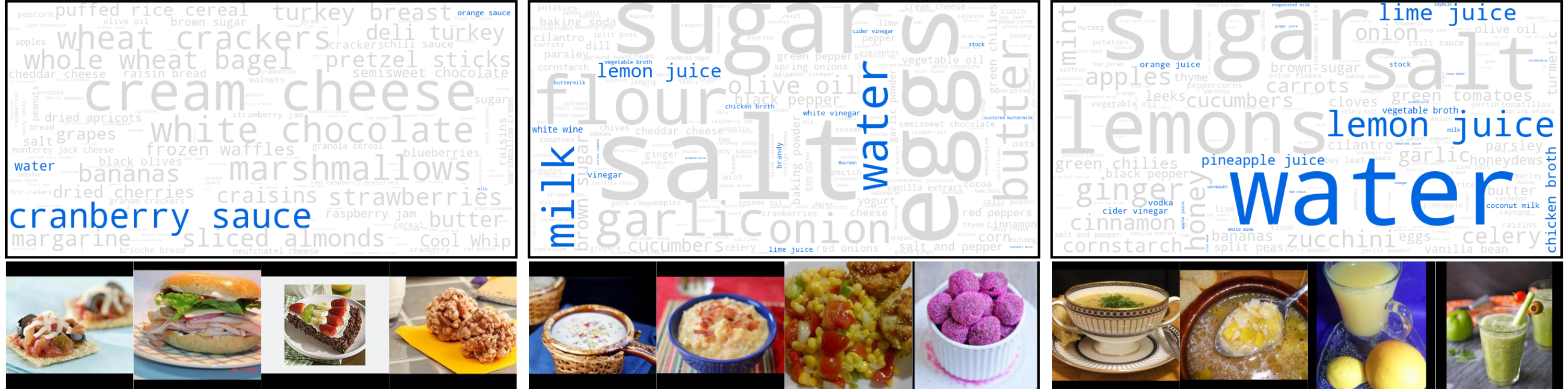}\vspace{-0.5em}\caption{$z_{0}$ ({\em wateriness}): watery ingredients as blue -- non-watery ingredients as gray.}
    \label{fig:supp_wordcloud_z0}
  \end{subfigure}
  \\ \vspace{+0.3em}%
  \centering
  \begin{subfigure}{1.0\textwidth}
    \centering
    \includegraphics[width=1.0\linewidth]{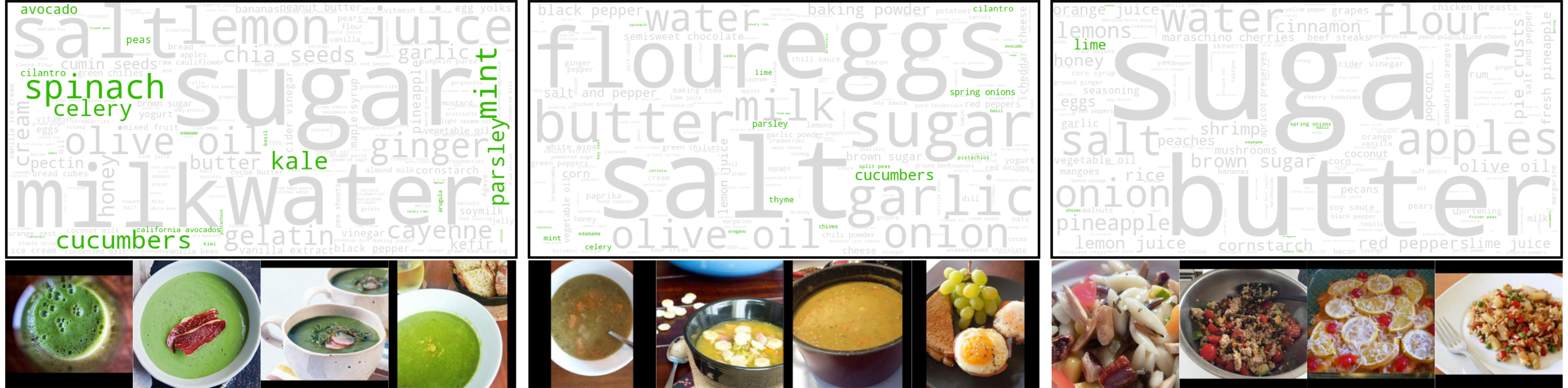}\vspace{-0.5em}\caption{$z_{3}$ ({\em greenness}): green ingredients as green -- non-green ingredients as gray.}
    \label{fig:supp_wordcloud_z3}
  \end{subfigure}
%   \\ \vspace{+0.3em}%
%   \centering
%   \begin{subfigure}{1.0\textwidth}
%     \centering
%     \includegraphics[width=1.0\linewidth]{figs/R1M/word-clouds/dimension1_word-clouds.png}\vspace{-0.5em}\caption{$z_{1}$ (strength$=0.35$): Non-noodle -- Noodle (brown)}
%     \label{fig:supp_wordcloud_z1}
%   \end{subfigure}
  %
\vspace{-0.8em}
\caption{(Recipe1M) Ingredient word clouds obtained from latent traversal (at three values, low/medium/high, in left/middle/right columns) along each of four latent variables.  %$z_{23}$, $z_2$, $z_0$, $z_3$, and $z_1$, in order of strength as in Table~\ref{tab:interpretations}. 
For each row, ingredients are colored differently according to the relevance/irrelevance to the governing factor (best seen in color). Below each word cloud image, we also place four retrieved food images corresponding to it. For instance, in (b), the factor of {\em savoriness}, we manually color the savory ingredients as red, while typical dessert (non-savory) ingredients are shown as pink. The light grey color is reserved for ingredients with no particular association to the factor in question. %The visual distinction between the two extreme cases $z_i = \mu_i \pm 10 \sigma_i$ is more pronounced for the stronger factors.  
%For example, the rightmost panel in (b) is largely pink, associating this factor value with ``sweet/baked dessert''.  On the other hand, the leftmost panel in (b) is largely red, indicating an association with ``savory''. 
}
\label{fig:supp_wordcloud}
%\vspace{-1.0em}
\end{figure}

\begin{figure}%[t!]
\raisebox{-.5\height}{\includegraphics[height=3.99cm]{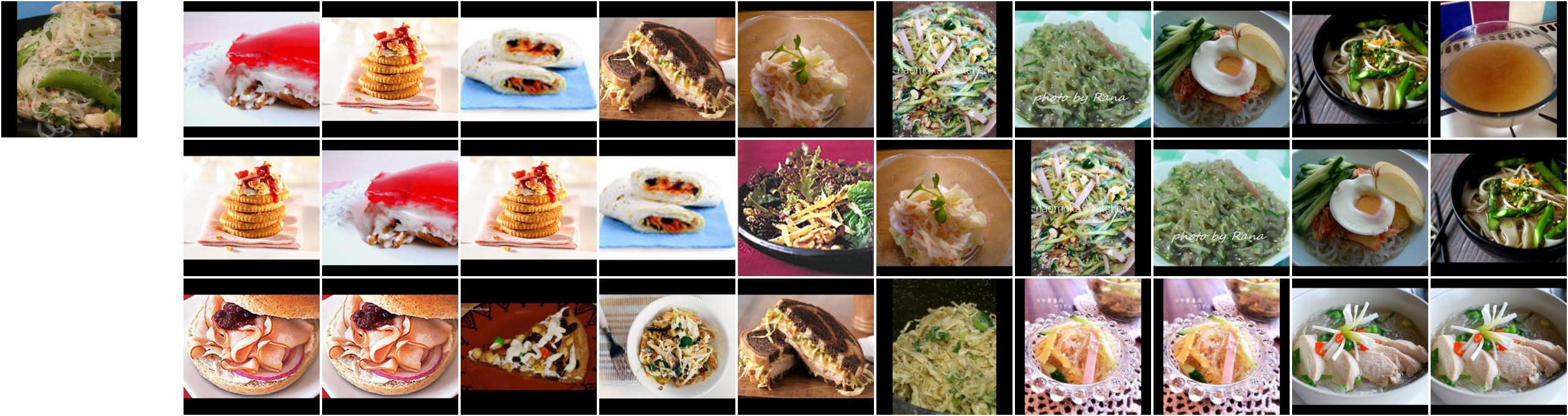}} \vspace{0.3em}\\ %\hline \hline
\raisebox{-.5\height}{\includegraphics[height=3.99cm]{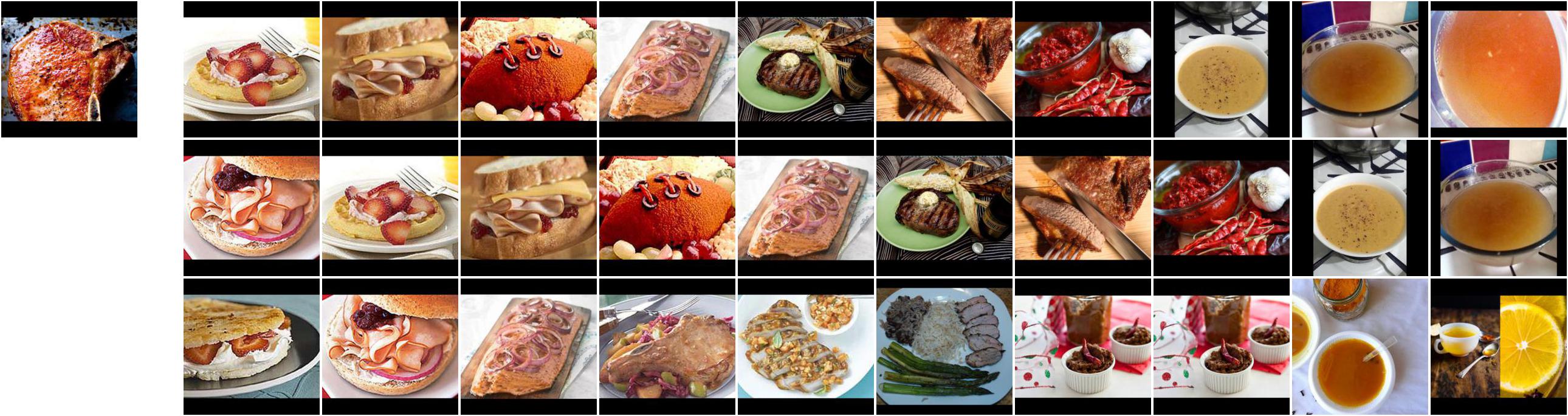}} \vspace{0.3em}\\ %\hline\hline
\raisebox{-.5\height}{\includegraphics[height=3.99cm]{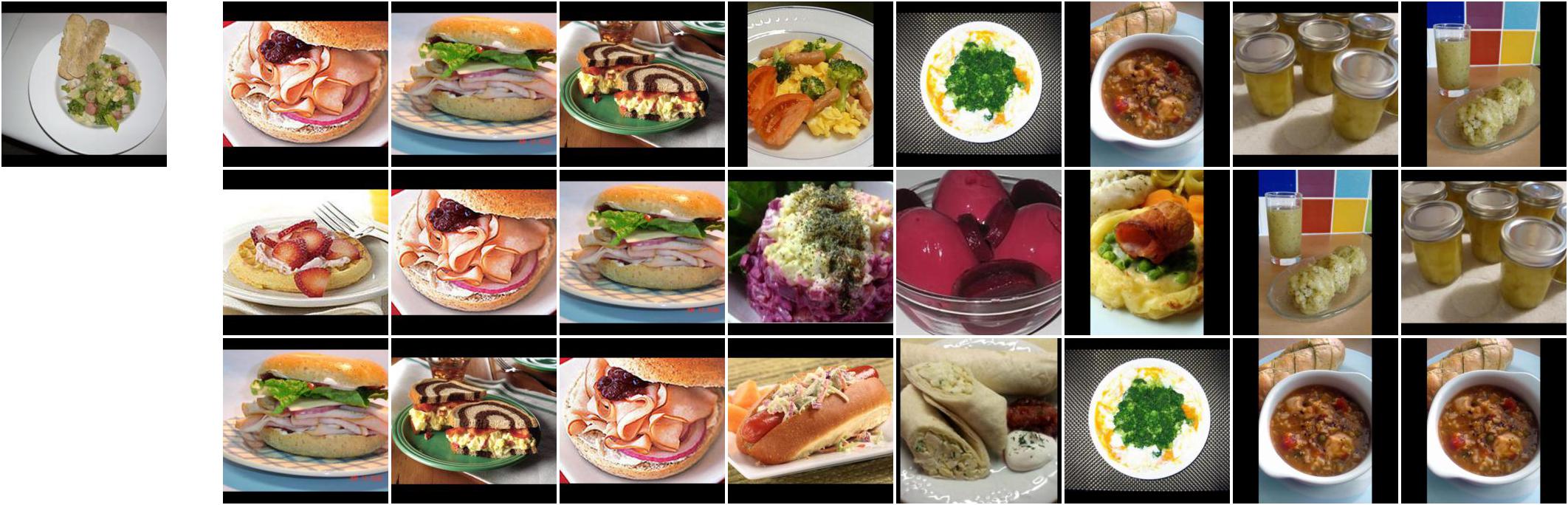}} \vspace{0.3em}\\ %\hline\hline
\raisebox{-.5\height}{\includegraphics[height=3.99cm]{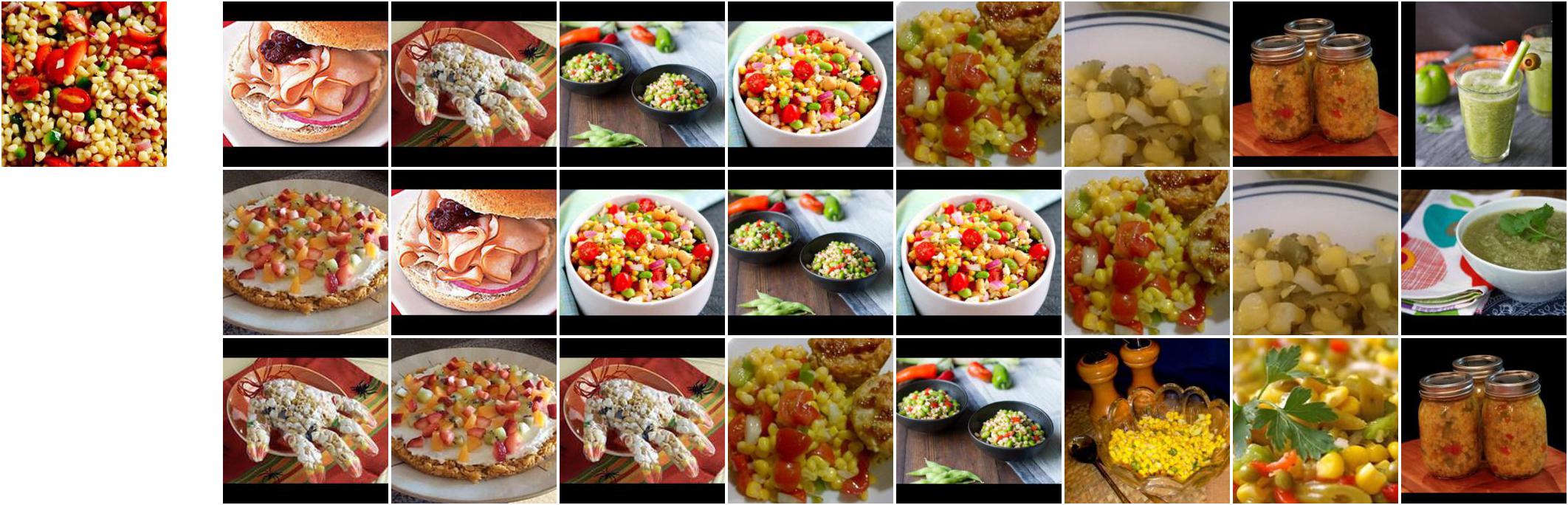}} \vspace{0.3em}\\ %\hline\hline
\raisebox{-.5\height}{\includegraphics[height=3.99cm]{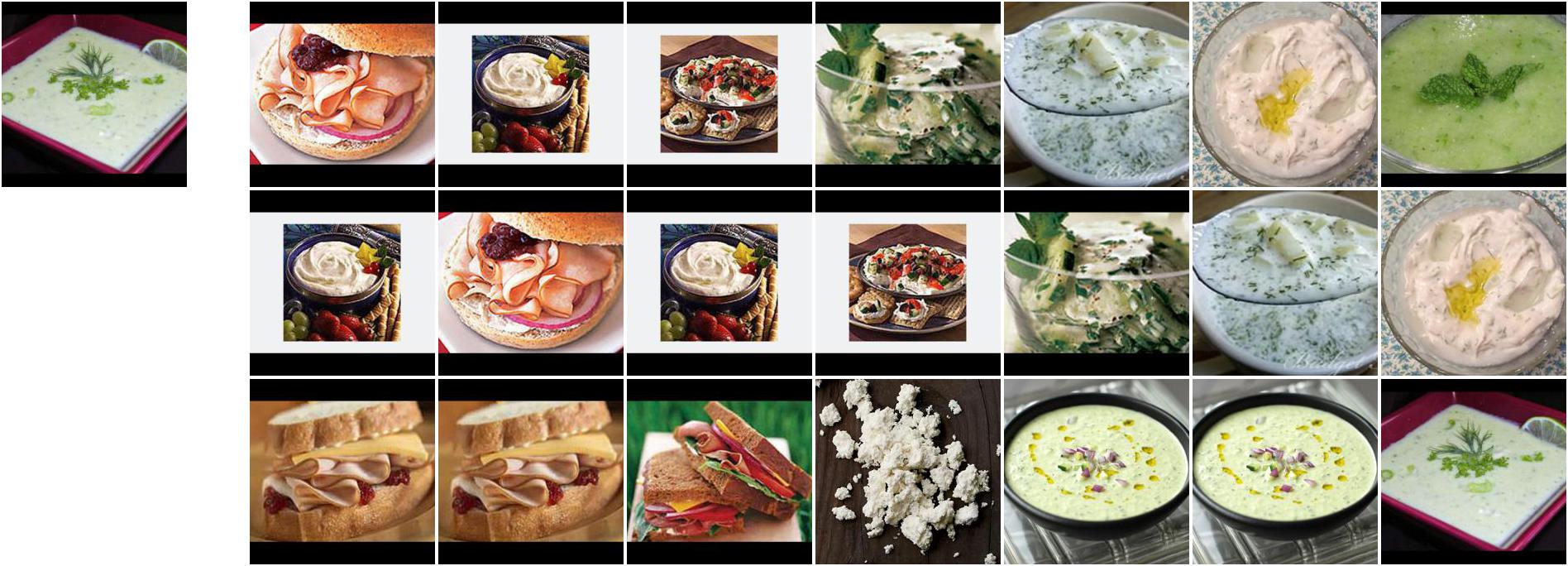}} \\ 
\vspace{-0.0em}
\caption{Images of the retrieved recipes from latent traversal along the latent variable $z_0$ ({\em wateriness}).  \textbf{Image grids}: the leftmost are the query images, and each column contains top-3 retrieved items at a point in traversal. Among the traversed points, we only show those that have change in the top-1 retrieved recipes from the previous traversed/retrieved ones. Note that the images on the left end of the traversal are less watery, while those on the right are more watery. 
%(Rows) Retrieved images for latent traversal, along $z_0$ (non watery - watery). \textbf{Image grids}: query image (left), each column top 3 NNs at a point in traversal $[\mu_i-10\sigma_i,\mu_i+10\sigma_i]$. Note that NNs are only shown for those points with a unique top 1, therefore, there are different amounts of retrieved images depending on the query.
}
\label{fig:supp_traversal0}
\vspace{-0.5em}
\end{figure}

\begin{figure}%[t!]
\raisebox{-.5\height}{\includegraphics[height=3.99cm]{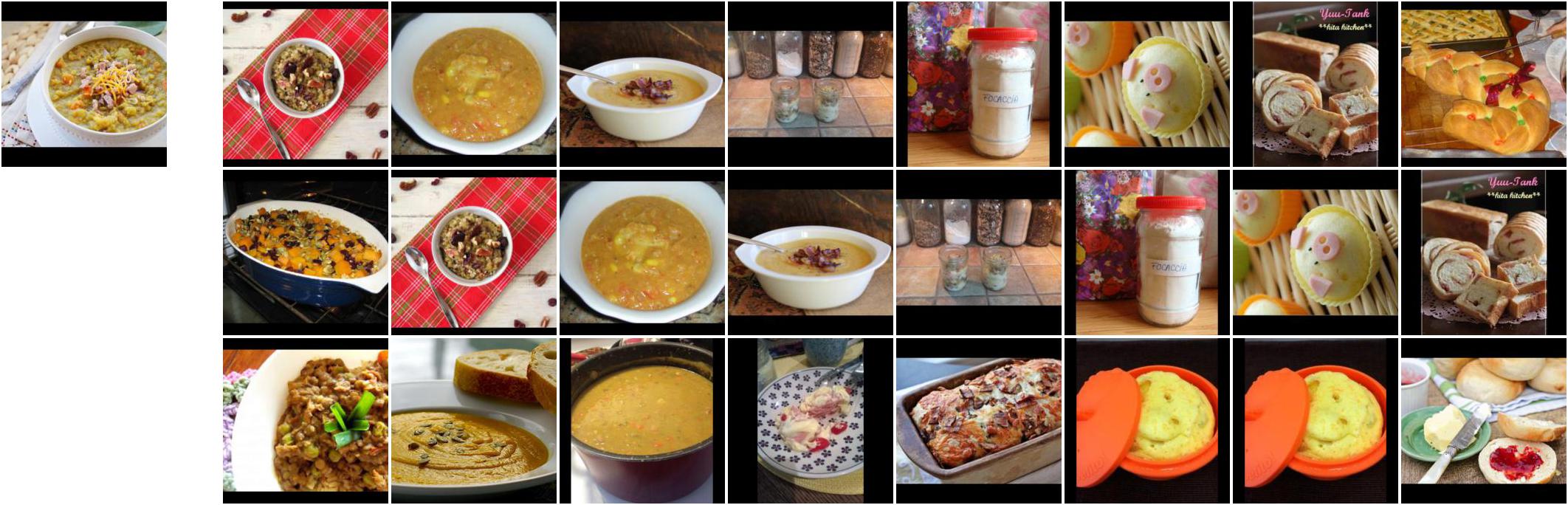}} \vspace{0.3em}\\ %\hline \hline
\raisebox{-.5\height}{\includegraphics[height=3.99cm]{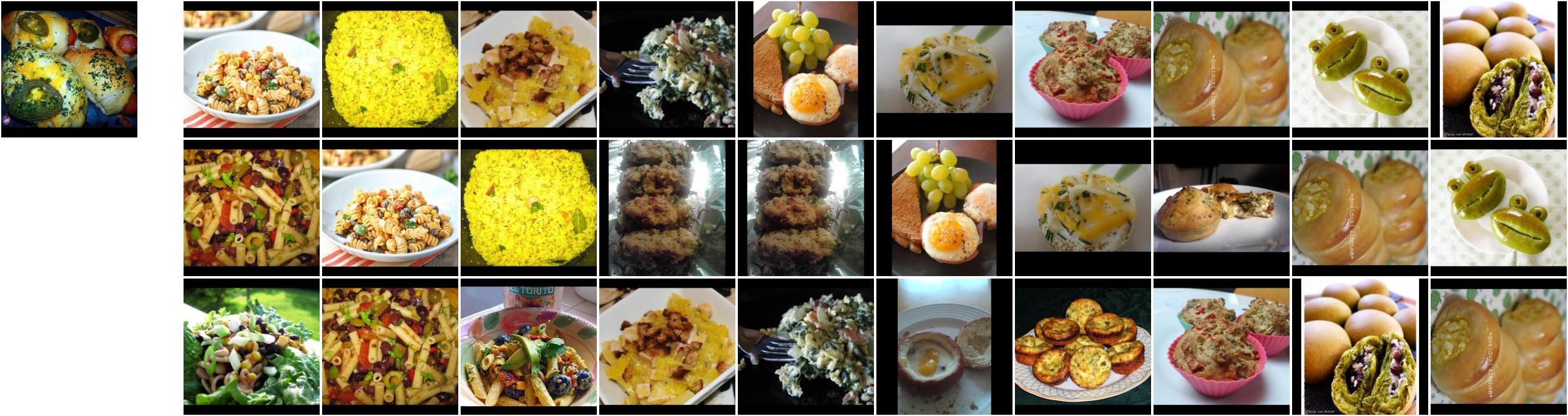}} \vspace{0.3em}\\ %\hline\hline
\raisebox{-.5\height}{\includegraphics[height=3.99cm]{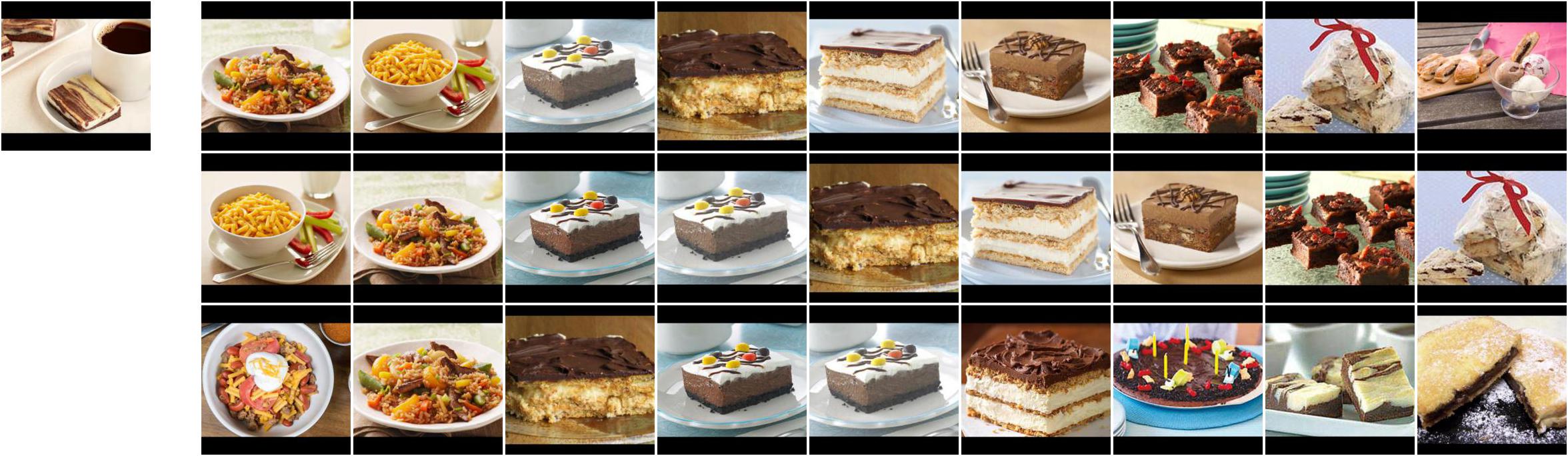}} \vspace{0.3em}\\ %\hline\hline
\raisebox{-.5\height}{\includegraphics[height=3.99cm]{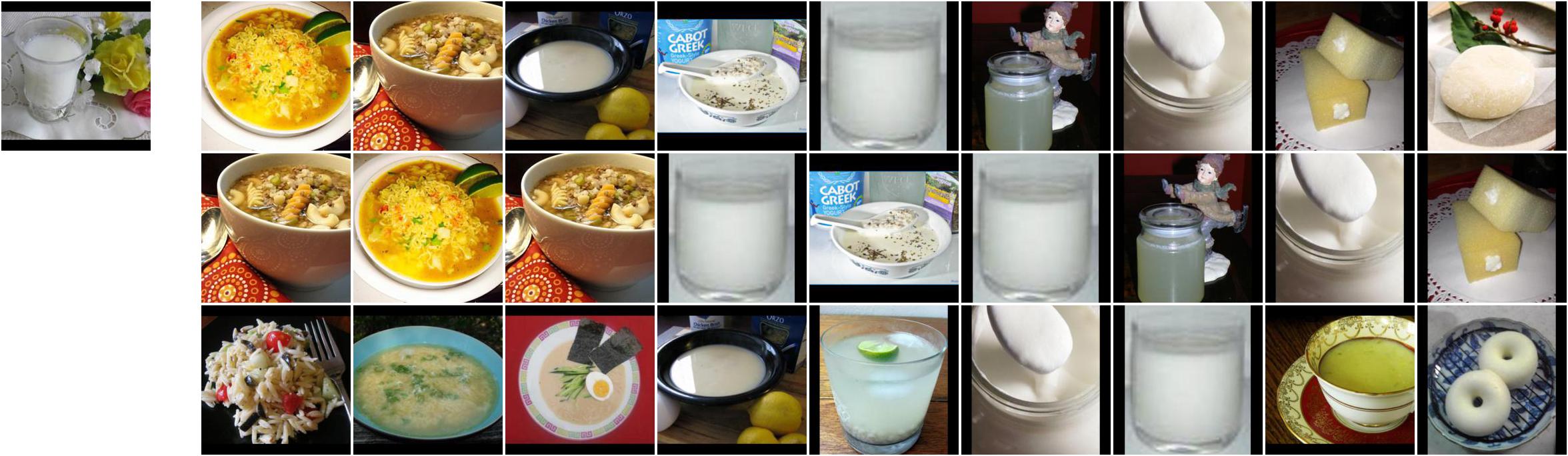}} \vspace{0.3em}\\ %\hline\hline
\raisebox{-.5\height}{\includegraphics[height=3.99cm]{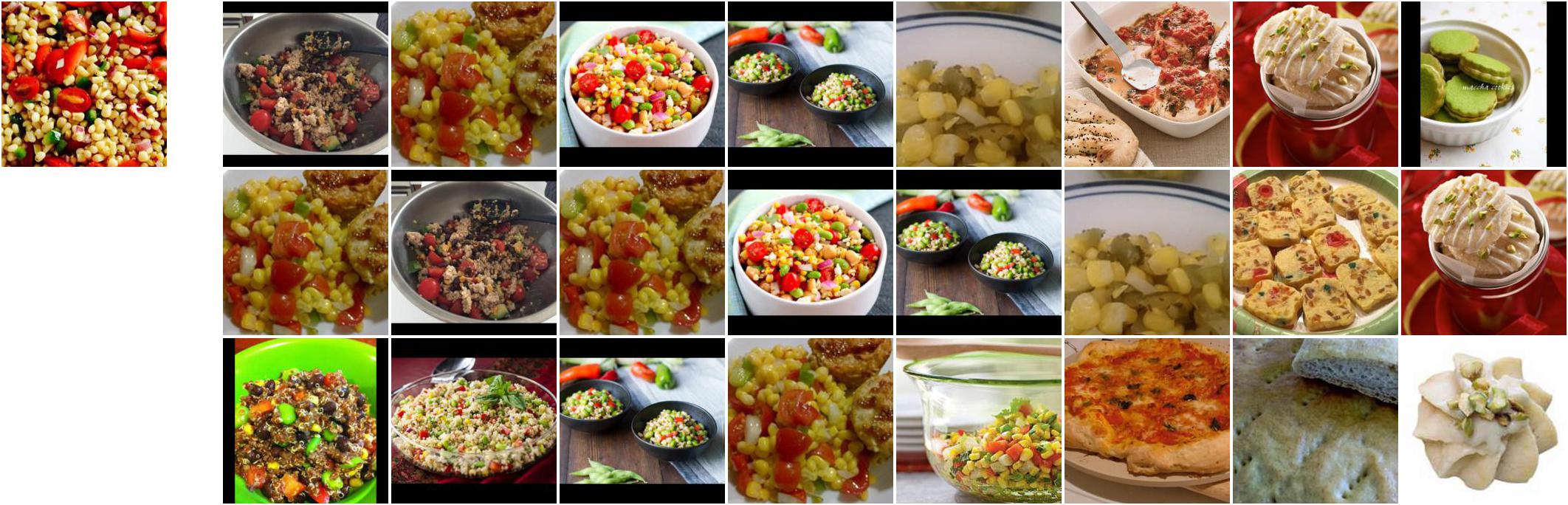}} \\ 
\vspace{-0.0em}
\caption{Images of the retrieved recipes from latent traversal along $z_2$ ({\em savoriness}). 
\textbf{Image grids}: the leftmost are the query images, and each column contains top-3 retrieved items at a point in traversal. Among the traversed points, we only show those that have change in the top-1 retrieved recipes from the previous traversed/retrieved ones. 
Note that the images on the left end of the traversal are more savory, while those on the right are less savory or sweet. 
%The same interpretation as Fig.~\ref{fig:supp_traversal0}.
%(Rows) Retrieved images for latent traversal, along $z_2$ (savory - sweet (baked)). \textbf{Image grids}: query image (left), each column top 3 NNs at a point in traversal $[\mu_i-10\sigma_i,\mu_i+10\sigma_i]$. Note that NNs are only shown for those points with a unique top 1, therefore, there are different amounts of retrieved images depending on the query.
}
\label{fig:supp_traversal2}
\vspace{-0.5em}
\end{figure}

\begin{figure}[t!]
\vspace{+1.5em}
\raisebox{-.5\height}{\includegraphics[height=3.99cm]{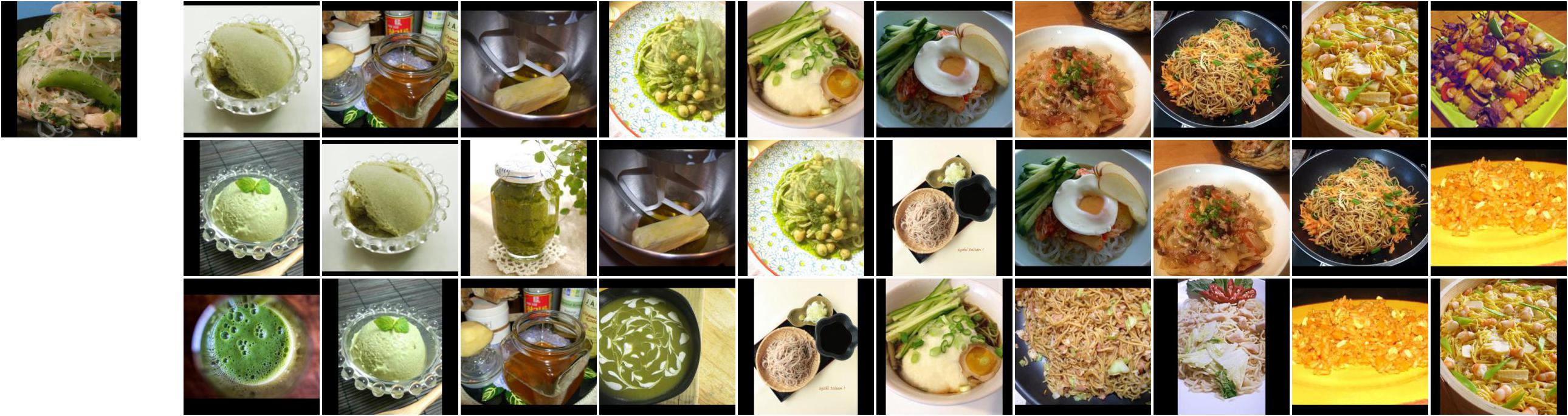}} \vspace{0.3em}\\ %\hline \hline
\raisebox{-.5\height}{\includegraphics[height=3.99cm]{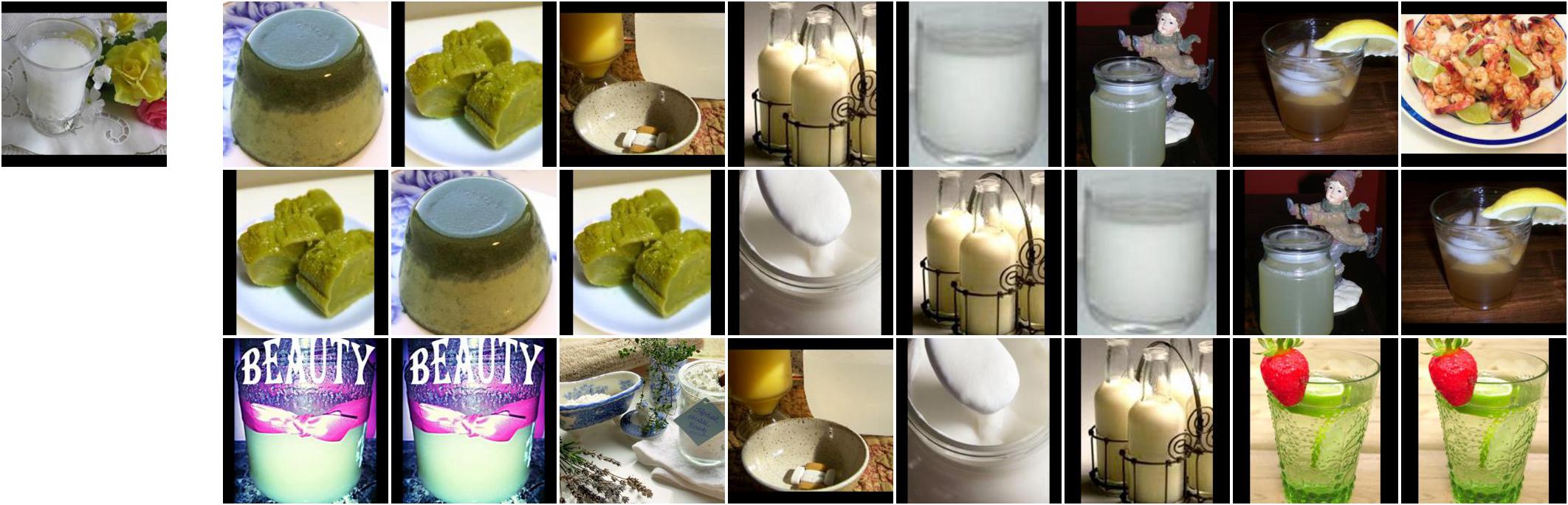}} \vspace{0.3em}\\ %\hline\hline
\raisebox{-.5\height}{\includegraphics[height=3.99cm]{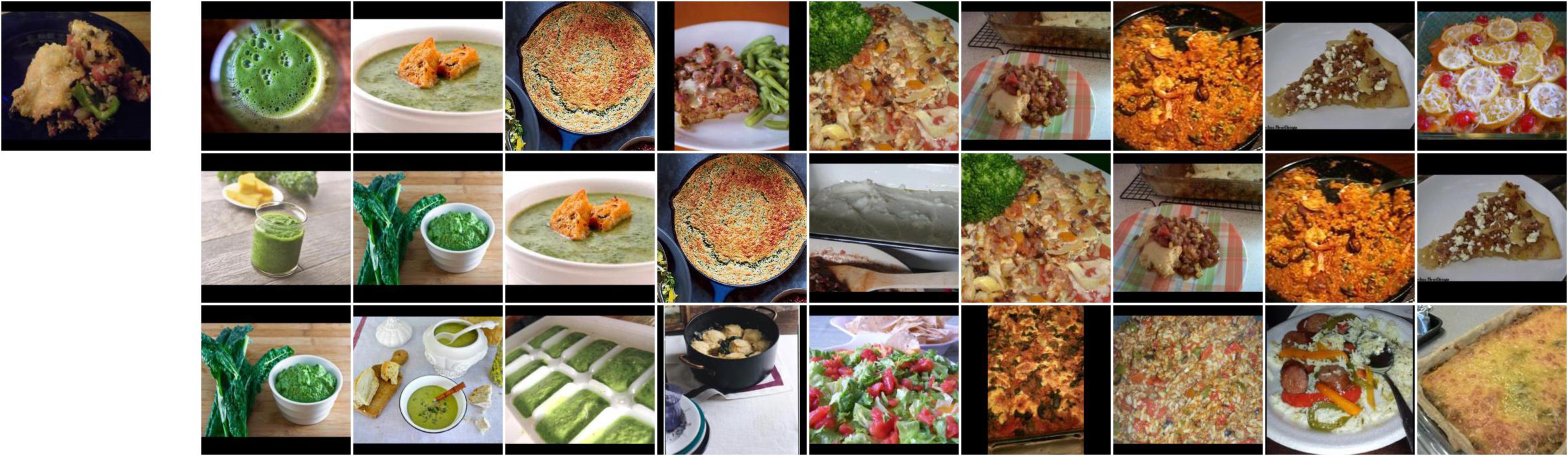}} \vspace{0.3em}\\ %\hline\hline
\raisebox{-.5\height}{\includegraphics[height=3.99cm]{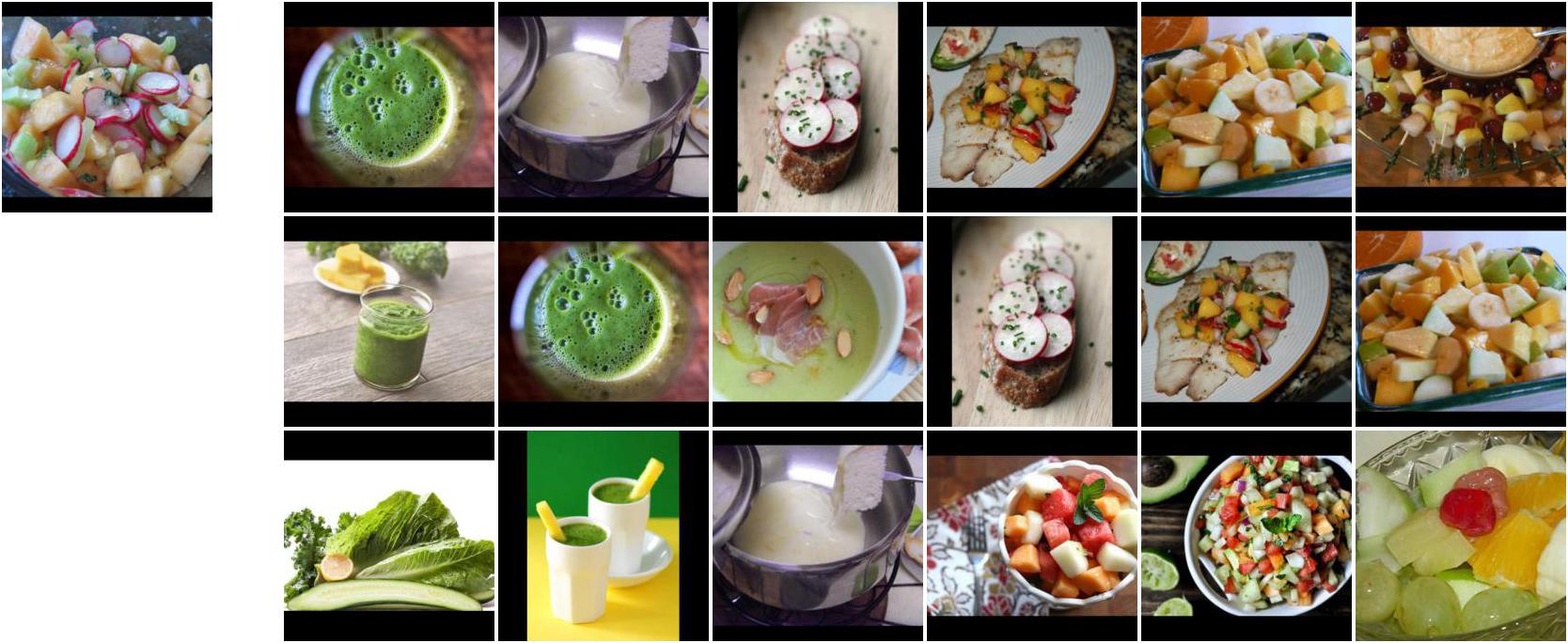}} \vspace{0.3em}\\ %\hline\hline
\raisebox{-.5\height}{\includegraphics[height=3.99cm]{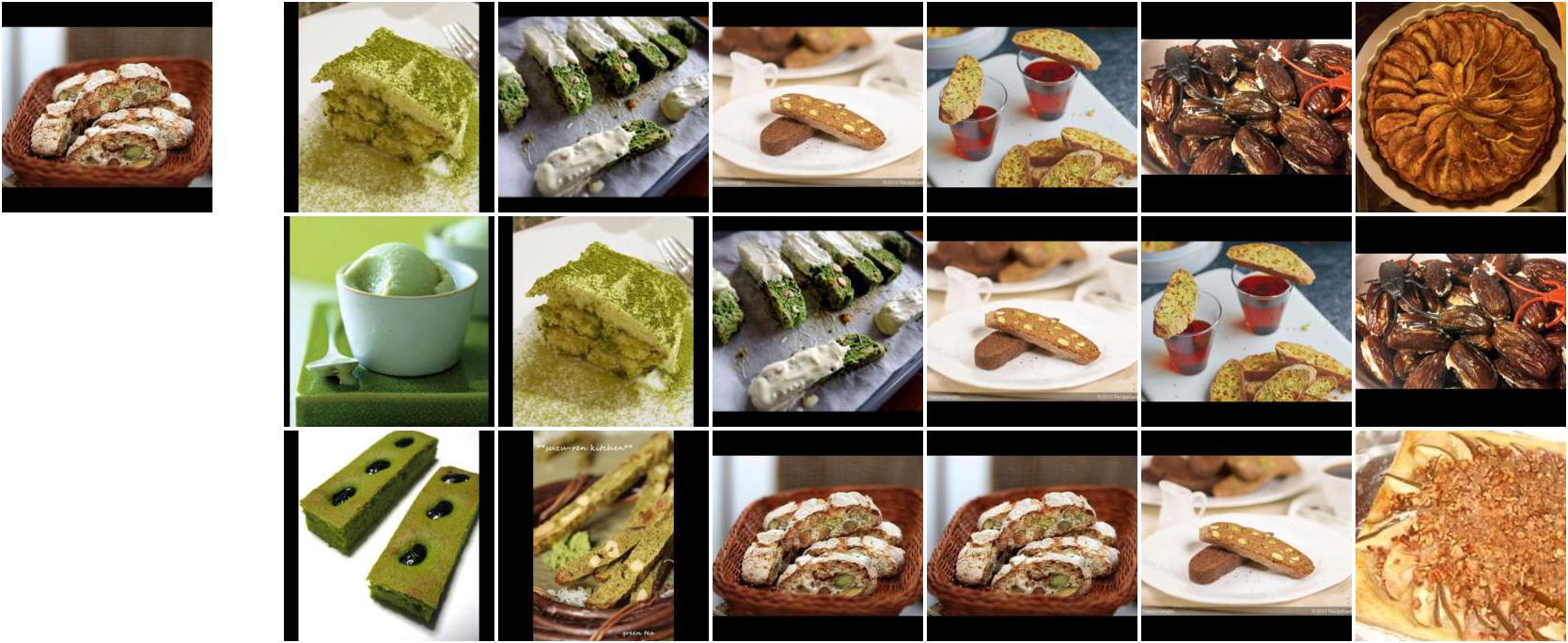}} \\ 
\vspace{-0.0em}
\caption{Images of the retrieved recipes from latent traversal along $z_3$ ({\em greenness}). %The same interpretation as Fig.~\ref{fig:supp_traversal0}. 
\textbf{Image grids}: the leftmost are the query images, and each column contains top-3 retrieved items at a point in traversal. Among the traversed points, we only show those that have change in the top-1 retrieved recipes from the previous traversed/retrieved ones. Note that the images on the left end of the traversal are more greenish, while those on the right are less greenish.
%(Rows) Retrieved images for latent traversal, along $z_3$ (greenish - non greenish). \textbf{Image grids}: query image (left), each column top 3 NNs at a point in traversal $[\mu_i-10\sigma_i,\mu_i+10\sigma_i]$. Note that NNs are only shown for those points with a unique top 1, therefore, there are different amounts of retrieved images depending on the query.
}
\label{fig:supp_traversal3}
\vspace{-0.5em}
\end{figure}

\begin{figure}[t!]
\vspace{+1.5em}
\raisebox{-.5\height}{\includegraphics[height=3.99cm]{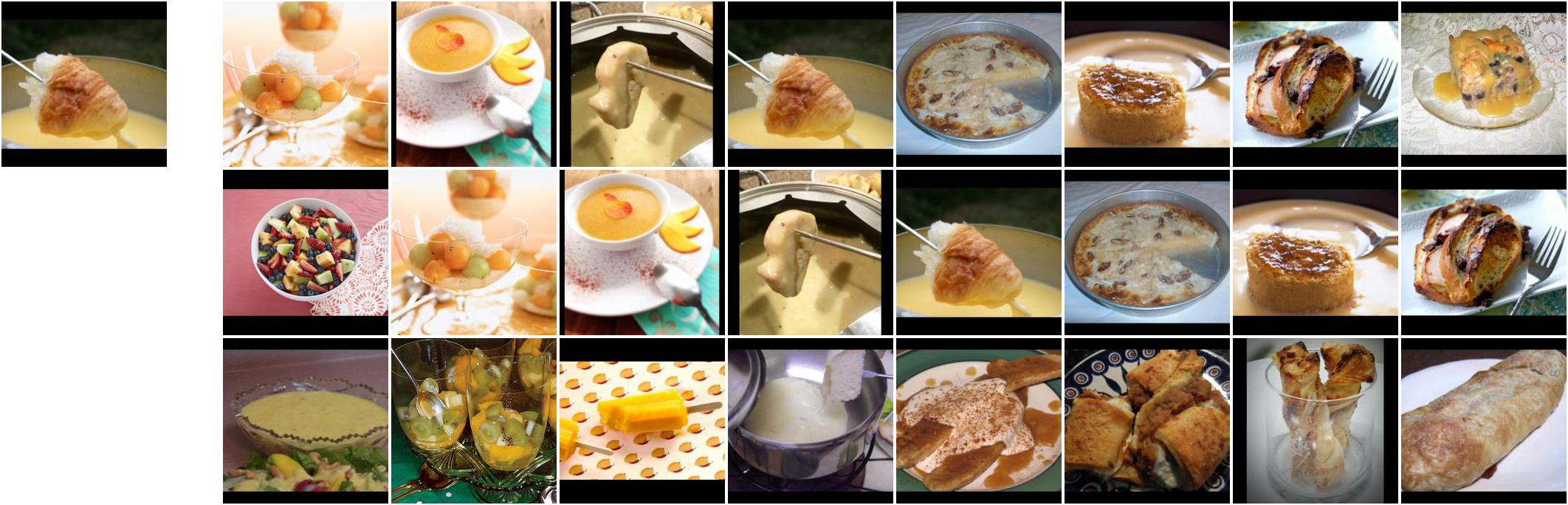}} \vspace{0.3em}\\ %\hline \hline
\raisebox{-.5\height}{\includegraphics[height=3.99cm]{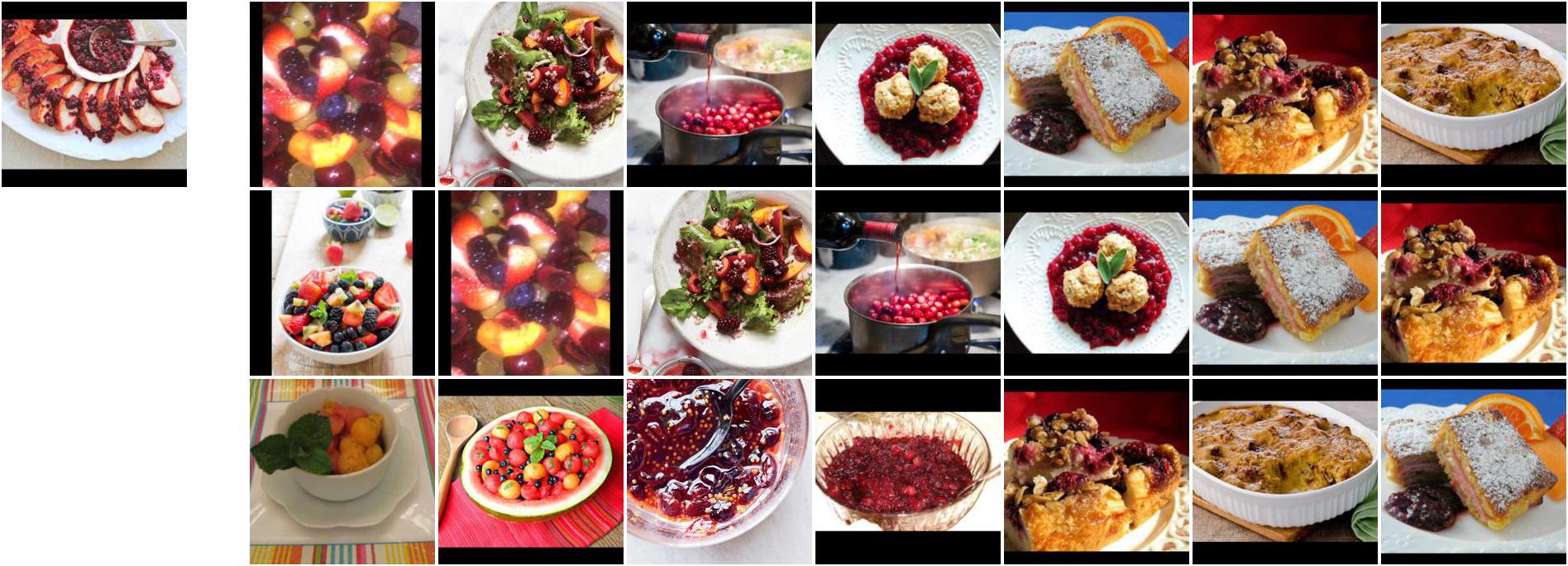}} \vspace{0.3em}\\ %\hline\hline
\raisebox{-.5\height}{\includegraphics[height=3.99cm]{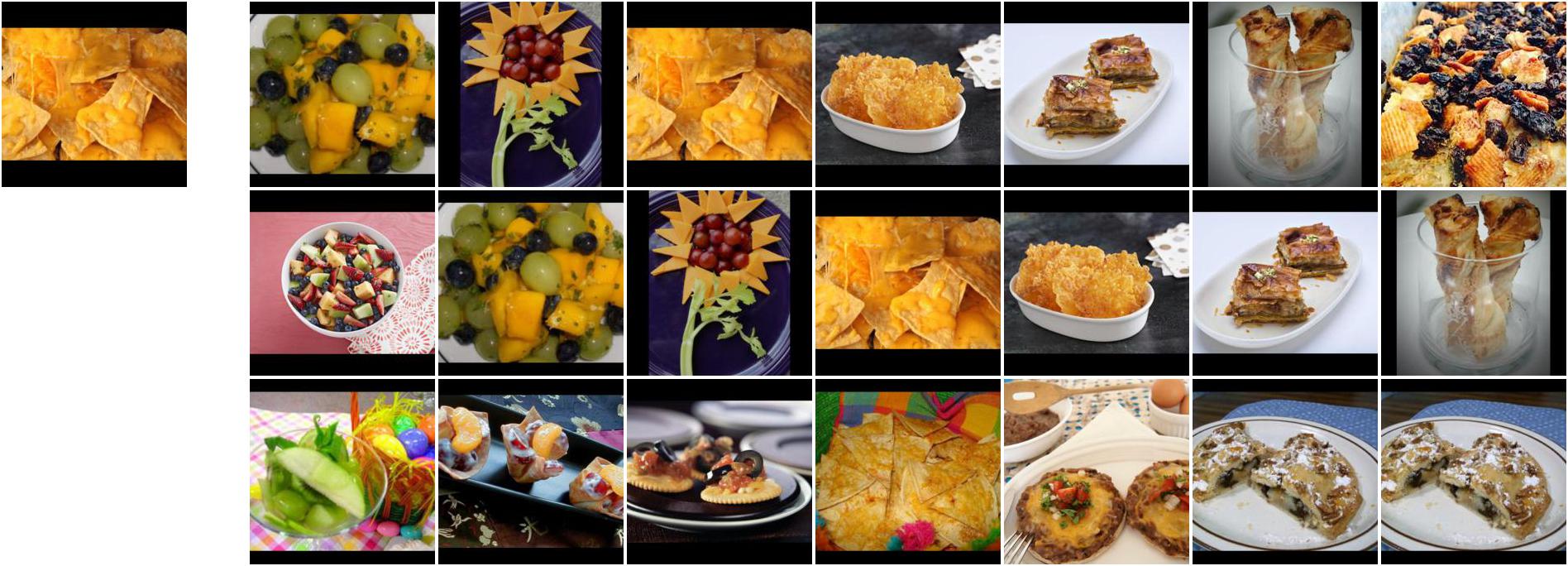}} \vspace{0.3em}\\ %\hline\hline
\raisebox{-.5\height}{\includegraphics[height=3.99cm]{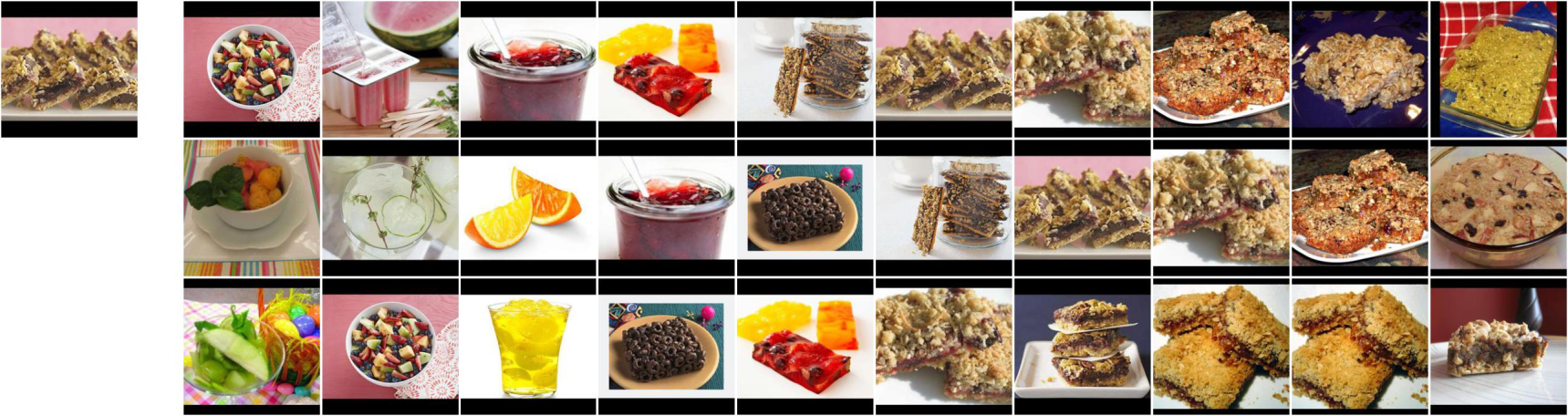}} \vspace{0.3em}\\ %\hline\hline
\raisebox{-.5\height}{\includegraphics[height=3.99cm]{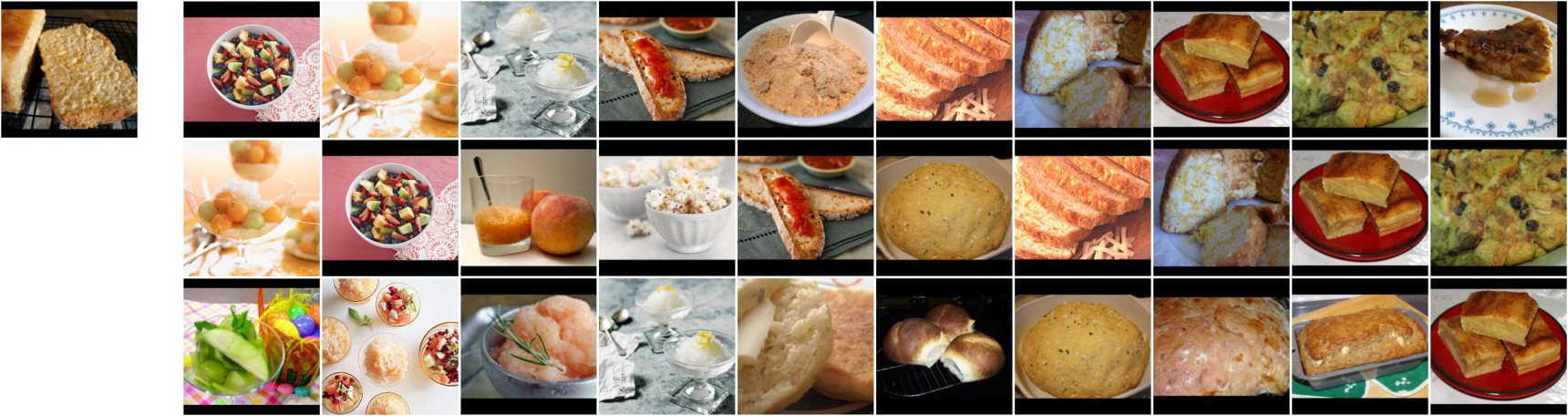}} \\ 
\vspace{-0.0em}
\caption{Images of the retrieved recipes from latent traversal along $z_{23}$ ({\em fruit-no-fruit}). %The same interpretation as Fig.~\ref{fig:supp_traversal0}. 
\textbf{Image grids}: the leftmost are the query images, and each column contains top-3 retrieved items at a point in traversal. Among the traversed points, we only show those that have change in the top-1 retrieved recipes from the previous traversed/retrieved ones. Note that the images on the left end of the traversal contain fruits, while those on the right have less or no fruits.
%(Rows) Retrieved images for latent traversal, along $z_3$ (greenish - non greenish). \textbf{Image grids}: query image (left), each column top 3 NNs at a point in traversal $[\mu_i-10\sigma_i,\mu_i+10\sigma_i]$. Note that NNs are only shown for those points with a unique top 1, therefore, there are different amounts of retrieved images depending on the query.
}
\label{fig:supp_traversal23}
\vspace{-0.5em}
\end{figure}

The qualitative latent traversal results for our RIVAE are depicted in Figs.~\ref{fig:supp_traversal0},  \ref{fig:supp_traversal2}, \ref{fig:supp_traversal3}, and \ref{fig:supp_traversal23}, each of which shows food images of the retrieved recipes in the latent traversal along the latent variables corresponding to: {\em wateriness}, {\em savoriness}, {\em greenness}, and {\em fruit-no-fruit}, respectively. 
Fig.~\ref{fig:supp_wordcloud} also displays the word clouds of the food ingredients, which are generated from the retrieved top 10 nearest neighbors (NN) from 20 query data points, where each latent variable takes three different values (two extrema and the mean, namely $\mu_i-10\sigma_i$, $\mu_i$, and $\mu_i+10\sigma_i$). Hence, they 
corresponding to {\em weak}, {\em mild}, and {\em strong} impact of each factor.
The size of each ingredient word in the word clouds corresponds to its relative frequency in the top 10 NN (from each of the 20 queries) set. 
Interpretation: For instance, in  Fig.~\ref{fig:supp_wordcloud_z23}, traversal along $z_{23}$ ({\em fruit-no-fruit}), we see that the ingredients are mostly associated with fruits on the left panel, while on the right panel no fruit related ingredients are found. 
Also, in Fig.~\ref{fig:supp_wordcloud_z2}, traversal along $z_{2}$ ({\em savoriness}), the ingredients are mostly associated with savory dishes on the left panel, while typical dessert ingredients on the right, and a mixture of these on the middle panel. 
In Fig.~\ref{fig:supp_wordcloud_z0}, traversal along $z_0$ ({\em wateriness}, there is an inclination of more liquid ingredients on the right side. In particular, it is noticeable that the word {\em water} increases in frequency (word size) from left to right. 
Similarly, in Fig.~\ref{fig:supp_wordcloud_z3}, $z_3$ ({\em greenness}), the left side contains more green ingredients than the right. %However, Fig.~\ref{fig:supp_wordcloud_z1}, that is, $z_1$ (non-noodle -- noodle), shows weaker evidence of extensive use of noodle (or pasta) ingredients, as indicated by the low strength score (0.35).

% \textbf{Failure of Existing Approaches.} Unlike our proposed models, the existing approaches fail to identify the underlying disentangled hidden factors. To verify this, we have conducted latent traversal in the {\em embedded space} $\mathcal{V}$ in the Cos-Sim model that is retrieval oriented. Due to the high dimensionality of the original Cos-Sim model, we use the Cos-Sim Bottleneck model that has 30-dim $\mathcal{V}$ space. The results are shown in Fig.~\ref{fig:supp_traversal_cossim} where the traversal is done in a similar manner as ours. We also encourage the reader to see the video version of the figure in the supplementary material. As shown, visually it is not easy to interpret the results; certain images often appear repeatedly across different dimensions. It implies that the Cos-Sim model merely yields latent representation that is significantly entangled with hidden factors. 

%%%%%%%%%%%%%%%%%%%%%%%%%%%%%%%%%%%%%%%%%%%%%%%%%%%%%%%%

\section{Recipe1M: Association between food factors and \url{food.com} tags}\label{sec:supp_food_tag_assoc}

\begin{figure}[t!]
%\vspace{+1.5em}
  %
  \centering
  \begin{subfigure}{0.33\textwidth}
    \centering
    \includegraphics[width=1.0\linewidth]{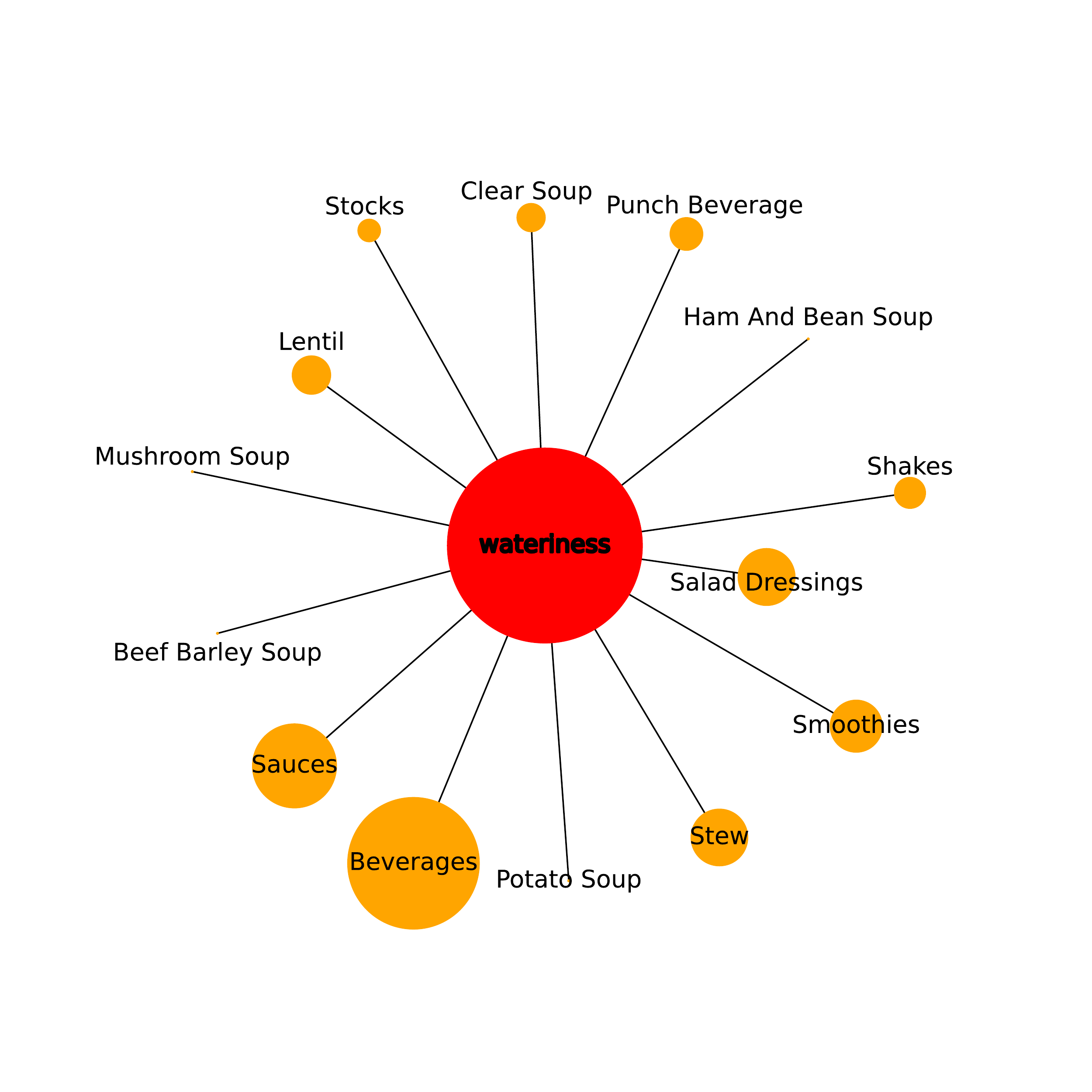}\vspace{-0.5em}
    % \caption{$z_{23}$ ({\em wateriness}): All tags associated with mostly watery dishes.}
    \label{fig:supp_association_wateriness}
  \end{subfigure}
  \centering
  \begin{subfigure}{0.33\textwidth}
    \centering
    \includegraphics[width=1.0\linewidth]{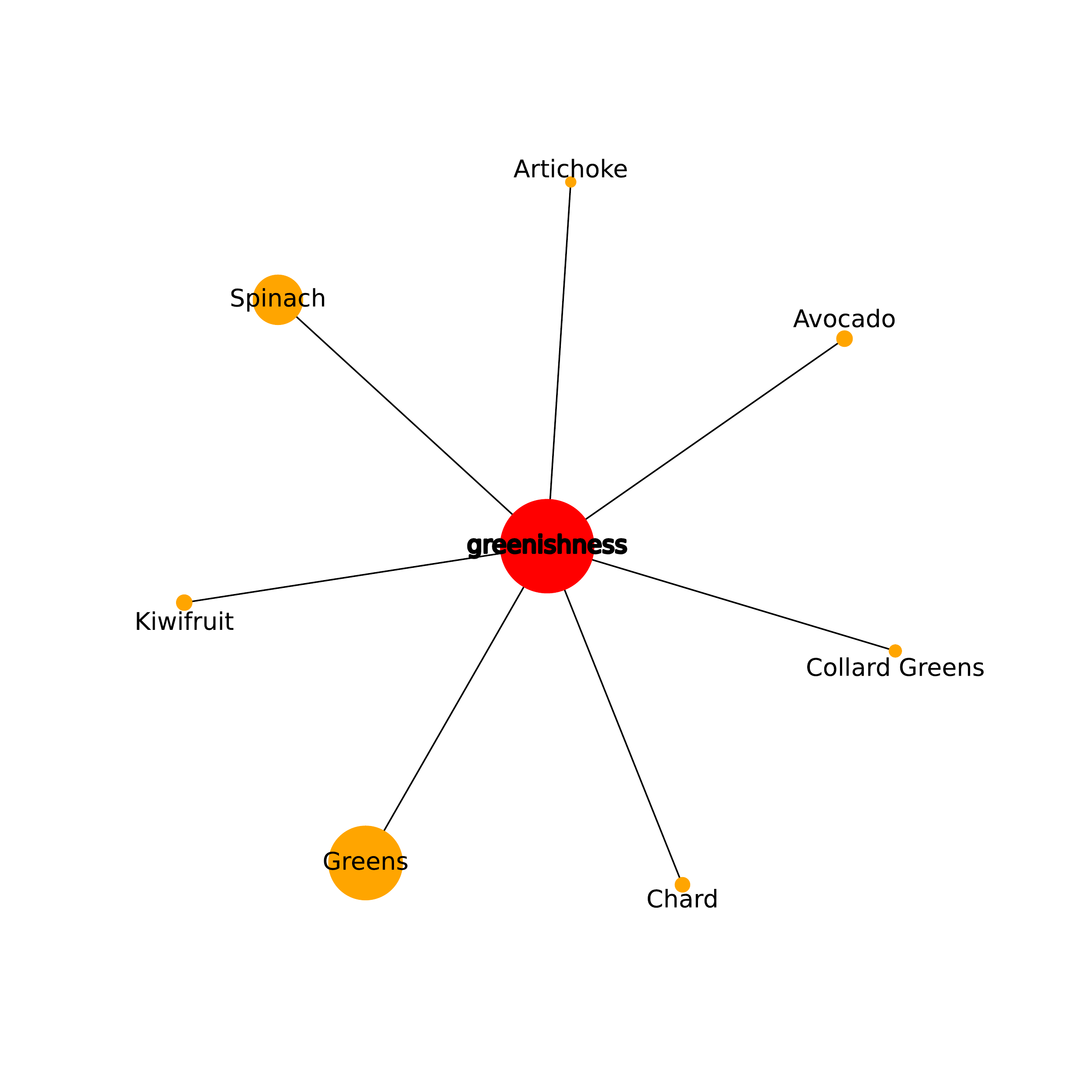}\vspace{-0.5em}
    % \caption{$z_{23}$ ({\em greenness}): All tags associated with mostly green dishes (mostly tagged as ingredients).}
    \label{fig:supp_association_wateriness}
  \end{subfigure}
  \centering
  \begin{subfigure}{0.33\textwidth}
    \centering
    \includegraphics[width=1.0\linewidth]{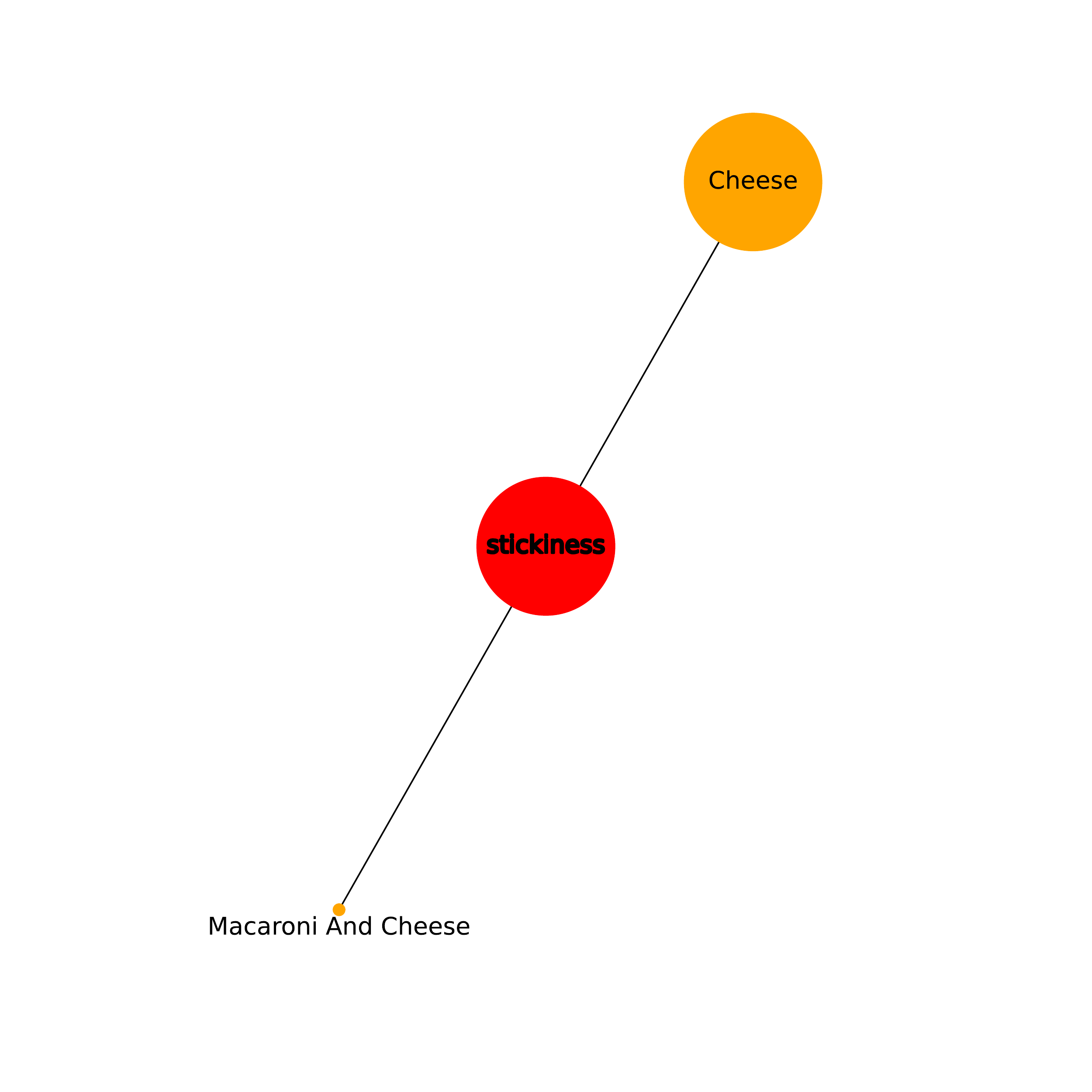}\vspace{-0.5em}
    % \caption{$z_{23}$ ({\em greenness}): All tags associated with mostly sticky dishes (mostly tagged as ingredients).}
    \label{fig:supp_association_stickiness}
  \end{subfigure}
  \\ \vspace{+0.3em}%
  \centering
  \begin{subfigure}{0.33\textwidth}
    \centering
    \includegraphics[width=1.0\linewidth]{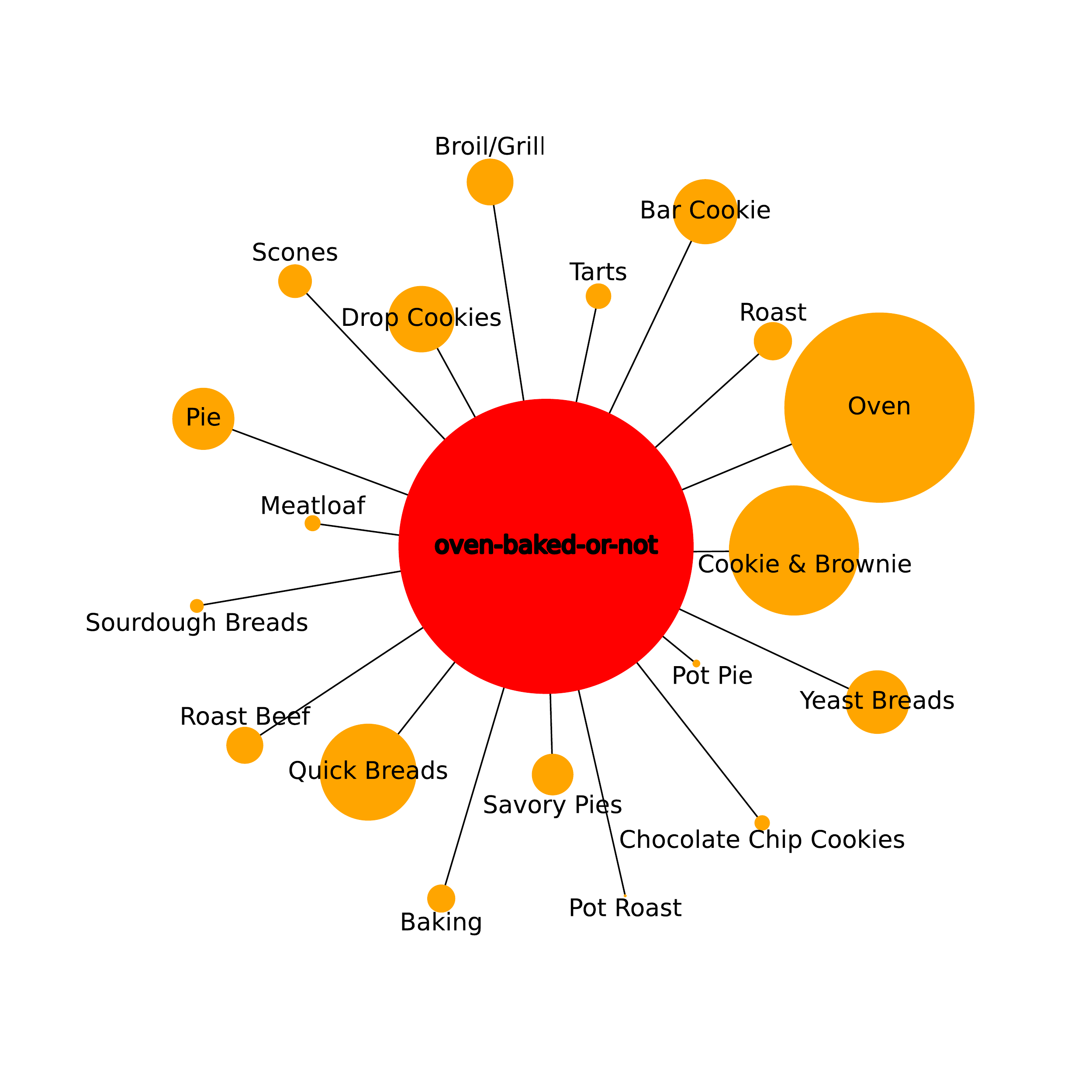}\vspace{-0.5em}
    % \caption{$z_{23}$ ({\em wateriness}): All tags associated with mostly watery dishes.}
    \label{fig:supp_association_oven}
  \end{subfigure}
  \centering
  \begin{subfigure}{0.33\textwidth}
    \centering
    \includegraphics[width=1.0\linewidth]{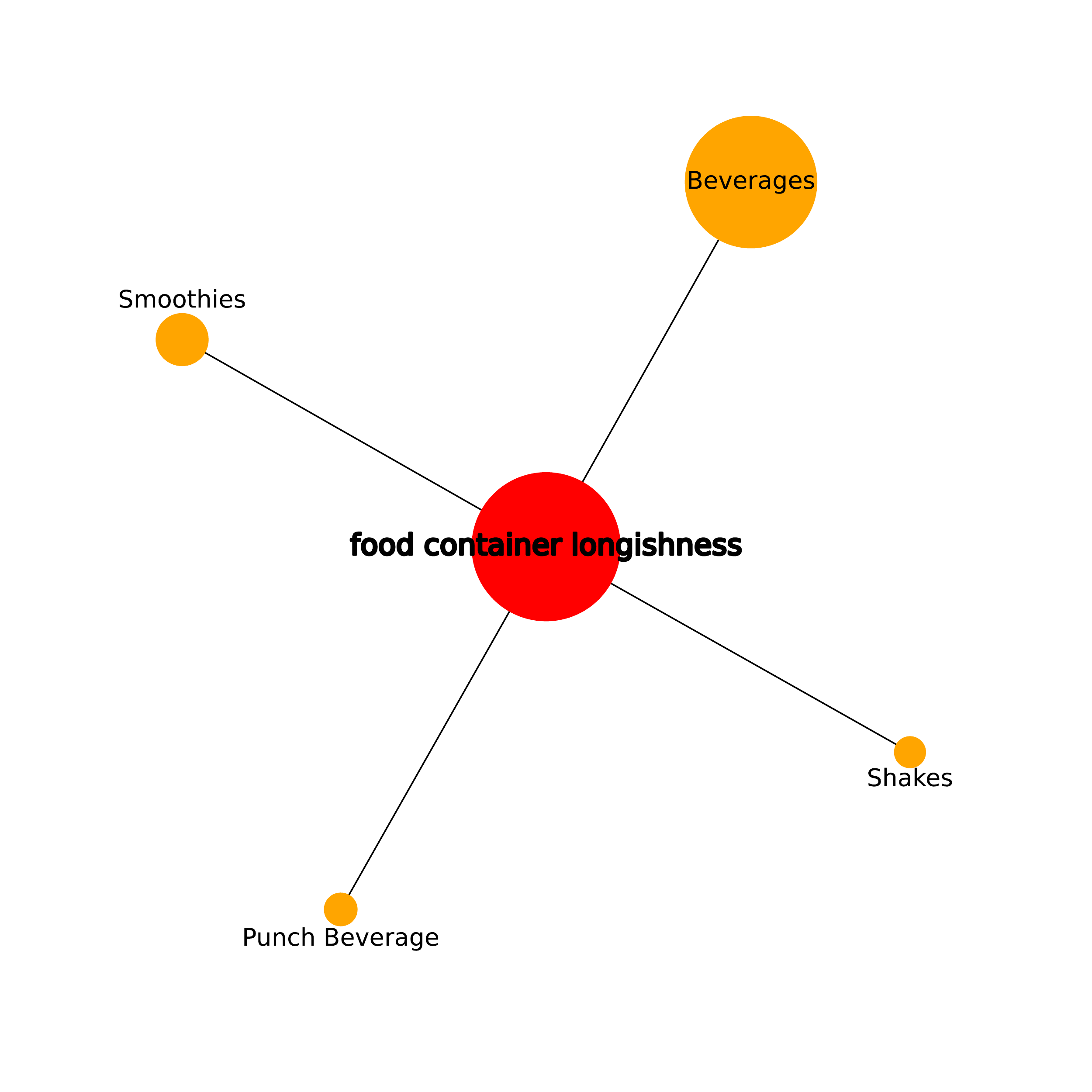}\vspace{-0.5em}
    % \caption{$z_{23}$ ({\em greenness}): All tags associated with mostly green dishes (mostly tagged as ingredients).}
    \label{fig:supp_association_container}
  \end{subfigure}
  \centering
  \begin{subfigure}{0.33\textwidth}
    \centering
    \includegraphics[width=1.0\linewidth]{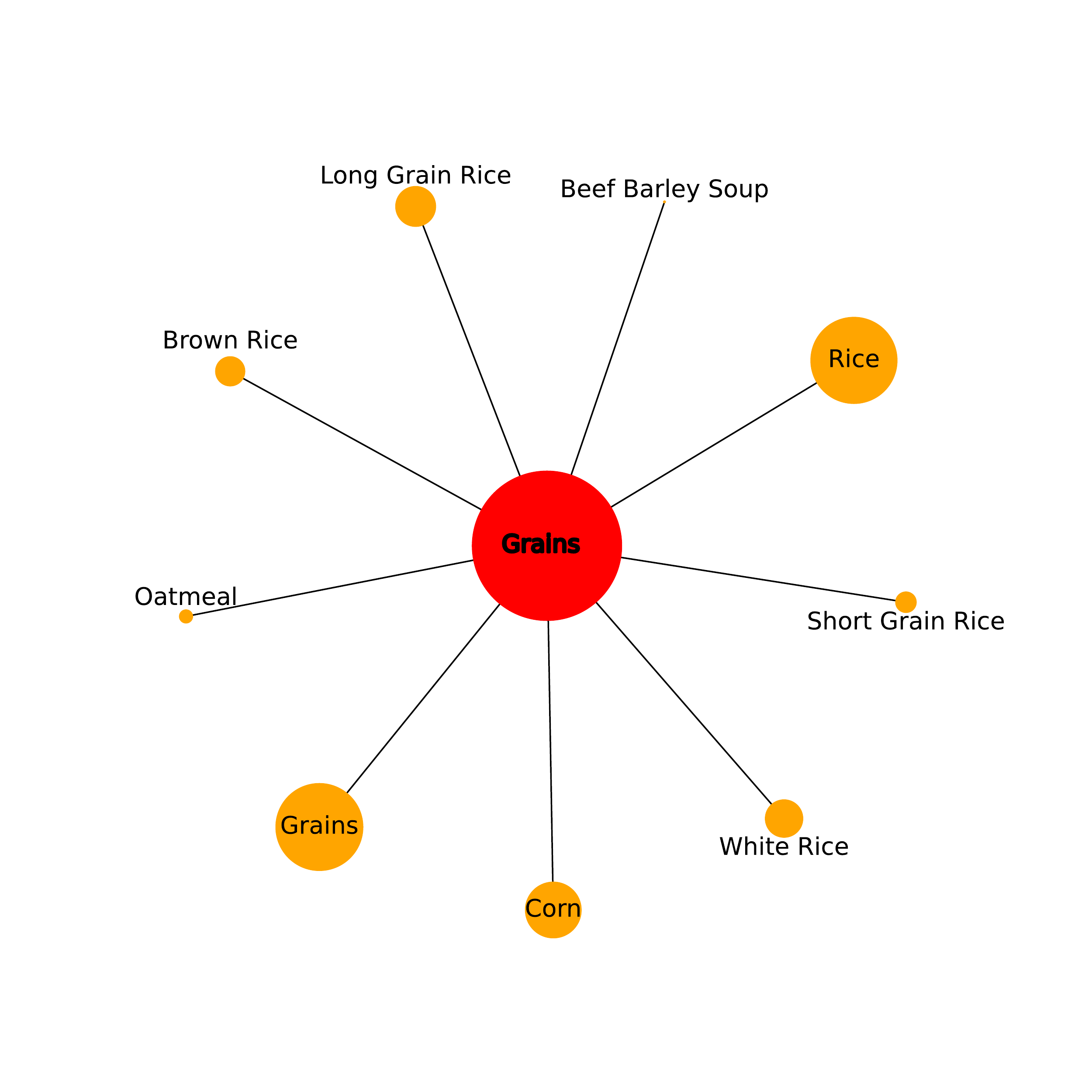}\vspace{-0.5em}
    % \caption{$z_{23}$ ({\em greenness}): All tags associated with mostly sticky dishes (mostly tagged as ingredients).}
    \label{fig:supp_association_grains}
  \end{subfigure}
  \\ \vspace{+0.3em}%
  \centering
  \begin{subfigure}{0.33\textwidth}
    \centering
    \includegraphics[width=1.0\linewidth]{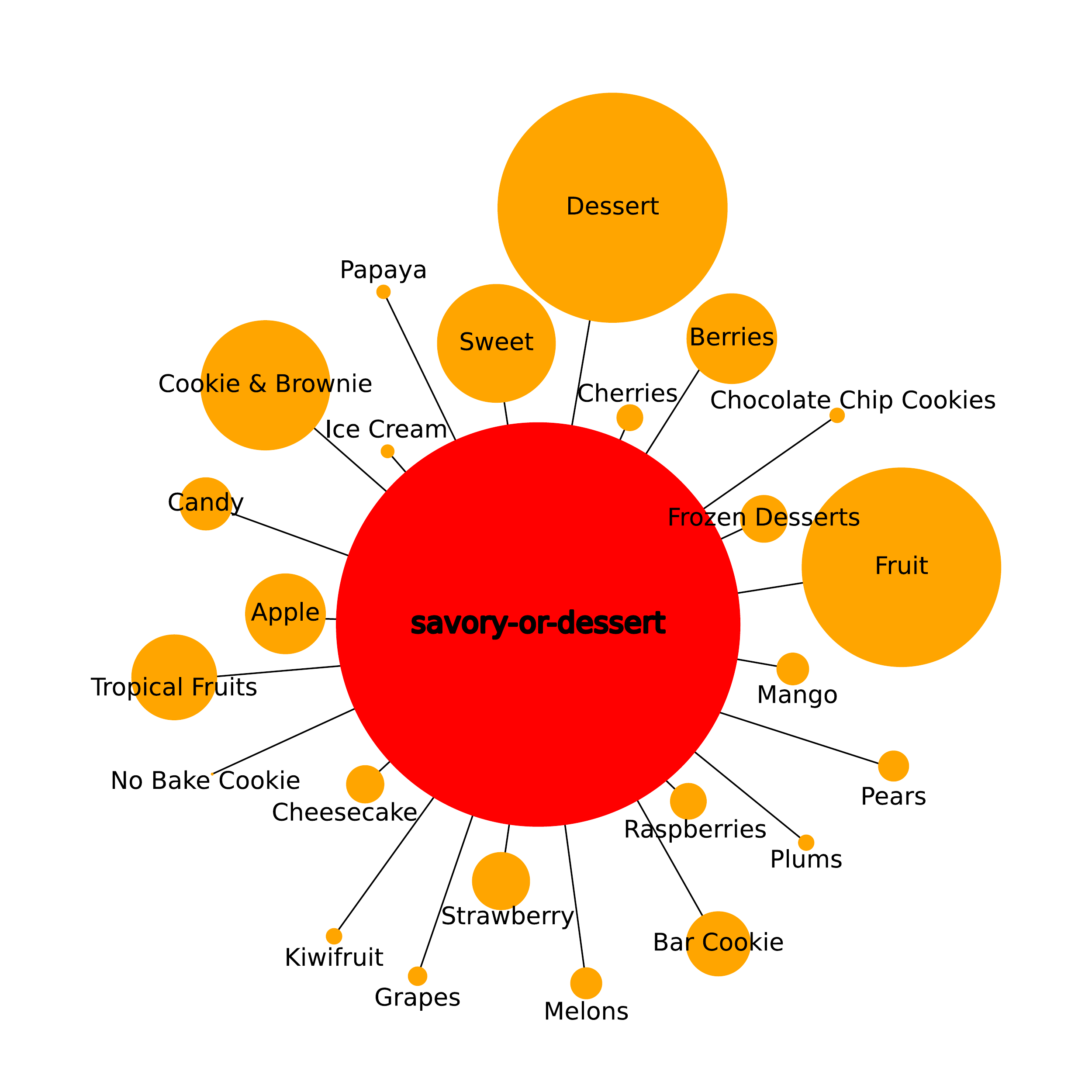}\vspace{-0.5em}
    % \caption{$z_{23}$ ({\em wateriness}): All tags associated with mostly watery dishes.}
    \label{fig:supp_association_dessert}
  \end{subfigure}
  \centering
  \begin{subfigure}{0.33\textwidth}
    \centering
    \includegraphics[width=1.0\linewidth]{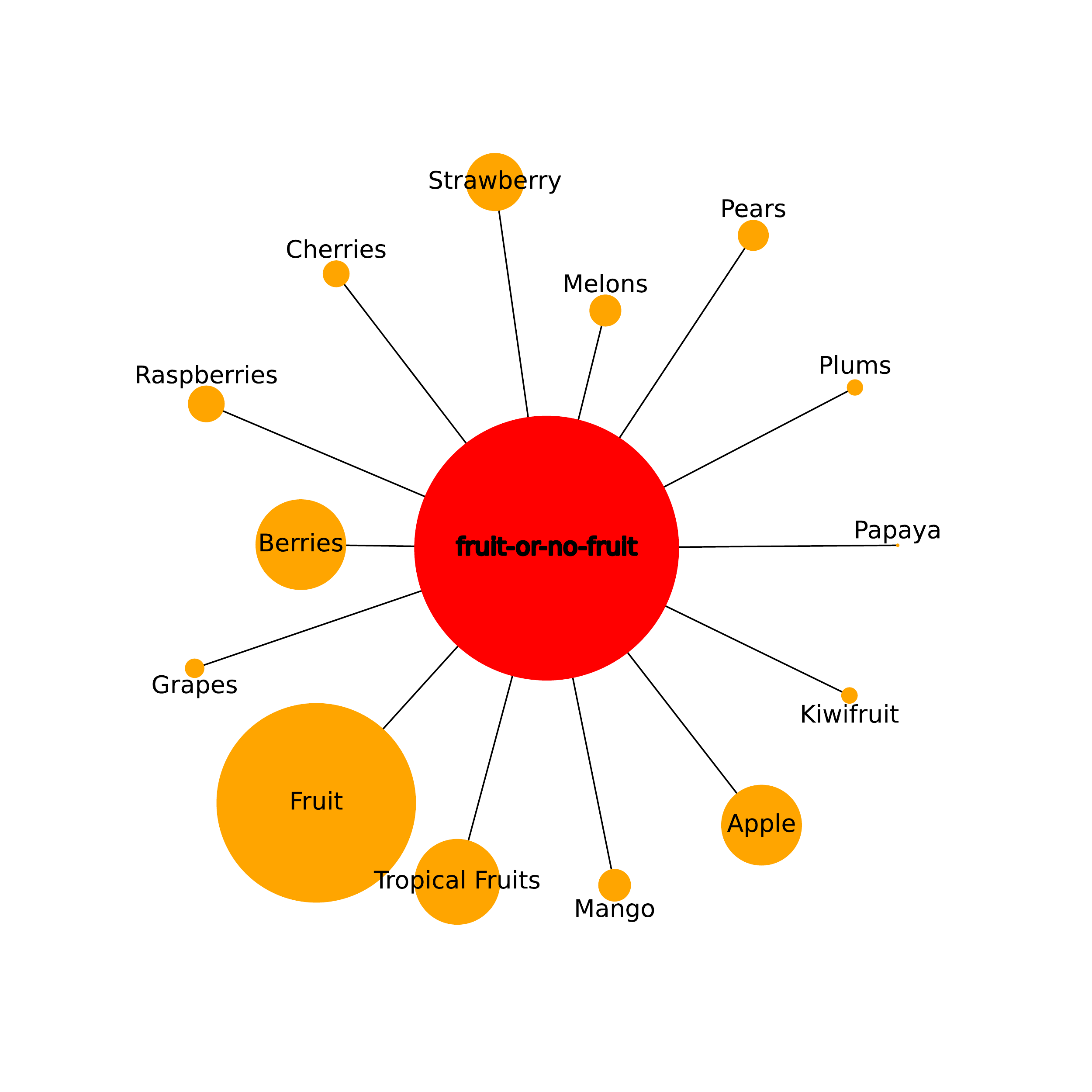}\vspace{-0.5em}
    % \caption{$z_{23}$ ({\em greenness}): All tags associated with mostly green dishes (mostly tagged as ingredients).}
    \label{fig:supp_association_fruit}
  \end{subfigure}
\vspace{-0.8em}
\caption{(Recipe1M) Association between tags (shown as orange) extracted from \url{food.com} recipes (a subset of Recipe1M) and our %discovered 
8 food factors (red). The sizes of the nodes are proportional to their frequencies. 
}
\label{fig:supp_wordgraph}
%\vspace{-1.0em}
\end{figure}

Recall that we selected a small subset of food factors by re-craping all the recipes in the Recipe1M that are associated with the Internet domain \url{food.com}, and parsing their keywords and categories. We then formed the recipe {\em tags} as the unique terms union between keywords and categories, and manually grouped the tags that are related to one another. Among the tag groups, we chose 8 factors the most dependent on the latent variables by visually inspecting the latent traversal results for the competing methods. 
They are: 1) wateriness, 2) greenness, 3) stickiness, 4) oven-baked-or-not, 5) food container longishness (e.g., bottle/cup or plate), 6) grains, 7) savory-or-dessert, and 8) fruit-or-no-fruit.  Fig.~\ref{fig:supp_wordgraph} shows the association between the tags and these 8 factors.

%%%%%%%%%%%%%%%%%%%%%%%%%%%%%%%%%%%%%%%%%%%%%%%%%%%%%%%%%%%%%%%%%%%%%%%%%%%%%%%
%%%%%%%%%%%%%%%%%%%%%%%%%%%%%%%%%%%%%%%%%%%%%%%%%%%%%%%%%%%%%%%%%%%%%%%%%%%%%%%
\section{Details of the joint model,  Retrieval-Bi-VAE (RBi-VAE)}\label{sec:supp_rbi_vae}

In this section we describe the details of the joint model RBi-VAE that served as one of the competing approaches in the empirical study. Note that unlike our proposed RIVAE, the model has no {\em model identifiability} property, which leads to inferior performance to our RIVAE in the experimental results. 

This model builds a latent variable generative model (e.g., VAE) on top of the shared embedding space $\mathcal{V}$. To this end, the embedded vectors ${\bf v}$'s are regarded as the observed data for the VAE. That is, the RBi-VAE model is defined as:
%%%%
\begin{align}
&P({\bf z}) = \mathcal{N}(0, {\bf I}), \ %P({\bf v}_1,{\bf v}_2|{\bf z}) = 
%P({\bf v}_1|{\bf z}) \cdot P({\bf v}_2|{\bf z}) \  \textrm{where} 
P({\bf v}_i|{\bf z}) = \mathcal{N}({\bf v}_i; {\boldsymbol\mu}_i({\bf z}), \textrm{Diag}({\boldsymbol\sigma}_i({\bf z}))) \ \textrm{for} \ i=1,2,
\label{eq:supp_dfr_pvz} \\
& {\bf v}_1 = {\bf e}_1({\bf x}_1), \ {\bf v}_2 = {\bf e}_2({\bf x}_2). \label{eq:supp_embedding}
\end{align}
%%%%
The latent variables ${\bf z}\in\mathbb{R}^d$ represent the underlying factors, and ${\boldsymbol\mu}_{1/2}(\cdot)$ and ${\boldsymbol\sigma}_{1/2}(\cdot)$ are the deep networks whose outputs determine means and variances of $P({\bf v}_{1/2}|{\bf z})$.

Similarly as RIVAE, we train RBi-VAE with two goals: maximizing the cross-modal prediction performance and maximizing the  (embedded) data likelihood. 
Given the embedding networks, with ${\bf v}_1$ and ${\bf v}_2$ fixed, learning the Bi-VAE model can be done by maximizing the variational evidence lower bound (ELBO), 
%%%%
\begin{equation}
%\log P({\bf v}_1,{\bf v}_2) \geq 
\textrm{ELBO} := \mathbb{E}_{Q({\bf z}|{\bf v}_1,{\bf v}_2)}[\log P({\bf v}_1,{\bf v}_2|{\bf z})] - 
\textrm{KL}( Q({\bf z}|{\bf v}_1,{\bf v}_2)||P({\bf z})) %\nonumber
\label{eq:supp_elbo}
\end{equation}
%%%%
where the variational density $Q({\bf z}|{\bf v}_1,{\bf v}_2)$ is defined as a product of the single-modal inference networks~\cite{poe}, 
$Q({\bf z}|{\bf v}_1,{\bf v}_2) \propto Q({\bf z}|{\bf v}_1) \cdot  Q({\bf z}|{\bf v}_2)$, where
%%%%
\begin{equation}
%Q({\bf z}|{\bf v}_1) &:= \mathcal{N}({\bf z}; {\bf m}_1({\bf v}_1), \textrm{Diag}({\bf s}_1({\bf v}_1))) \\
%Q({\bf z}|{\bf v}_2) &:= \mathcal{N}({\bf z}; {\bf m}_2({\bf v}_2), \textrm{Diag}({\bf s}_2({\bf v}_2))) \\
Q({\bf z}|{\bf v}_i) := \mathcal{N}({\bf z}; {\bf m}_i({\bf v}_i), \textrm{Diag}({\bf s}_i({\bf v}_i))) \ \ \textrm{for} \ i=1,2. %\\
%Q({\bf z}|{\bf v}_1,{\bf v}_2) &\propto Q({\bf z}|{\bf v}_1) \cdot  Q({\bf z}|{\bf v}_2), \label{eq:supp_poe_q}
\end{equation}
%%%%

Furthermore, to encourage the model to learn disentangled latent variables, we adopt the total correlation (TC) loss~\cite{factor_vae18,tcvae}, $\textrm{TC} = \textrm{KL} \big( Q({\bf z}) \Vert \prod_{j} Q(z_j)) \big)$ where $Q({\bf z}) = \mathbb{E}_{({\bf x}_1,{\bf x}_2)\sim \textrm{Data}} [Q({\bf z}|{\bf v}_1,{\bf v}_2)]$. 
Estimating the TC loss can be done by the density ratio estimation proxy~\cite{density_ratio_jordan,density_ratio_sugiyama} followed by adversarial training~\cite{factor_vae18}, or the weighted sampling strategy~\cite{tcvae}. 
Combining the two leads to the following objective function for the Bi-VAE model: 
%%%%
\begin{equation}
\mathcal{L}_{\textrm{Bi-VAE}}(\Theta_\textrm{Bi-VAE}) = -\textrm{ELBO} + \gamma \cdot \textrm{TC},
\label{eq:supp_obj_vae}
\end{equation}
%%%%
%where $\gamma$ is the impact of the TC loss. 
where $\Theta_\textrm{Bi-VAE}$ in (\ref{eq:supp_obj_vae}) refers to all parameters of the Bi-VAE model (i.e., parameters of ${\boldsymbol\mu}_{1/2}(\cdot)$, ${\boldsymbol\sigma}_{1/2}(\cdot)$, ${\bf m}_{1/2}(\cdot)$, ${\bf s}_{1/2}(\cdot)$), and $\gamma$ %is the hyperparameter that 
trades off the TC loss against the ELBO. We use $\gamma=10.0$ for all datasets except $\gamma=100.0$ for Recipe1M. 

The cross-modal retrieval loss, similar to our RIVAE, is also adopted to train both embedding networks and the Bi-VAE model. The related loss is:
%%%%
\begin{equation}
\mathcal{L}_\textrm{Retrieval}(\Theta_\textrm{Bi-VAE},\Theta_\textrm{Emb}) = \Big( 1 + \mathbb{E}_{Q({\bf z}|{\bf v}_1)} \big[ \log P({\bf v}'_2 | {\bf z}) - \log P({\bf v}_2 | {\bf z}) \big] \Big)_+ 
\label{eq:supp_retr_loss} 
\end{equation}
%%%%
where ${\bf v}'_2 = {\bf e}_2({\bf x}'_2)$ is the embedding of the mismatch sample ${\bf x}'_2$,  $\Theta_\textrm{Emb}$ indicates the parameters of the embedding networks ${\bf e}_1(\cdot)$ and ${\bf e}_2(\cdot)$, and $(a)_+ = \max(0,a)$. 
Finally, we impose the embedding regularization loss for ${\bf e}_2(\cdot)$, similar to RIVAE:
%%%%
\begin{equation}
\mathcal{L}_\textrm{Reg}(\Theta_\textrm{Emb}) = \mathbb{E}_{{\bf x}_2,{\boldsymbol\epsilon}} \Big[ \big(||{\bf e}_2({\bf x}_2) - {\bf e}_2({\bf x}_2+{\boldsymbol \epsilon})|| - c \big)^2 \Big]
\label{eq:supp_emb_reg}
\end{equation}
%%%%
where $\boldsymbol{\epsilon} \sim P(\boldsymbol{\epsilon})$ is a random sample from a noise distribution $P(\boldsymbol{\epsilon})$ with small magnitude (e.g., $||\boldsymbol{\epsilon}||=0.001$). Then the training of the RBi-VAE can be written as the following optimization:
%%%%
\begin{equation}
%\max_{\textrm{VAE}} & \ \ -\textrm{ELBO} + \gamma \cdot \textrm{TC} \\
%\min_{\textrm{VAE},\textrm{Emb}} \ \ \mathcal{L}_\textrm{Retrieval} + \lambda \cdot \textrm{Reg}({\bf e}_2)
\min \ \mathcal{L}_\textrm{Bi-VAE}(\Theta_\textrm{Bi-VAE}) + \lambda_\textrm{Retrieval} \mathcal{L}_\textrm{Retrieval}(\Theta_\textrm{Bi-VAE}, \Theta_\textrm{Emb}) +  \lambda_\textrm{Reg} \mathcal{L}_\textrm{Reg}(\Theta_\textrm{Emb}). 
\label{eq:supp_obj_joint}
\end{equation}
%%%%
Note that the arguments in each loss $\mathcal{L}(\cdot)$ indicates which parameters should be updated regarding the loss.

%%%%%%%%%%%%%%%%%%%%%%%%%%%%%%%%%%%%%%%%%%%%%%%%%%%%%%%%
%%%%%%%%%%%%%%%%%%%%%%%%%%%%%%%%%%%%%%%%%%%%%%%%%%%%%%%%

% %%%%%%%%%%%%%%%%%%%%%%%%%%%%%%%%%%%%%%%%%%%%%%%%%%%%%%%%
% \begin{figure}%[t!]
% \raisebox{-.5\height}{\includegraphics[height=3.16cm]{figs/R1M/traversals/dim1/dimension-1_anchor-0_traversal.jpg}} \vspace{0.3em}\\ %\hline \hline
% \raisebox{-.5\height}{\includegraphics[height=3.16cm]{figs/R1M/traversals/dim1/dimension-1_anchor-5_traversal.jpg}} \vspace{0.3em}\\ %\hline\hline
% \raisebox{-.5\height}{\includegraphics[height=3.16cm]{figs/R1M/traversals/dim1/dimension-1_anchor-8_traversal.jpg}} \vspace{0.3em}\\ %\hline\hline
% \raisebox{-.5\height}{\includegraphics[height=3.16cm]{figs/R1M/traversals/dim1/dimension-1_anchor-10_traversal_2.png}} \vspace{0.3em}\\ %\hline\hline
% \raisebox{-.5\height}{\includegraphics[height=3.16cm]{figs/R1M/traversals/dim1/dimension-1_anchor-15_traversal.jpg}} \\ 
% \vspace{-0.0em}
% \caption{Images of the retrieved recipes from latent traversal along $z_1$ (non noodle -- noodle). The same interpretation as Fig.~\ref{fig:supp_traversal0}.
% %(Rows) Retrieved images for latent traversal, along $z_1$ (non noodle - noodle). \textbf{Image grids}: query image (left), each column top 3 NNs at a point in traversal $[\mu_i-10\sigma_i,\mu_i+10\sigma_i]$. Note that NNs are only shown for those points with a unique top 1, therefore, there are different amounts of retrieved images depending on the query.
% }
% \label{fig:supp_traversal1}
% \vspace{-0.5em}
% \end{figure}
% %%%%%%%%%%%%%%%%%%%%%%%%%%%%%%%%%%%%%%%%%%%%%%%%%%%%%%%%

%%%%%%%%%%%%%%%%%%%%%%%%%%%%%%%%%%%%%%%%%%%%%%%%%%%%%%%%
%%%%%%%%%%%%%%%%%%%%%%%%%%%%%%%%%%%%%%%%%%%%%%%%%%%%%%%%

%%%%%%%%%%%%%%%%%%%%%%%%%%%%%%%%%%%%%%%%%%%%%%%%%%%%%%%%
%%%%%%%%%%%%%%%%%%%%%%%%%%%%%%%%%%%%%%%%%%%%%%%%%%%%%%%%

%%%%%%%%%%%%%%%%%%%%%%%%%%%%%%%%%%%%%%%%%%%%%%%%%%%%%%%%
%%%%%%%%%%%%%%%%%%%%%%%%%%%%%%%%%%%%%%%%%%%%%%%%%%%%%%%%

%\newpage

\end{document}